# A deep learning framework for efficient pathology image analysis

Peter Neidlinger (1), Tim Lenz (1), Sebastian Foersch (2), Chiara M. L. Loeffler (1, 3, 4), Jan Clusmann (1, 5), Marco Gustav (1), Lawrence A. Shaktah (1), Rupert Langer (6), Bastian Dislich (7), Lisa A. Boardman (8), Amy J. French (9), Ellen L. Goode (10), Andrea Gsur (11), Stefanie Brezina (11), Marc J. Gunter (12, 13), Robert Steinfelder (14), Hans-Michael Behrens (15), Christoph Röcken (15), Tabitha Harrison (14, 16), Ulrike Peters (14, 16), Amanda I. Phipps (14, 16), Giuseppe Curigliano (17, 18), Nicola Fusco (18, 19), Antonio Marra (17), Michael Hoffmeister (20), Hermann Brenner (20, 21), Jakob Nikolas Kather+ (1, 3, 22)

+ Correspondence to jakob-nikolas.kather@alumni.dkfz.de

1. Else Kroener Fresenius Center for Digital Health, Faculty of Medicine and University Hospital Carl Gustav Carus, TUD Dresden University of Technology, 01307 Dresden, Germany
2. Institute of Pathology, University Medical Center Mainz, Mainz, Germany
3. Department of Medicine I, Faculty of Medicine and University Hospital Carl Gustav Carus, TUD Dresden University of Technology, 01307 Dresden, Germany
4. National Center for Tumor Diseases Dresden (NCT/UCC), a partnership between DKFZ, Faculty of Medicine and University Hospital Carl Gustav Carus, TUD Dresden University of Technology, and Helmholtz-Zentrum Dresden - Rossendorf (HZDR), Dresden, Germany
5. Department of Medicine III, University Hospital RWTH Aachen, Aachen, Germany
6. Institute of Pathology and Molecular Pathology, Kepler University Hospital, Johannes Kepler University Linz, Linz, Austria
7. Institute of Tissue Medicine and Pathology, University of Bern, Bern, Switzerland
8. Division of Gastroenterology and Hepatology, Mayo Clinic, Rochester, Minnesota, USA.
9. Division of Laboratory Genetics, Department of Laboratory Medicine and Pathology, Mayo Clinic, Rochester, Minnesota, USA.
10. Department of Quantitative Health Sciences, Division of Epidemiology, Mayo Clinic, Rochester, Minnesota, USA.
11. Center for Cancer Research, Medical University of Vienna, Vienna, Austria.
12. Nutrition and Metabolism Branch, International Agency for Research on Cancer, World Health Organization, Lyon, France.
13. Cancer Epidemiology and Prevention Research Unit, School of Public Health, Imperial College London, London, United Kingdom.
14. Division of Public Health Sciences, Fred Hutchinson Cancer Center, Seattle, WA, USA.
15. Department of Pathology, University Hospital Schleswig-Holstein, Kiel, Germany
16. Department of Epidemiology, University of Washington, Seattle, WA, USA.
17. Division of New Drugs and Early Drug Development, European Institute of Oncology IRCCS, Milan, Italy
18. Department of Oncology and Hemato-Oncology, University of Milan, Italy
19. Division of Pathology, European Institute of Oncology IRCCS, Milan, Italy
20. Division of Clinical Epidemiology and Aging Research, German Cancer Research Center (DKFZ), Heidelberg, Germany
21. German Cancer Consortium (DKTK), German Cancer Research Center (DKFZ), Heidelberg, Germany
22. Medical Oncology, National Center for Tumor Diseases (NCT), University Hospital Heidelberg, Heidelberg, Germany

# Abstract

Artificial intelligence (AI) has transformed digital pathology by enabling biomarker prediction from high-resolution whole-slide images (WSIs). However, current methods are computationally inefficient, processing thousands of redundant tiles per WSI and requiring complex aggregator models. We introduce EAGLE (Efficient Approach for Guided Local Examination), a deep learning framework that emulates pathologists by selectively analyzing informative regions. EAGLE incorporates two foundation models: CHIEF for efficient tile selection and Virchow2 for extracting high-quality features. Benchmarking was conducted against leading slide- and tile-level foundation models across 43 tasks from nine cancer types, spanning morphology, biomarker prediction, treatment response and prognosis. EAGLE outperformed state-of-the-art patch aggregation methods by up to 23% and achieved the highest AUROC overall. It processed a slide in 2.27 seconds, reducing computational time by more than 99% compared to existing models. This efficiency enables real-time workflows, allows rapid review of the exact tiles used for each prediction, and reduces dependence on high-performance computing, making AI-powered pathology more accessible. By reliably identifying meaningful regions and minimizing artifacts, EAGLE provides robust and auditable outputs, supported by systematic negative controls and attention concentration analyses. Its unified embedding enables rapid slide searches, integration into multi-omics pipelines and emerging clinical foundation models.

# Main

Artificial intelligence (AI) has significantly advanced computational pathology (CPath) by enabling the extraction of clinically relevant information from gigapixel-scale whole-slide images (WSIs)[1–6]. Existing methods use resource-intensive vision transformers trained with self-supervised learning (SSL) to encode detailed morphological features essential for diagnosis, prognosis, and treatment planning in oncology[7–11]. While these approaches have shown great promise across a wide range of tasks, their inefficiencies and limited scalability highlight the need for solutions that better align with real-world diagnostic workflows. Recently, pathology-specific multimodal large language models (MLLMs) have emerged as AI copilots for clinical decision-making, but they often underperform in biomarker prediction and the regulatory pathway for approving such models as medical devices remains uncertain[12–16].

Current methods predominantly operate at the tile level, requiring the extraction and analysis of thousands of tiles per WSI, with datasets in this study averaging approximately 18,000 tiles per slide at a resolution of 0.5 µm/pixel (MPP). This computationally intensive process deviates from how pathologists evaluate slides, as they selectively focus on regions of interest[17–19]. Moreover, tile-wise features are aggregated into slide-level predictions using models trained separately for each task, limiting scalability and interpretability[8,20,21]. The complexity of these models often obscures the decision-making process, making it challenging to understand how predictions are derived and which tissue regions are influential. These systems also struggle in data-scarce scenarios, where tile selection often fails to identify the most relevant regions, leading to suboptimal predictions[22]. Such scenarios are often a clinical reality, for example during the evaluation of small biopsy specimens.

To address these limitations, we developed EAGLE (Efficient Approach for Guided Local Examination), a framework that emulates the diagnostic strategy of pathologists by focusing on a small, informative subset of tiles within WSIs. EAGLE combines CHIEF[23], a pre-trained and task-agnostic model used for global tissue representation and guided tile selection, with Virchow2[24], for detailed feature extraction from selected tiles. This combination drastically reduces computational demands while increasing performance (**Fig. 1a-c**). By selecting a small, reproducible subset of regions, EAGLE enhances auditability and scalability, particularly in biomarker prediction tasks where subtle morphological features are critical[25]. Unlike MLLMs, which emphasize multimodal interaction, EAGLE prioritizes efficient high-quality WSI analysis. Still, it can integrate with MLLMs to provide valuable inputs for enhanced contextual analysis. Through comprehensive evaluation against state-of-the-art models, including multiple instance learning (MIL) and slide-encoder approaches, we demonstrate the efficacy and robustness of EAGLE across 43 tasks spanning nine cancer types[8–10,23,24,26–30].

# Results

## EAGLE improves upon state-of-the-art approaches

We first evaluated EAGLE within a previously established benchmarking framework encompassing 31 CPath tasks across breast (BRCA), colorectal (CRC), gastric (STAD), and non-small cell lung cancers (NSCLC)[22]. This dataset integrates diverse prediction objectives spanning morphology, biomarker and prognosis, and enables systematic comparison on large, truly external datasets. The benchmark includes both state-of-the-art slide-encoder and tile-encoder approaches: slide-level models (TITAN[30], COBRA[29], CHIEF[23], Prism[26], MADELEINE[27], and Prov-GigaPath[9]) and tile encoders or foundation models (Virchow2[24], CONCH v1.5[30], CONCH[28], Prov-GigaPath[9], CTransPath[8], Virchow[10]), which were aggregated using attention-based multiple instance learning (ABMIL)[21], the standardized pipeline STAMP[18], or simply by averaging all embeddings. All classifiers were trained using a five-fold cross-validation setup on The Cancer Genome Atlas (TCGA) data, resulting in five models per task. Each model was then evaluated on the full external test cohorts (CPTAC, DACHS, Kiel, Bern, IEO), ensuring external validation without data leakage.

Across all 31 tasks, EAGLE and TITAN achieved the highest average area under the receiver operating characteristic curve (AUROC) scores of 0.742 and 0.740, respectively. Following were Virchow2 in STAMP (0.723) and CONCH v1.5 in STAMP (0.721), suggesting that modern slide encoders often outperform strong tile-level baselines (**Fig. 1c, 2a, Extended Data Fig. 2**). EAGLE exceeded key AUROC thresholds more often than other models, surpassing 0.800 in 39% of tasks and 0.650 in 77% of tasks—higher than TITAN (35% and 68%) and Virchow2 (26% and 65%) (**Fig. 2b**). Looking at task-specific performance, EAGLE (0.772), TITAN (0.763), and COBRA (0.757) excelled on biomarker tasks, which predict molecular alterations or protein expression. In morphology tasks, which classify tumor location or histological patterns, TITAN achieved the highest performance (0.814), followed by Virchow2 (0.785) and EAGLE (0.782). Tile-level CONCH v1.5 in STAMP (0.648) held an advantage in prognosis, which assesses tumor spread, with EAGLE achieving the highest prognosis performance among slide-level frameworks (0.630) (**Fig. 2c,Extended Data Fig. 3a,e**). EAGLE also scored highest in three of the four cancer types—BRCA (0.737), CRC (0.710), and STAD (0.755)—while TITAN scored highest in lung (0.810) (**Extended Data Fig. 3b,c,f**). Beyond AUROC, EAGLE achieved the highest average area under the precision-recall curve (AUPRC) scores (0.566), followed by COBRA (0.556). In balanced accuracy, EAGLE and TITAN performed best with average scores of 0.657 and 0.655, respectively. TITAN held the highest average F1 scores (0.498), followed by EAGLE (0.490) (**Supplementary Fig. 1-3**).

We assessed the statistical significance of AUROC differences with two-sided DeLong's tests, controlling the false discovery rate using the Benjamini–Hochberg procedure. For each patient, predictions from the five cross-validation models were averaged to yield a single ensemble score. Across all tasks and models, ensemble AUROC values were highly correlated with mean fold performance, with most model–task pairs lying close to the diagonal, indicating only marginal performance gains from ensembling (**Extended Data Fig. 4a**). Under these ensemble conditions, EAGLE achieved an average AUROC of 0.750, followed by TITAN (0.744) and CONCH (0.736) (**Supplementary Fig. 4a**). Statistical comparisons based on ensemble predictions confirmed that EAGLE significantly outperformed slide-level encoders across numerous tasks, with AUROC improvements exceeding 0.05 and occasionally 0.2 (**Extended Data Fig. 4b**). Similar trends were observed when comparing EAGLE with tile-level foundation models (**Extended Data Fig. 4c**). When restricting the comparison to the Virchow2 model aggregated through STAMP, ABMIL, or mean pooling, differences were smaller, with EAGLE outperforming in six task–model combinations and performing worse in four (**Extended Data Fig. 4d**). EAGLE's strongest advantages were observed for DACHS CRC tasks (MSI, *BRAF*, *KRAS*, M-Status, and N-Status), potentially due to the large cohort size, which provided increased statistical power. In DACHS *BRAF* mutation prediction, EAGLE statistically outperformed all models except COBRA (**Supplementary Fig. 4b**).

Together, these analyses show that EAGLE achieves consistently high performance across tasks, cancer types, and external cohorts, often statistically outperforming state-of-the-art slide- and tile-level models. Its strong biomarker prediction accuracy and robust generalization across heterogeneous datasets highlight the quality and versatility of its representations.

## Design analysis of EAGLE and negative controls

To understand which design choices yield EAGLE's performance gains, we conducted extensive ablation studies. First, we compared five different options for aggregating Virchow2 tile embeddings at 2 MPP. EAGLE selects the 25 most informative tiles using CHIEF and averages their embeddings, whereas alternative approaches either aggregate tiles through supervised models (STAMP, ABMIL or gated ABMIL) or compute a simple mean across all tiles. Among these, EAGLE achieved the highest mean AUROC (0.742), followed by gated ABMIL (0.723), STAMP (0.723), mean pooling (0.720), and regular ABMIL (0.711), confirming that EAGLE provides the most effective tile aggregation strategy. (**Fig. 3a, Extended Data Fig. 5a**). Second, we varied the tile encoder while keeping EAGLE aggregation fixed. Performance remained robust across encoders, indicating that EAGLE benefits from stronger tile features but does not rely on a single backbone. We then asked whether EAGLE can replace bespoke slide encoders trained for specific tile embeddings. Across four backbones, EAGLE exceeded the native slide encoders for three and trailed slightly for one. Gains were largest with GigaPath (+12%) and CONCH (+4%), modest with CTransPath (+1%), and negative with Virchow (-2%). Pooled across all comparisons, EAGLE led in about two thirds of tasks with a small increase in mean AUROC, supporting its role as an effective universal slide-level aggregator (**Fig. 3b,d, Extended Data Fig. 5b,c**). Third, we systematically analyzed how the number and selection strategy of tiles affect EAGLE's performance. We compared three approaches: CHIEF-based selection of the top-ranked tiles aggregated either (i) by simple averaging (unweighted) or (ii) by CHIEF-derived attention weights (weighted), and (iii) random tile selection as a negative control repeated 100 times per tile budget. For both CHIEF-based strategies, performance increased up to 25 tiles, where the highest AUROC was reached (0.745 for unweighted, 0.744 for weighted), indicating that a small, targeted subset captures most predictive information. Notably, using only the top five selected tiles (0.727) already outperformed mean pooling of all embeddings (0.720). A 100-replicate random baseline defined an empirical null distribution for each tile budget ( N = 5, 10, 25, 50, 100), enabling uncertainty estimation and Monte Carlo testing. The mean AUROC of the replicate distribution increased with larger tile budgets, approaching the mean pooling baseline as N increased. For every N, CHIEF-based selection exceeded the maximum random replicate, corresponding to a Monte Carlo p-value of 1/101 and excluding the hypothesis that any reasonable unguided subset performs comparably (**Fig. 3c,**

**Supplementary Table 1**). Because weighted and unweighted aggregation performed similarly at 25 tiles, we adopted equal averaging of the top 25 tiles as the default to enable transparent region-level auditability (**Fig. 3c, Extended Data Fig. 5e**). To address intra-patient heterogeneity, we next compared two methods for creating a patient-level representation. Early fusion aggregates across all slides jointly before forming a single patient embedding. Late fusion averages slide-level embeddings after separate aggregation per slide. On average the difference was small with mean delta equals −0.005, although early fusion tended to help for morphology and prognosis tasks while biomarker tasks showed little change. (**Extended Data Fig. 5d**). Moreover, we examined how magnification choices influence performance. While EAGLE and COBRA each peaked at 2 MPP (0.742 and 0.719, respectively), TITAN and Prism favored 0.5 MPP (0.740 and 0.695). For Prov-GigaPath (0.628), CHIEF (0.684) and MADELEINE (0.672), 1.14 MPP worked best, implying that the optimal resolution differs between models (**Extended Data Fig. 5f**). Finally, to address concerns about potential bias from suboptimal hyperparameters, we performed extensive MLP hyperparameter tuning using only internal validation cohorts. On internal data, tuning led to modest AUROC gains of 0.02–0.03 on average, with larger improvements for lower-performing models such as GigaPath (+0.05) and MADELEINE (+0.037). However, these gains did not generalize to external test cohorts, where performance changes were negligible (<0.01 on average across models) (**Fig. 3e**). Correlation analysis between internal and external AUROC deltas ($R^2 = 0.04$) confirmed that tuning improvements largely reflect overfitting rather than transferable optimization. These findings indicate that extensive hyperparameter tuning provides little benefit for large embedding-based models in CPath (**Extended Data Fig. 6**).

Together, these investigations highlight that EAGLE's performance gains arise from its architecture and aggregation strategy rather than hyperparameter bias. Selecting and averaging the most informative tiles provides clear advantages over established aggregation methods, slide encoders, and random or exhaustive tile inclusion. EAGLE's approach remains robust across magnifications and effectively integrates multiple slides per patient, underscoring the stability and generalizability of its representations.

## End-to-end weak supervision drives attention toward near-uniform aggregation

To directly test whether EAGLE's fixed, task-agnostic tile selection is a principled response to a measurable limitation of end-to-end weak supervision, we quantified attention concentration for CHIEF, ABMIL, and a strengthened gated ABMIL baseline. CHIEF attention scores were obtained from the frozen CHIEF model operating on CTransPath embeddings, whereas ABMIL and gated ABMIL attention scores were obtained from task-specific models trained on Virchow2 embeddings. For each method, we computed Lorenz curves of cumulative attention mass as a function of the fraction of tiles ranked by attention, along with summary statistics including the Gini coefficient, the fraction of tiles required to accumulate 50% and 80% of the attention mass, and top-k mass metrics (**Fig. 4, Supplementary Table 2**). These analyses were performed per patient and then aggregated across tasks and folds by averaging attention rank profiles rather than tile identities, preventing trivial averaging effects.

Across tasks, CHIEF induced a stable non-uniform saliency ordering. The median fraction of tiles required to accumulate 50% of attention mass was 8.4% for CHIEF, compared with 44.1% for ABMIL and 42.0% for gated ABMIL, with corresponding 80% mass fractions of 20.5%, 75.8%, and 73.6% (**Fig. 4a,f, Supplementary Table 2**). This pattern was strongest for subtle biomarker endpoints, where (gated) ABMIL attention frequently remained broadly distributed and approached mean pooling behavior, but it became more selective for overt morphology endpoints such as NSCLC subtyping and CRC sidedness (**Fig. 4b-d**). Consistent with these concentration curves, the mean Gini coefficient was 0.702 for CHIEF versus 0.087 for ABMIL and 0.104 for gated ABMIL (**Fig. 4e**).

To assess whether CHIEF-based selection is dominated by very few tiles, we quantified cumulative attention mass in the top 1, 2, and 25 ranked tiles. These metrics indicate that CHIEF attention is concentrated yet not driven by a single region (**Fig. 4g,h, Supplementary Table 2**). Finally, we provide direct visual comparisons on the same slides, showing CHIEF and gated ABMIL attention maps with explicit overlays of the exact top 25 tiles used by EAGLE, which makes the difference between concentrated saliency rankings and diffuse task-specific attention immediately observable (**Fig. 5, Supplementary Figures 6-7**).

## Generalization to additional cancer types and complex clinical tasks

To evaluate EAGLE's generalization beyond the benchmarking tasks, we next applied it to the PathoBench dataset, which includes 12 survival and treatment response prediction tasks covering five additional cancer types[31,32]. In contrast to the benchmarking framework, PathoBench does not include external cohorts, and training and testing were performed within the same cohort, often with smaller sample sizes (**Fig. 6a,b**). The seven survival tasks comprise one progression-free (BOEHMK PFS) and six overall survival (OS) endpoints (SURGEN, LUAD, HNSC, PDA, CCRCC, and MBC), while the five treatment-response tasks assess binary or multi-class outcomes including five-year mortality, RECIST-based therapy response, lymphovascular invasion, and drug sensitivity.

Across the seven survival prediction tasks, EAGLE achieved the highest mean concordance index (C-index = 0.584), followed by PRISM (0.582) and TITAN (0.577) (**Fig. 6c**). For treatment-response prediction, EAGLE again outperformed all other approaches, reaching a mean AUROC of 0.689, while MADELEINE (0.676) and ABMIL (0.646) were the second- and third-best performing models, respectively (**Fig. 6d**). To further assess prognostic stratification, we analyzed model-derived risk scores obtained from the CoxNet survival models. For each model, patients were divided at the median predicted risk score into high- and low-risk groups, and Kaplan–Meier curves were generated to visualize survival separation. On the CPTAC PDA cohort, EAGLE achieved clear stratification between high- and low-risk patients (hazard ratio = 2.02, p = 0.0038), whereas TITAN (1.18), CHIEF (1.16), and GigaPath (1.02) showed no significant separation (p > 0.5) (**Fig. 6e**). Across additional survival tasks, EAGLE frequently showed stronger risk group separation than other models, including high hazard ratios on CPTAC CCRCC (2.3) and HNSC (1.88) (**Extended Data Fig. 7**).

Overall, these results demonstrate that EAGLE generalizes effectively to new cancer types and clinically relevant endpoints, including complex survival and treatment response tasks.

## Efficiency and Performance in Data-Scarce Scenarios

We next quantified computational requirements for the major pipeline steps, measuring average inference times for tile extraction and slide encoding using 25 representative slides of the benchmarking dataset (**Fig. 7a**). All measurements were performed on a workstation equipped with an L40 GPU (48GB), considering only model inference times. The slowest step is the tile-level feature extraction. On average, CTransPath at 0.5 MPP required 25.4 s/WSI and only 2.01 s/WSI at 2 MPP. By contrast, CONCH v1.5 at 0.5 MPP cost 191.85 s/WSI, and Prov-GigaPath 16 min/WSI (**Supplementary Table 3**). The faster slide encoding took on average 0.36 ms for CHIEF (2 MPP) , while TITAN (0.5 MPP) took 3296 ms and Prism (0.5 MPP) 153 ms. EAGLE first processed each WSI with CTransPath at 2 MPP (2.01 s/WSI), applied CHIEF (0.36 ms/WSIs), and selected 25 key tiles. Those tiles were then re-extracted with Virchow2 (0.26 s/WSI). On average, ~2% of tiles are reprocessed in detail at 2 MPP using Virchow2 (or ~0.1% at 0.5 MPP). When plotted against overall AUROC, EAGLE obtained the highest performance while ranking among the most efficient in terms of run time and floating point operations (FLOPs) (**Fig. 7b-d**). Prov-GigaPath, in contrast, was both the most time-consuming and the poorest-performing. TITAN, though competitive with EAGLE, required markedly more compute (second slowest overall).

Medical imaging datasets often contain limited samples per class, prompting few-shot or low-resource scenarios. We first assessed linear probing on EAGLE (and other slide encoder) embeddings with k = 1, 2, 4, 8, 16, 32 samples per class using logistic regression. To ensure meaningful comparisons, we selected only the top three binary tasks per cancer type based on AUROC in the main results, as linear probing performs poorly on low-performance tasks, increasing randomness and diluting differences between models. While performance dips in this challenging setting, both EAGLE and TITAN consistently outperformed others. EAGLE performed best in all cancer types except lung, where Prism excelled. With k = 1, 2, 4, TITAN lead across all 12 tasks, followed closely by EAGLE, with both models achieving notable gaps over others (e.g., k=4: TITAN/EAGLE AUROC = 0.72, Prism = 0.68, Prov-GigaPath = 0.58). At k = 8, 16, 32, EAGLE surpassed all other models, showcasing the advantage of its focused tile selection in low-data settings (**Fig. 2d, Extended Data Fig. 8b,d**).

We further evaluated EAGLE and other slide-encoder vs. tile-encoder models by training regular multilayer perceptron (MLP) classifiers on 300, 150, or 75 patients. Over 29 tasks with sufficient data, EAGLE maintained the highest average AUROC across all three subsets, particularly at 150 patients (0.689 vs. 0.669 for TITAN and 0.618 for the best tile encoder) (**Fig. 7e, Extended Data Fig. 8a,c**). This robust performance gap highlights how slide-level embeddings support model training when only small cohorts are available. To demonstrate EAGLE's practicality for rare-biomarker discovery, we examined almost 300 biomarkers in a multicenter cohort of over 1,000 patients. The entire processing time from the raw images to the final predictions took <24 hours and we detected 20 biomarkers with AUROCs >0.800 (**Fig. 7f, Supplementary Table 4**).

Together, these efficiency metrics confirm that EAGLE's guided, two-step approach minimizes computation without compromising performance and scales well to large or multi-task pathology pipelines. EAGLE's adaptability in data-constrained contexts, indicate that the combination of slide-level embeddings, selective tile processing, and efficient feature extraction yields robust performance when training data are limited.

## Versatility and Interpretability of EAGLE

EAGLE generates a single, compact embedding per whole-slide or patient by restricting computation to a small number of highly informative tiles. The selection of only 25 top tiles, those identified as most relevant by the model, makes it explicit which tissue regions are used to construct the slide representation and can therefore be reviewed. To illustrate this interpretability advantage, we examined a subset of 50 randomly selected WSIs from the DACHS CRC cohort for MSI prediction, where 35 WSIs contained pen marks. Across all top tiles of these 35 WSIs, pen marks were present in 16% of EAGLE-selected tiles compared to 23% of tiles selected by the supervised baseline, which uses Virchow2 tile embeddings aggregated with STAMP. Notably, when considering only dominant pen marks (covering >50% of the tile), EAGLE almost completely avoided them (1%), whereas the supervised baseline selected those tiles in 15% of cases, even though they did not provide valuable information. Expanding the analysis to include all artifacts (e.g., tissue folds, slide edges, air bubbles, oil drops, scratches, pen marks, foreign objects, dark spots, and out-of-focus regions), EAGLE-selected tiles showed artifacts in 22% of cases, compared to 32% in the tiles selected by the supervised baseline (**Fig. 8b,c**). Furthermore, a board-certified pathologist inspected top selected tiles and found that EAGLE reliably focused on the most representative tumor tissue of each case. In contrast, the supervised baseline often focused on a mixture of healthy, tumorous and artifact-rich tiles, providing puzzling and sometimes contradictory features for its prediction (**Extended Data Fig. 9a**). This capacity to confirm which tiles are chosen could further support clinical acceptance.

We then visualized slide embeddings with Uniform Manifold Approximation and Projection (UMAP)[33]. We noted that TITAN formed well-separated clusters by cancer type, while EAGLE displayed moderate clustering and surpassed CHIEF and Prov-GigaPath Slide Encoder in capturing morphological diversity (**Fig. 8a**). To further evaluate embedding structure, we compared UMAP projections of CHIEF and EAGLE across 29 TCGA cohorts, observing clearer separation with EAGLE (**Extended Data Fig. 9b**). Moreover, we explored slide retrieval applications, wherein a query WSI's embedding was compared against a database of stored slide embeddings for quick retrieval of similar cases, thereby enabling pathologists to rapidly find relevant reference slides, accelerating diagnosis or facilitating training[34,35]. This approach was computationally lightweight, with near-instant matches. Qualitative reviews by a board-certified pathologist indicated that the slide search approach yielded highly similar cases both within a given cohort and across multiple cohorts. Interestingly, in a few instances, cases from other entities would be identified (e.g. a BRCA case while looking for a STAD case). This happened, when special subtypes were queried (e.g. a medullary carcinoma, which can occur both in the stomach as well as in the mammary gland) (**Fig. 8d**). This could prove to be a very valuable tool, for example for cohort assembly for basket trials.

Together, these data indicate that EAGLE enables more transparent slide-level decision-making and simplifies downstream analyses such as multi-omic and clinical integration, and efficient retrieval of relevant comparator slides (**Fig. 1d**).

## Comparison with Multimodal Large Language Models (MLLMs)

Finally, we compared EAGLE with emerging MLLMs employing in-context learning low-resolution WSI thumbnails. Using k=2 examples per class in a few-shot classification setup, EAGLE embeddings combined with a lightweight logistic regression model surpassed GPT-4o's in-context classification for tasks including NSCLC subtyping, MSI prediction in CRC, and ER expression in BRCA. In NSCLC subtyping and ER expression prediction, GPT-4o usually defaulted to a single class (e.g., predicting adenocarcinoma in NSCLC and ER-positive in BRCA), while in all three tasks, including MSI prediction, its classification accuracy remained at chance level (0.5). To evaluate whether input resolution constrained GPT-4o's performance, we provided the model with EAGLE's top 25 most informative tiles instead of WSI thumbnails. Despite this targeted input, MLLM accuracy remained unchanged, suggesting that the bottleneck lies in the models' lack of pathology-specific feature representations rather than the resolution or relevance of the input (**Fig. 2e, Extended Data Fig. 10a**). Interestingly, despite its poor predictive accuracy, GPT-4o's textual re-sponses frequently referenced key diagnostic features, such as “mucin production” and “keratinization” for NSCLC subtyping, “lymphocytic infiltration” for MSI status prediction, indicating an awareness of relevant pathological concepts (**Extended Data Fig. 10b**).

Taken together, these results underscore the limitations of general-purpose MLLMs in specialized pathology tasks and highlight the need for domain-adapted models. While MLLMs such as GPT-4o demonstrate impressive contextual reasoning and linguistic capabilities, their ability to generalize to specialized pathology biomarker tasks is notably limited.

# Discussion

Weakly supervised CPath has fundamentally reshaped cancer research, allowing diagnosis, biomarker status prediction, and survival outcome estimation directly from WSIs. Recent progress with pathology foundation models, often trained using SSL, has enriched these pipelines, enhancing accuracy and generalizability across diverse clinical tasks. However, concerns remain about the scalability of tile-level methods and the difficulty of interpreting predictions when massive neural networks process thousands of patches per slide. Here, we introduced EAGLE, a slide-level framework that emulates how pathologists target limited, high-yield regions.

EAGLE’s design centers on processing only 25 tiles per WSI, a strategy that substantially reduces computational requirements while preserving predictive power. This design is not a heuristic truncation but reflects a bias–variance trade-off under weak supervision: when the predictive signal is spatially sparse relative to total tissue area, restricting inference to a reproducible, high-saliency subset can improve statistical conditioning. This approach aligns with established clinical practice, in which pathologists focus on diagnostically relevant or suspicious tissue regions rather than exhaustively scanning each tile. Our results show that EAGLE cuts feature extraction times dramatically compared to tile-level pipelines relying on resource-intensive foundation models. These efficiency gains translate to quicker classifier training and inference, making EAGLE especially suited for large-scale or multi-institutional datasets. Such performance improvements are likely to be essential for real-time or near-real-time diagnostic workflows. By substantially reducing computational requirements, EAGLE brings the deployment of digital pathology tools on lower-resource hardware, including tablets and potentially smartphones, within reach. This improved efficiency broadens accessibility, lowers infrastructure barriers and supports adoption in a wider range of clinical and resource-limited settings. Our comparative evaluations indicate that EAGLE frequently outperforms or matches leading slide- and tile-encoder baselines, particularly in biomarker prediction tasks central to precision oncology. Whereas many CPath models demonstrate high performance for commonly studied tasks such as tumor subtyping, they can struggle in more challenging settings like molecular status prediction for biomarkers or survival analyses. EAGLE addresses these obstacles by focusing on the regions that matter most morphologically, offering refined predictive performance. Our benchmarking across 31 externally validated tasks shows that combining CHIEF and Virchow2 in EAGLE outperforms both individual models in the majority of cases and is never worse than both on any task. Targeted selection of informative tiles proved essential for maximizing accuracy, whereas selecting more than 25 tiles tended to introduce noise and reduce performance. The strength of this approach also reflects the extensive pretraining of CHIEF on more than 60,000 slides, a scale not feasible for typical task-specific datasets in digital pathology. Importantly, performance gains cannot be attributed to arbitrary subsampling. Across multiple tile budgets, saliency-guided selection consistently outperformed repeated uniform random selection under identical splits, indicating that the improvement arises from structured region ranking rather than from processing fewer tiles. Quantitative analysis of attention concentration further showed that the pretrained saliency prior induces a stable and non-uniform ordering of tissue regions, whereas task-specific attention learning frequently approaches near-uniform aggregation for subtle biomarker endpoints. Across models, extensive hyperparameter tuning provided only negligible benefit on external test cohorts, suggesting that performance in large embedding-based CPath models is primarily determined by representation quality rather than classifier optimization. Beyond the benchmarking study, EAGLE maintained strong performance on the PathoBench survival and treatment response tasks, demonstrating its applicability to more complex clinical endpoints.

A hallmark of EAGLE's design is the generation of a single embedding per slide or patient. Although this embedding remains abstract, EAGLE provides explicit spatial localization because it is computed solely from the features of the top selected tiles, all of which contribute equally to the prediction. While sparsity alone does not guarantee faithful explanation, this design provides strong auditability: the exact regions used for each prediction can be enumerated, reproduced, and rapidly reviewed. In contrast, standard attention heatmaps can be diffuse or unstable across folds. Deep learning systems are often perceived as "black boxes" with limited transparency in medical settings; in contrast, EAGLE allows rapid review of the exact regions used for representation construction. In our qualitative assessment, selected tiles frequently corresponded to plausible tumor rich regions and showed reduced artifact exposure[36]. Beyond improving interpretability, the resulting slide- and patient-level embeddings open new possibilities for content-based slide retrieval, clustering, and integration with other data modalities. By compressing the most relevant morphological cues into a single vector, EAGLE provides an efficient bridge to genomic, transcriptomic, and radiologic data. In the long term, such patient-level embeddings could serve as the backbone of comprehensive multimodal models, enabling more holistic characterizations of disease states. The field of CPath is expanding beyond image-only pipelines into vision language models or MLLMs that can annotate slides in a chatbot-like manner, but reveal significant limitations in specialized tasks. Despite referencing key diagnostic features, GPT-4o fails to achieve accuracy beyond chance levels in biomarker prediction, even when provided with EAGLE's selected tiles.

While MLLMs may complement traditional approaches in interpretability, their predictive capacity in pathology remains limited without domain adaptation.

Although we tested EAGLE extensively on multiple external cohorts spanning nine major cancer types, additional large-scale validation is warranted to assess its performance on rare pathologies. Encouragingly, recent studies have demonstrated that large-scale pathology foundation models can generalize well to rare and underrepresented cancer types, supporting the promise of such approaches for broader clinical use[7]. Furthermore, our approach might show some limitations for tasks which focus on very specific tissue areas (e.g. detection of vascular invasion) or in highly heterogeneous tumors (e.g. consisting of various different subtypes). CHIEF was primarily pre-trained on cancer slides, so its region selection may be less effective for non-cancer tasks, thereby impairing EAGLE's performance. A limitation of our interpretability analysis is that pathologist review was limited to a single expert and not all methods were assessed in parallel. Future studies should involve multiple pathologists and a standardized evaluation protocol. Our study prioritizes benchmarking predictive performance over detailed confounder analysis, as we focus on common predictive targets to ensure comparability across models. While this facilitates systematic evaluation, it does not capture the influence of confounding variables or the deeper morphological basis of predictions. As training datasets primarily originate from major cancer centers, they may not fully reflect the diversity of global patient populations, highlighting the need for broader validation across more heterogeneous cohorts. While EAGLE achieved AUROCs >0.900 for some tasks, the average AUROC of 0.742 indicates that EAGLE's performance might not yet be adequate to replace standard clinical procedures. Future work may explore prospective clinical trials to evaluate real-world diagnostic improvements, as well as the integration of EAGLE's embeddings with patient-specific data, such as laboratory values or electronic health records.

In conclusion, EAGLE demonstrates that combining complementary pathology foundation models can yield representations that are both efficient and broadly generalizable. By integrating large-scale pretrained knowledge through targeted region selection, EAGLE establishes an effective paradigm for building high-performing slide-level models without requiring additional large-scale data collection.

# Author contributions

PN, TL and JNK designed the study. PN and TL developed the software. PN, MG, RL, BD, LAB, AJF, ELG, AG, SB, MJG, RS, HMB, CR, TH, UP, AIP, GC, NF, AM, MH, HB and JNK contributed to data collection and assembly. PN, TL, SF, CMLL, JC, LAS and JNK interpreted and analyzed the data. All authors substantially contributed to writing and reviewing the report, approved the final version for submission, and have agreed to be personally accountable for the author's own contributions and to ensure that questions related to the accuracy or integrity of any part of the work, even ones in which the author was not personally involved, are appropriately investigated, resolved, and the resolution documented in the report.

## Acknowledgements

The authors gratefully acknowledge the GWK support for funding this project by providing computing time through the Center for Information Services and HPC (ZIH) at TU Dresden. The authors gratefully acknowledge the Gauss Centre for Supercomputing e.V. (www.gauss-centre.eu) for funding this project by providing computing time through the John von Neumann Institute for Computing (NIC) on the GCS Supercomputer JUWELS at Jülich Supercomputing Centre (JSC).

## Disclosures

JNK declares ongoing consulting services for AstraZeneca and Bioptimus. Furthermore, he holds shares in StratifAI, Synagen, and Spira Labs, has received institutional research grants from GSK and AstraZeneca, as well as honoraria from AstraZeneca, Bayer, Daiichi Sankyo, Eisai, Janssen, Merck, MSD, BMS, Roche, Pfizer, and Fresenius. SF has received honoraria from MSD and BMS. MG has received honoraria for lectures sponsored by Techniker Krankenkasse (TK) and AstraZeneca. RL declares consulting services and honoraria from MSD, Janssen, AstraZeneca, Astellas, Roche. UP declares consulting services for AbbVie and her husband is holding individual stocks for the following companies: BioNTech SE – ADR, Amazon, CureVac BV, NanoString Technologies, Google/Alphabet Inc Class C, NVIDIA Corp, Microsoft Corp. AM has received honoraria as a consultant, advisor or speaker from Roche, Lilly and Menarini/Stemline, and has received support for accommodation and travel from AstraZeneca, all outside the submitted work. NF: Consulting/advisory role: MSD, Merck, Novartis, AstraZeneca, Sysmex, Roche, Menarini Group, Gilead, Veracyte, Sakura, Abbvie. Speaker bureau: MSD, Novartis, AstraZeneca, Daiichi Sankyo, Sysmex, GSK, Gilead, Roche, Menarini, Leica Biosystems, ThermoFisher, Genomic Health, Veracyte, Lilly. Research grants: Novartis, Gilead, AstraZeneca, GSK, Pfizer. Travel grants: Roche, Novartis. No other conflicts of interest are declared by any of the authors.

## Funding

JNK is supported by the German Cancer Aid DKH (DECADE, 70115166), the German Federal Ministry of Research, Technology and Space BMFTR (PEARL, 01KD2104C; CAMINO, 01EO2101; TRANSFORM LIVER, 031L0312A; TANGERINE, 01KT2302 through ERA-NET Transcan; Come2Data, 16DKZ2044A; DEEP-HCC, 031L0315A; DECIPHER-M, 01KD2420A; NextBIG, 01ZU2402A; PROSURV, 01KD2509C), the German Research Foundation (DFG, Deutsche Forschungsgemeinschaft) as part of Germany's Excellence Strategy – EXC 2050/2 – Project ID 390696704 – Cluster of Excellence "Centre for Tactile Internet with Human-in-the-Loop" (CeTI) of Technische Universität Dresden, as well as through DFG-funded collaborative research projects (TRR 412/1, 535081457; SFB 1709/1 2025, 533056198), the German Academic Exchange Service DAAD (SECAI, 57616814), the German Federal Joint Committee G-BA (TransplantKI, 01VSF21048), the European Union EU's Horizon Europe research and innovation programme (ODELIA, 101057091; GENIAL, 101096312), the European Research Council ERC (NADIR, 101114631), the Breast Cancer Research Foundation (BELLA-DONNA, BCRF-25-225) and the National Institute for Health and Care Research NIHR (Leeds

Biomedical Research Centre, NIHR203331). The views expressed are those of the author(s) and not necessarily those of the NHS, the NIHR or the Department of Health and Social Care. This work was funded by the European Union. Views and opinions expressed are, however, those of the author(s) only and do not necessarily reflect those of the European Union. Neither the European Union nor the granting authority can be held responsible for them. SF is supported by the German Federal Ministry of Education and Research (SWAG, 01KD2215C), the German Cancer Aid (DECADE, 70115166 and TargHet, 70115995) and the German Research Foundation (504101714). The DACHS study was supported by the German Research Council (BR 1704/6-1, BR 1704/6-3, BR 1704/6-4, CH 117/1-1, HO 5117/2-1, HO 5117/2-2, HE 5998/2-1, HE 5998/2-2, KL 2354/3-1, KL 2354 3-2, RO 2270/8-1, RO 2270/8-2, BR 1704/17-1, BR 1704/17-2); the Interdisciplinary Research Program of the National Center for Tumor Diseases (NCT), Germany; and the German Federal Ministry of Education and Research (01KH0404, 01ER0814, 01ER0815, 01ER1505A, and 01ER1505B). JC is supported by the Mildred-Scheel-Postdoktorandenprogramm of the German Cancer Aid (grant #70115730). AM is supported by the European Society for Medical Oncology José Baselga Fellowship for Clinician Scientists founded by AstraZeneca (2023–2025). The Genetics and Epidemiology of Colorectal Cancer Consortium (GECCO) is funded by: National Cancer Institute, National Institutes of Health, U.S. Department of Health and Human Services (U01 CA137088, R01 CA488857, P20 CA252733, P50 CA285275). Genotyping/Sequencing services were provided by the Center for Inherited Disease Research (CIDR) contract number HHSN268201700006I. This research was funded in part through the NIH/NCI Cancer Center Support Grant P30 CA015704. Scientific Computing Infrastructure at Fred Hutch funded by ORIP grant S10OD028685· The CORSA study was funded by Austrian Research Funding Agency (FFG) BRIDGE (grant 829675, to Andrea Gsur), the "Herzfelder'sche Familienstiftung" (grant to Andrea Gsur) and was supported by COST Action BM1206· CRA was supported by the National Institutes of Health grant R01 CA068535. The coordination of EPIC is financially supported by the International Agency for Research on Cancer (IARC) and also by the Department of Epidemiology and Biostatistics, School of Public Health, Imperial College London which has additional infrastructure support provided by the NIHR Imperial Biomedical Research Centre (BRC). The national cohorts are supported by: Danish Cancer Society (Denmark); Ligue Contre le Cancer, Institut Gustave Roussy, Mutuelle Générale de l'Education Nationale, Institut National de la Santé et de la Recherche Médicale (INSERM) (France); German Cancer Aid, German Cancer Research Center (DKFZ), German Institute of Human Nutrition Potsdam- Rehbruecke (DIfE), Federal Ministry of Education and Research (BMBF) (Germany); Associazione Italiana per la Ricerca sul Cancro-AIRC-Italy, Compagnia di SanPaolo and National Research Council (Italy); Dutch Ministry of Public Health, Welfare and Sports (VWS), Netherlands Cancer Registry (NKR), LK Research Funds, Dutch Prevention Funds, Dutch ZON (Zorg Onderzoek Nederland), World Cancer Research Fund (WCRF), Statistics Netherlands (The Netherlands); Health Research Fund (FIS) - Instituto de Salud Carlos III (ISCIII), Regional Governments of Andalucía, Asturias, Basque Country, Murcia and Navarra, and the Catalan Institute of Oncology - ICO (Spain); Swedish Cancer Society, Swedish Research Council and and Region Skåne and Region Västerbotten (Sweden); Cancer Research UK (14136 to EPIC-Norfolk; C8221/A29017 to EPIC-Oxford), Medical Research Council (1000143 to EPIC-Norfolk; MR/M012190/1 to EPIC-Oxford) (United Kingdom). The IWHS study was supported by NIH grants CA107333 (R01 grant awarded to P.J. Limburg) and HHSN261201000032C (N01 contract awarded to the University of Iowa). The WHI program is funded by the National Heart, Lung, and Blood Institute, National Institutes of Health, U.S. Department of Health and Human Services through contracts 75N92021D00001, 75N92021D00002, 75N92021D00003, 75N92021D00004, 75N92021D00005. This work was partially supported by the Italian Ministry of Health through Ricerca Corrente 5 × 1000 funds; the Italian Ministry of Innovations via the Sustainable Growth Fund – Innovation Agreements under the Ministerial Decree of December 31, 2021, and the Director's Decree of November 14, 2022 (2nd Call), Project No.: F/350104/01–02/X60; and the Italian Ministry of University and Research (MUR) 2023 through the "Future Artificial Intelligence Research – FAIR" program, PE0000013, CUP D53C22002380006, within the National Recovery and Resilience

Plan (PNRR), Mission 4, Component 2, Investment 1.3 – funded by the European Union – NextGenerationEU. Project: "AIDH – Adaptive AI Methods for Digital Health."We kindly thank all individuals who agreed to participate in the CORSA study. Furthermore, we thank all cooperating physicians and students and the Biobank Graz of the Medical University of Graz. We also acknowledge the TCGA Research Network and the Clinical Proteomic Tumor Analysis Consortium (CPTAC), which generated the data on which some of the results shown in this study are based. Where authors are identified as personnel of the International Agency for Research on Cancer/World Health Organization, the authors alone are responsible for the views expressed in this article and they do not necessarily represent the decisions, policy or views of the International Agency for Research on Cancer/World Health Organization. The authors thank the WHI investigators and staff for their dedication, and the study participants for making the program possible. A full listing of WHI investigators can be found at:
https://s3-us-west-2.amazonaws.com/www-whi-org/wp-content/uploads/WHI-Investigator-Long-List.pdf

# Figures

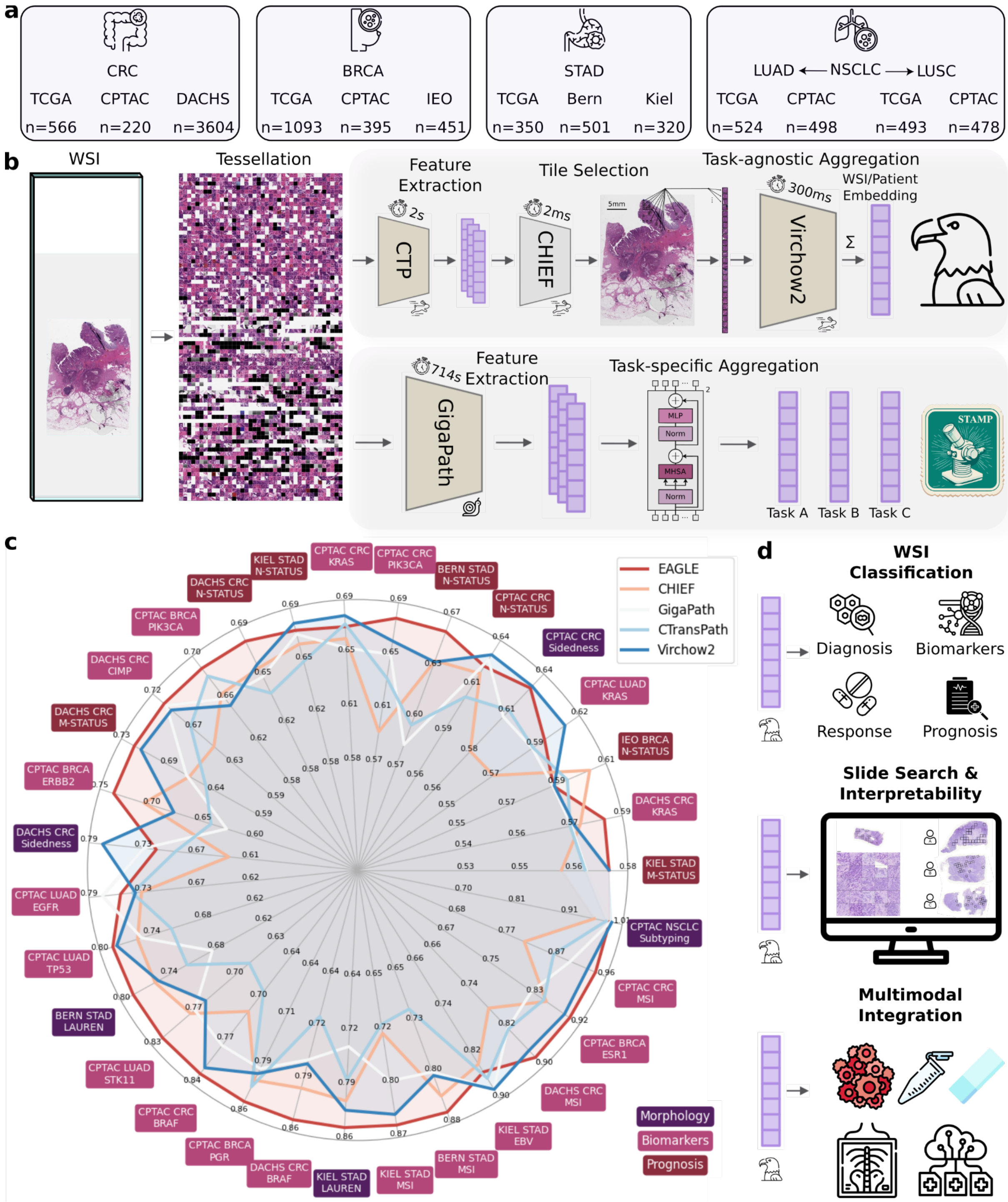

**Figure 1: EAGLE framework. a**) EAGLE is compared to state-of-the-art foundation models on 9,528 whole-slide images (WSIs) across 13 cohorts, encompassing four cancer types in the main benchmark. **b**) Workflow comparison of EAGLE and conventional supervised approaches. WSIs undergo tessellation into tiles, followed by feature extraction and aggregation. EAGLE employs a three-step process: tile feature extraction with CTransPath, tile selection

using CHIEF, and detailed feature extraction of 25 selected tiles using Virchow2, culminating in an averaged WSI embedding per patient. In contrast, supervised pipelines, such as STAMP, extract features from all tiles using models like Prov-GigaPath and aggregate these features using task-specific architectures (e.g., transformers). Measured time for average foundation model processing per WSI is included (EAGLE at 2 MPP vs. conventional supervised models at the commonly used 0.5 MPP). **c**) average AUROC scores comparing EAGLE against CHIEF, Prov-GigaPath, CTransPath, and Virchow2 across 31 computational pathology tasks. Each axis is normalized: the center equals 0.5, the outer edge the highest AUROC for that task. **d**) Applications of WSI/patient embeddings, including WSI classification for diagnosis, biomarker analysis, treatment response prediction, and prognosis; slide retrieval for similar-case search in clinical workflows; and integration into AI-powered clinical systems and multi-omics modeling.

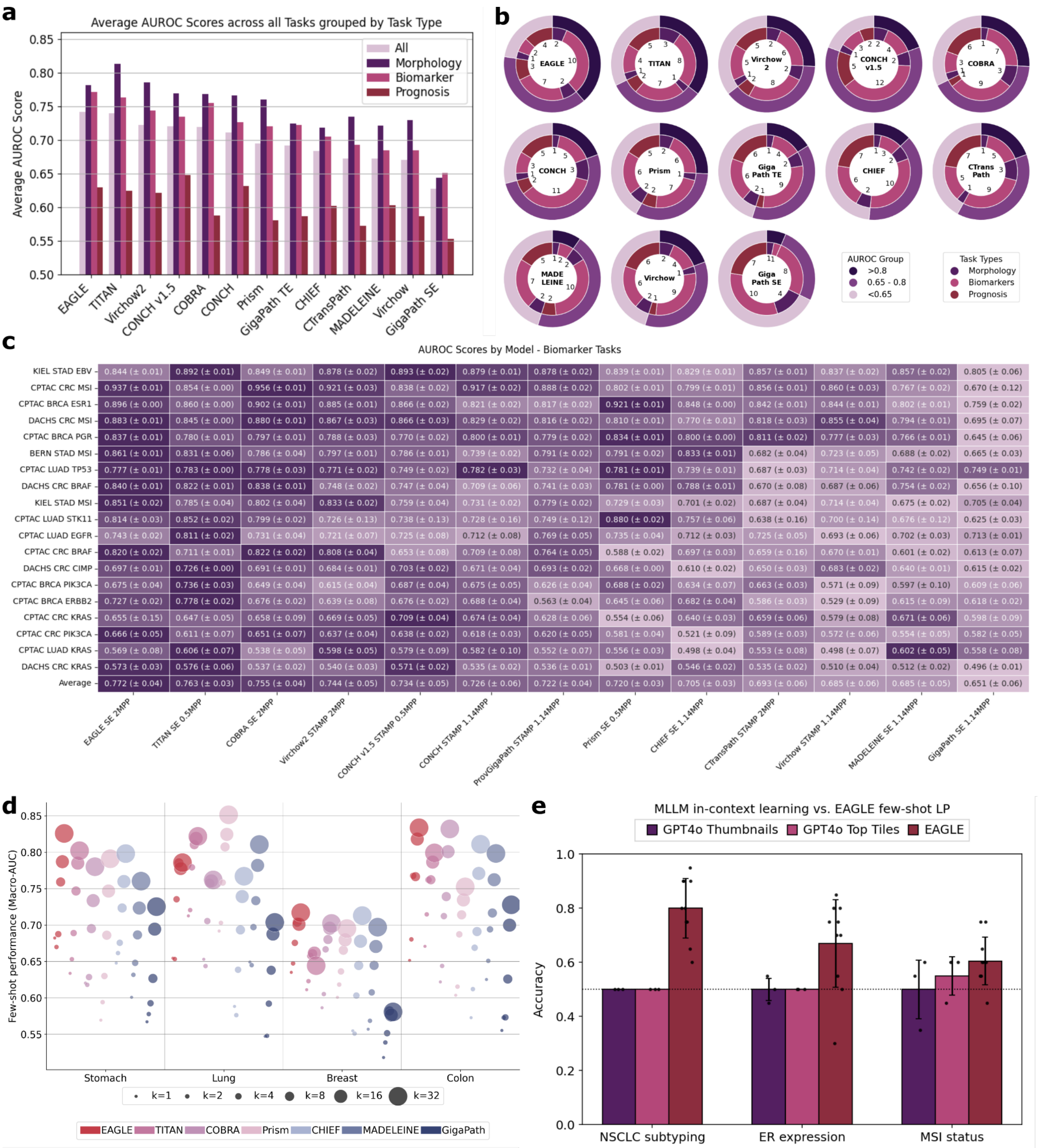

**Figure 2: Comparative performance of EAGLE vs. tile- and slide-level foundation models. a**) Average AUROC scores per task category across five folds for 13 models evaluated on 31 tasks. **b**) Distribution of tasks by model performance, showing the number of tasks where each model achieves an average AUROC of >0.800, 0.650 - 0.800, or <0.650, grouped by task type. **c**) Average AUROC scores for 19 biomarker prediction tasks (mean ± SD across five folds). Taskwise color normalization for better comparison of the models. Tasks are sorted by their mean AUROC across all models, while models are sorted by their mean AUROC across all tasks. **d**) Few-shot performance of slide encoders using linear probing with logistic regression for k=1, 2, 4, 8, 16, and 32 samples per class. To ensure robust evaluation, only the top three binary tasks per cancer type (selected based on highest AUROC across all models) are included. **e**) Comparison of EAGLE's few-shot linear probing (using k=2 examples per class) versus GPT-4o's in-context learning (k=2) for ER expression prediction in BRCA, MSI status in CRC, and NSCLC subtyping.

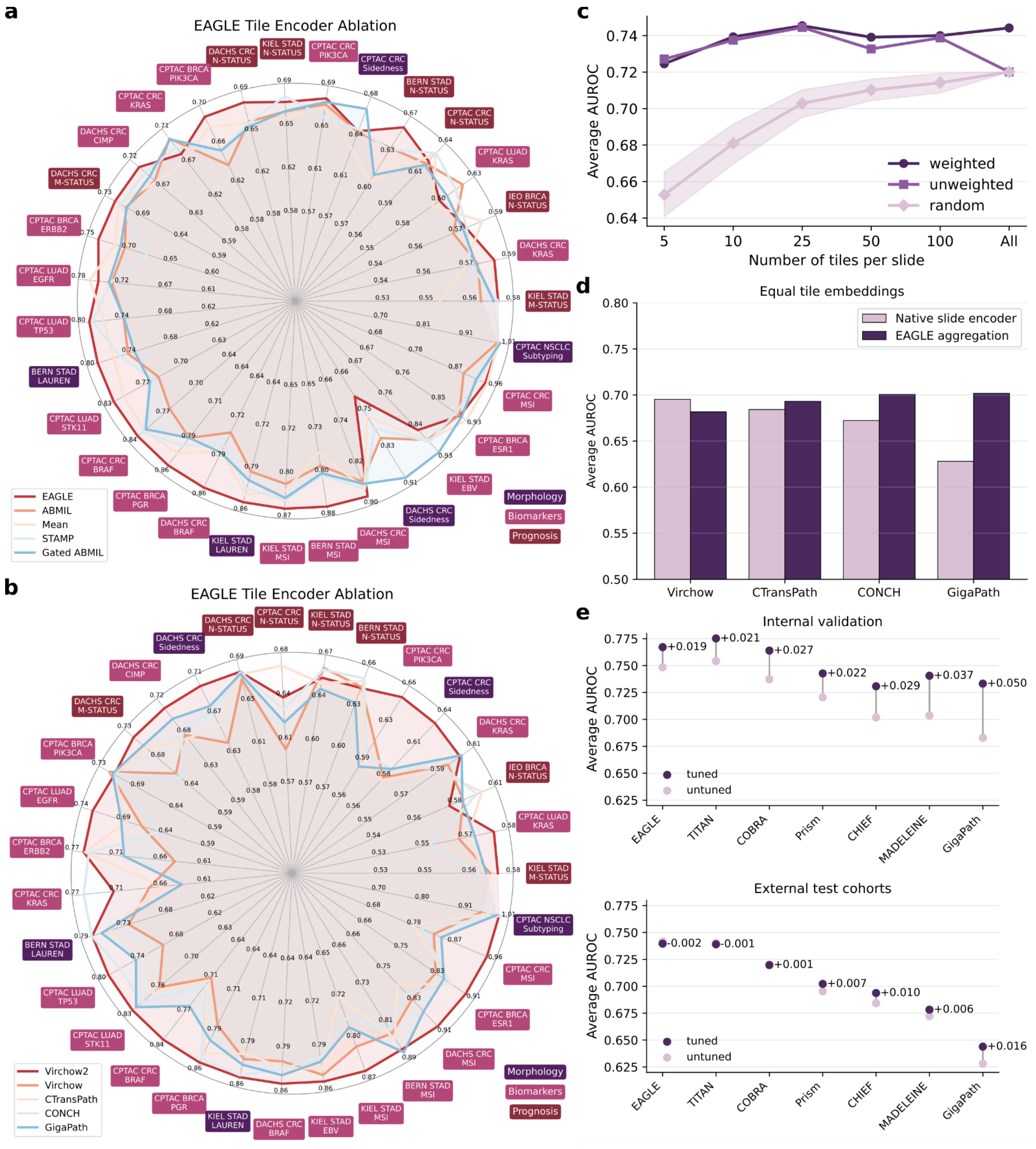

**Figure 3: Ablation studies to dissect the design choices of EAGLE. a**) Comparison of alternative aggregation strategies applied to Virchow2 tile embeddings at 2 MPP across 31 computational pathology tasks. Averaging (Mean), attention-based multiple instance learning (ABMIL), gated ABMIL, STAMP, and EAGLE were evaluated to quantify the impact of the aggregation mechanism on model performance. **b**) Comparison of five tile encoders, Virchow2, Virchow, CTransPath, CONCH, and GigaPath, used within the EAGLE framework across 31 tasks, isolating the influence of the underlying tile encoder on predictive accuracy. **c**) Effect of the number and selection strategy of tiles per patient embedding. CHIEF-based tile selection was evaluated for N = 5, 10, 25, 50, and 100 tiles per slide using either unweighted averaging or attention-weighted aggregation. Uniform random tile selection was repeated 100 times for each N to estimate an empirical null distribution (mean and 95% interval shown). Across all tile counts, CHIEF-based selection consistently outperformed the full range of random replicates (Monte Carlo P = 1/101 for each N). **d**) Comparison of native slide encoders versus EAGLE as a universal aggregator for four tile encoders. Bars show the mean AUROC across all 31 tasks for the native slide encoders and for EAGLE built on the same tile embeddings. **e**) Average change in AUROC on 23 internal and 31 external validation tasks when comparing tuned and untuned multilayer perceptron configurations. Each point is a model task pair. Tuning is performed once per model on internal data only, then evaluated out of sample on external cohorts.

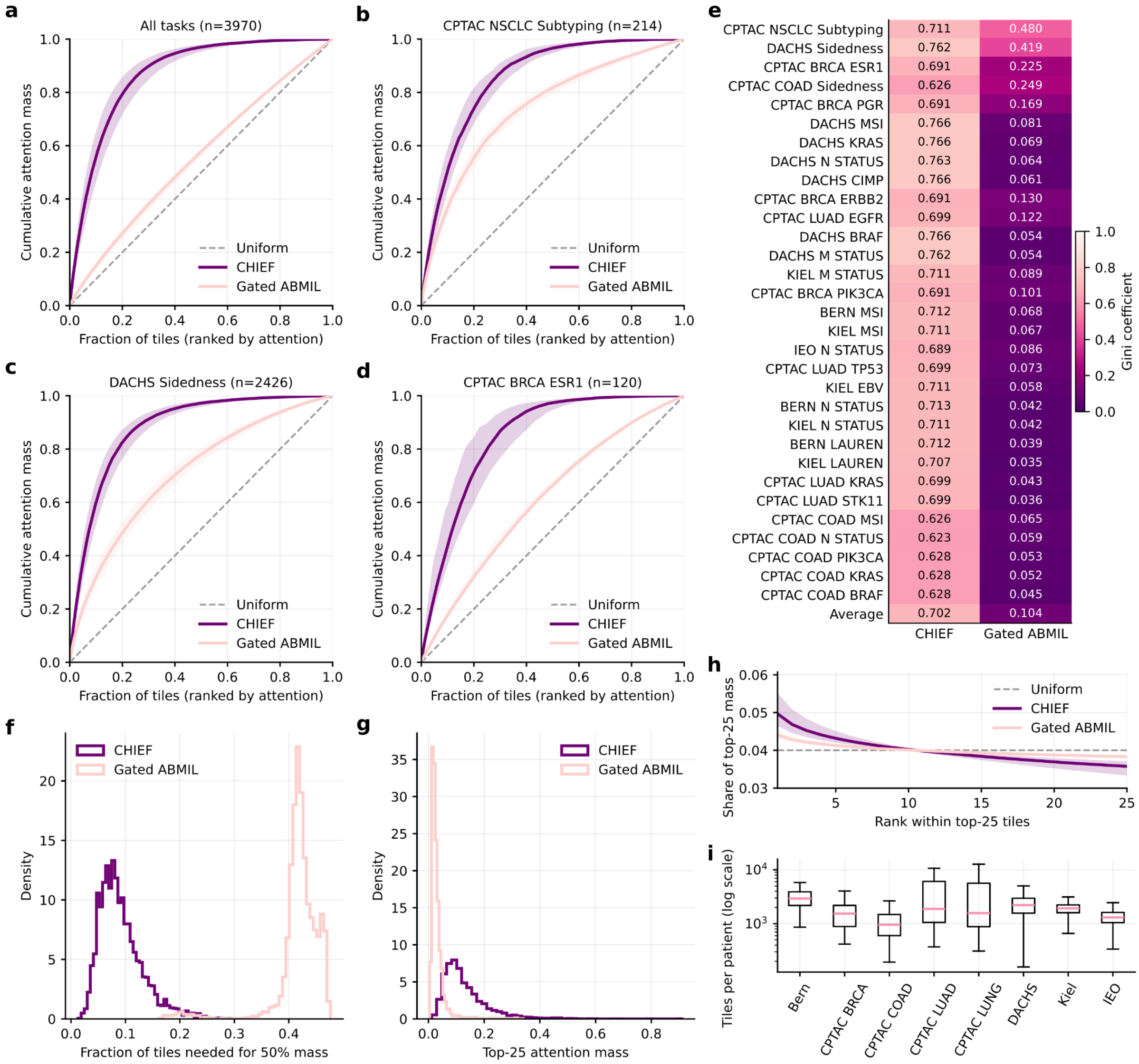

**Figure 4: Quantitative attention concentration analyses. a**) Lorenz curves of cumulative attention mass as a function of the fraction of tiles ranked by attention, aggregated across all tasks with each patient counted once by averaging rank profiles rather than tile identities. Curves are shown for uniform attention, CHIEF, and gated ABMIL. **b–d**) Representative task-specific Lorenz curves for CPTAC NSCLC subtyping, DACHS sidedness, and CPTAC BRCA ESR1. **e**) Gini coefficient heatmap per task and model, including an average row. **f**) Distribution of the fraction of tiles required to accumulate 50% of total attention mass, computed per patient and shown as density curves. **g**) Distribution of top 25 cumulative attention mass per patient. **h**) Contribution profile of the top 25 ranks, showing the share of top 25 mass contributed by each rank position. **i**) Tile count distributions for external test cohorts on a log scale, illustrating what 25 selected tiles represent as a fraction of available tissue area.

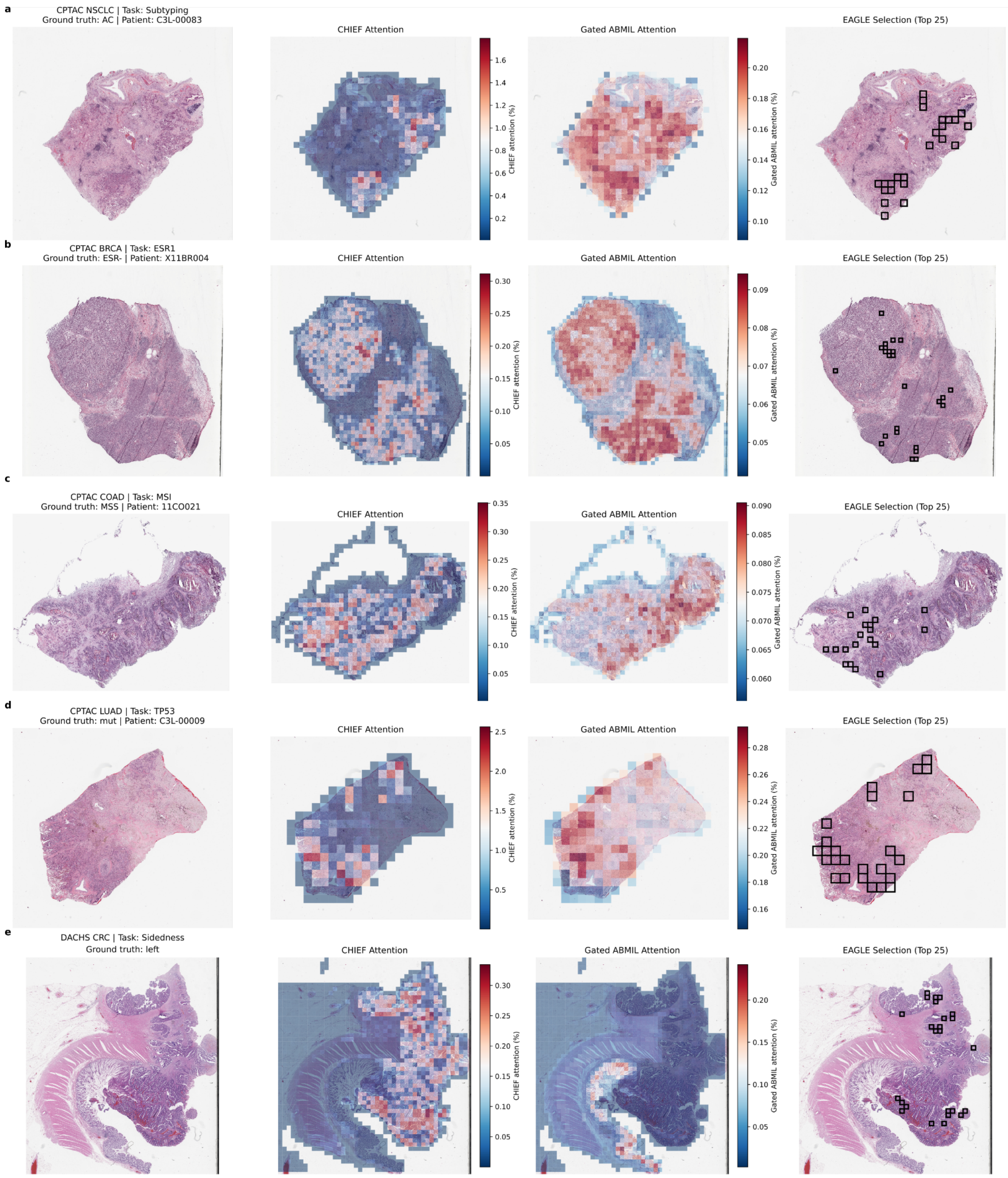

**Figure 5: Direct visualization of attention maps and EAGLE top 25 overlays.** For each representative whole-slide image, columns show (left to right) a low-resolution H&E thumbnail, the CHIEF attention heatmap (computed from frozen CHIEF operating on CTransPath embeddings), the gated attention-based multiple instance learning (gated ABMIL) heatmap (task-specific model trained on Virchow2 embeddings), and the EAGLE tile set visualized as black boxes marking the exact top-25 tiles selected by CHIEF and used to construct the EAGLE representation. Panels show representative external test cases for non-small cell lung cancer subtyping (**a**), ESR1 mutation prediction in CPTAC BRCA (**b**), microsatellite instability prediction in CPTAC COAD (**c**), TP53

mutation prediction in CPTAC LUAD (**d**), and colorectal cancer sidedness in DACHS (**e**). Heatmap intensities are shown after softmax normalization within each slide. Color scales are normalized per model and per panel to visualize each method's dynamic range. Matched-scale controls and additional examples are provided in **Supplementary Fig. 6**, and corresponding visualizations using standard ABMIL are provided in **Supplementary Fig. 7**.

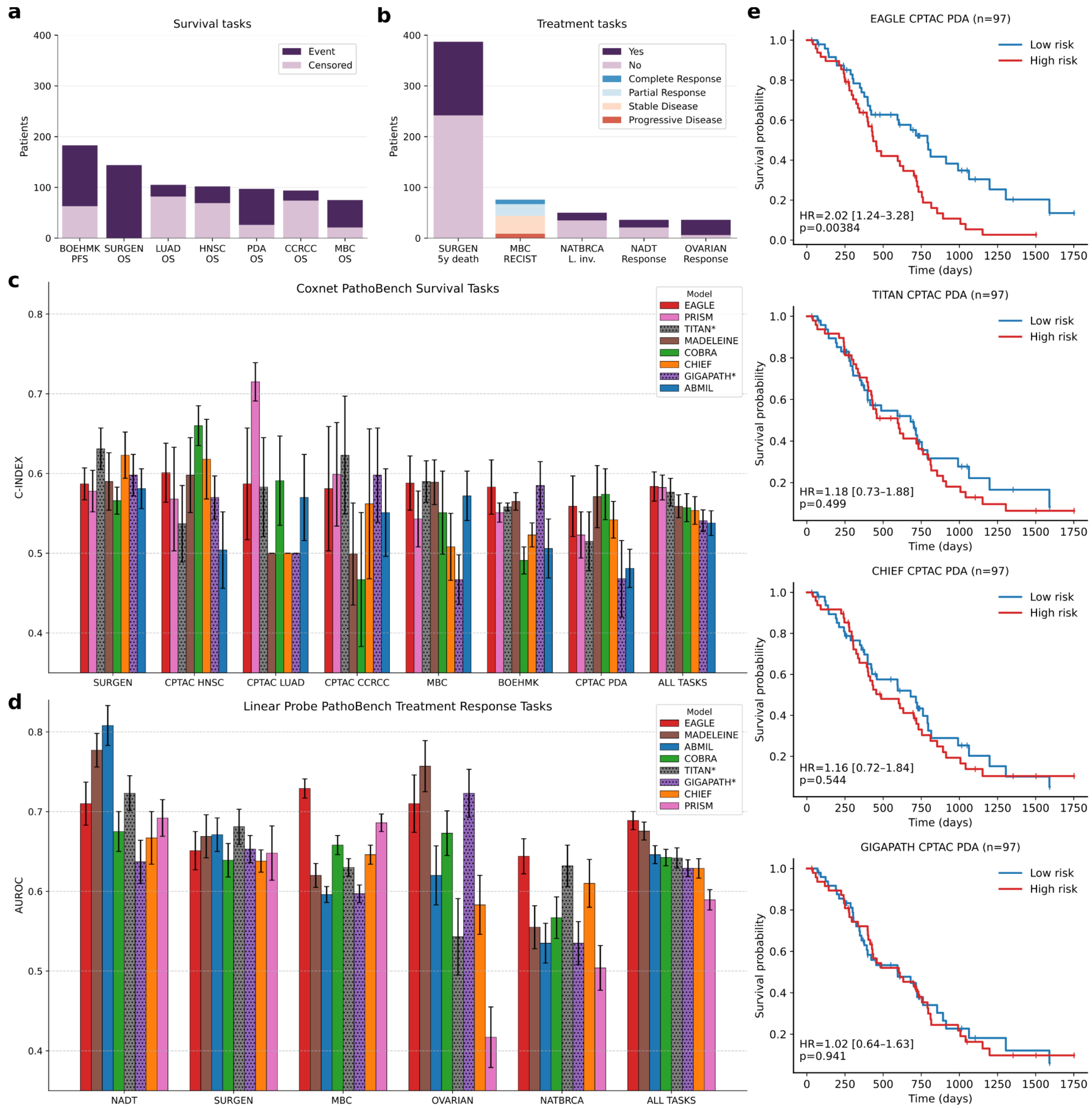

**Figure 6: Extended Evaluation of EAGLE on PathoBench Tasks. a**) Overview of seven survival prediction tasks included in the PathoBench extension, comprising six overall survival (OS) and one progression-free survival (PFS) endpoint. For each task, the number of events (deaths or progressions) and censored cases is shown. **b**) Overview of five treatment response prediction tasks, detailing the number of patients per response category across endpoints such as five-year mortality (death vs. survival), RECIST response classes (complete response, partial response, stable disease, progressive disease), lymphovascular invasion, and binary treatment response (yes or no). **c**) Concordance index (C-index) performance for seven survival tasks using the CoxNet implementation provided in PathoBench, comparing eight models. GigaPath and TITAN require spatial coordinates, and since PathoBench does not support multi-slide aggregation per patient, their patient-level embeddings were obtained by averaging individual slide embeddings. **d**) Mean AUROC performance across the five treatment response tasks, comparing the same eight

models under five-fold or 50-fold Monte Carlo cross-validation. Error bars represent confidence intervals as defined by the PathoBench evaluation protocol. **e**) Kaplan–Meier survival analysis for EAGLE, TITAN, CHIEF, and GigaPath on the CPTAC pancreatic ductal adenocarcinoma (PDA) cohort (n = 97). Patients were stratified by the median EAGLE risk score into low- and high-risk groups, and the hazard ratio and p-value were computed using a two-sided log-rank test.

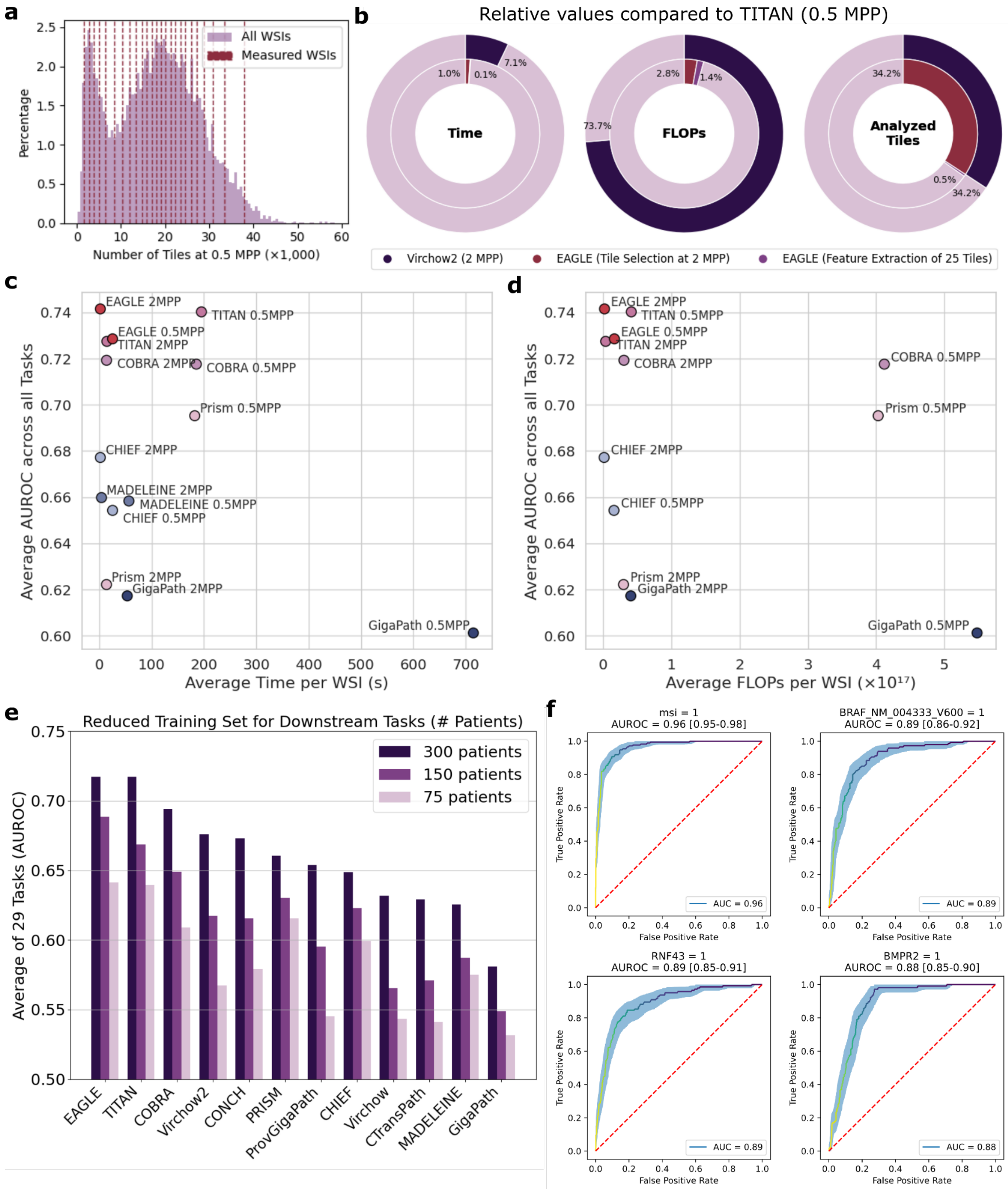

**Figure 7: Computational efficiency and performance in scarce data scenarios. a**) Tile distribution across 9,528 whole-slide images (WSIs) at 0.5 MPP, with 25 WSIs sampled at regular intervals from the 2nd to the 98th percentile for timing experiments. These WSIs (mean: 17,771 tiles) differ by less than 0.3% from the dataset-wide average (17,819 tiles). **b**) Time, floating point operations (FLOPs), and number of analyzed tiles for processing 25 WSIs with EAGLE and Virchow2 at 2 MPP, normalized to TITAN at 0.5 MPP. EAGLE metrics are split into the tile selection step consisting of CTransPath and CHIEF inference and the feature extraction of selected tiles

using Virchow2. These models and magnifications were chosen as they achieve the highest average AUROC scores in their respective optimal settings. **c**, **d**) Time (**c**) and FLOPs (**d**) comparisons across models, plotted against AUROC performance across 31 tasks. FLOPs include tile extraction; time includes both tile and slide processing. **e**) AUROC performance of seven slide encoders and their tile-based counterparts across 29 tasks, for classifiers trained with 300, 150 or 75 patients. **f**) AUROC curves for four biomarkers with high predictive performance (MSI status, *BRAF V600*, *RNF43*, and *BMPR2* mutations) from a dataset of over 1,000 patients across multiple centers. All biomarkers with an observed AUROC > 0.800 are listed in **Supplementary Table 4**.

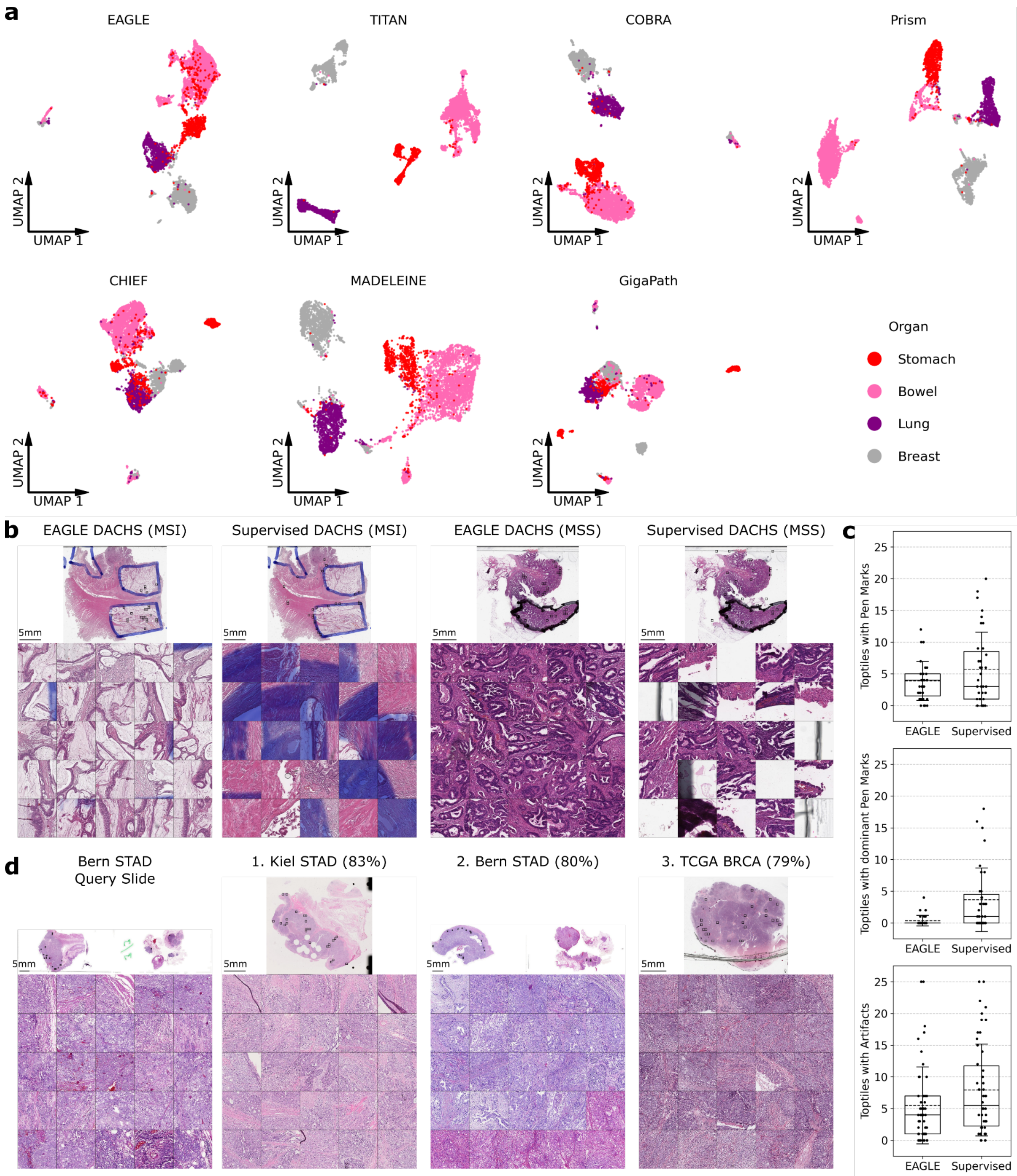

**Figure 8: Versatility and interpretability. a**) UMAPs of slide embeddings generated by EAGLE and tested slide encoders. Four tissue types are color-coded, illustrating separation by tissue type across models. **b**) Comparison of the top 25 tiles selected by EAGLE versus the supervised baseline on MSI status prediction (Virchow2 embeddings aggregated via the STAMP pipeline) for two representative DACHS slides (one MSI and one MSS case). The supervised baseline frequently highlights background, slide edge, pen mark, and other artifacts, whereas EAGLE predominantly selects representative, tissue-relevant regions. **c**) Prevalence of artifacts among top tiles from 50 randomly chosen DACHS WSIs. For each slide, the number of top tiles containing pen marks,

dominant pen marks (covering ≥50% of the tile), or other artifacts (e.g. tissue folds, slide edges, air bubbles, oil drops, scratches, foreign objects, dark spots, and out-of-focus regions) are shown for EAGLE and the supervised baseline. The median is shown as a solid center line, and the mean with standard deviation is represented by a dashed line and error bars. **d**) Examples of slide retrieval using EAGLE's embedding space. For a given query slide, the nearest neighboring slides across all cohorts are displayed, demonstrating EAGLE's potential for similar-case search in clinical workflows. The percentages indicate the cosine similarity between the query patient embedding from the Bern cohort and the three closest patient embeddings in the feature space.

# Materials and Methods

## Ethics statement

This study was conducted in accordance with the Declaration of Helsinki. Both TCGA and the Clinical Proteomic Tumor Analysis Consortium (CPTAC) collections comprise retrospective, anonymized data, so no additional ethics approval was required. The DACHS study is an epidemiological study overseen by the German Cancer Research Center (DKFZ, Heidelberg, Germany) and was approved by the ethics committee of the Medical Faculty of the University of Heidelberg (reference 310/2001)[37–39].

## Datasets

This study benchmarks prediction tasks spanning nine major cancer types—breast, colorectal, stomach, lung, prostate, ovary, head and neck, kidney and pancreas—across a range of clinically relevant endpoints, including tumor morphology, molecular biomarker status, treatment response and prognosis. All histopathology data consists of hematoxylin and eosin (H&E)-stained WSI, including both formalin-fixed paraffin-embedded (FFPE) and fresh frozen tissue. Cohorts span a range of staining protocols and scanner hardware, representing the variability encountered in real-world clinical practice. Our evaluation leverages large, multi-institutional datasets for both training and independent external validation, ensuring robust and generalizable performance assessment. High-level summaries of all training and validation datasets, prediction endpoints, and clinical variables are provided in **Supplementary Tables 5–7** for reference. All models used in this study, except where explicitly noted, were trained exclusively on TCGA whole-slide images (WSIs), which included histopathology data from lung adenocarcinoma (LUAD), lung squamous cell carcinoma (LUSC), colorectal cancer (CRC), stomach adenocarcinoma (STAD), and breast cancer (BRCA). External validation was carried out on CPTAC (including CPTAC-2 and CPTAC-3 prospective cohorts from 2018 and 2020, covering LUAD, LUSC, COAD, and BRCA), DACHS (CRC), proprietary STAD datasets (Kiel and Bern), and a BRCA dataset (IEO) (**Extended Data Fig. 1c**). None of these external cohorts were used to pretrain the foundational models or train the downstream classifiers, ensuring that no data leakage occurred. Additional information about the cohorts, including patient characteristics such as age, sex, race, cancer stage, and tumor stage, is provided in **Supplementary Table 5**.

For the PathoBench dataset, we evaluated 12 prediction tasks encompassing survival and treatment response endpoints across multiple cancer types, including breast, colorectal, ovarian, prostate, head and neck, kidney, and pancreatic cancers. The seven survival tasks comprise one progression-free survival endpoint (BOEHMK, ovarian) and six overall survival endpoints (SURGEN, colorectal; CPTAC-LUAD, lung; CPTAC-HNSC, head and neck; CPTAC-PDA, pancreas; CPTAC-CCRCC, kidney; and MBC, metastatic breast). The five treatment-response tasks included five-year mortality (SURGEN, colorectal), RECIST response (MBC, breast), lymphovascular invasion (Post-NAT-BRCA, breast), hormonal therapy response (NADT-Prostate, prostate), and bevacizumab response (OV-Bevacizumab, ovarian). In contrast to the main benchmarking framework, PathoBench does not include external validation cohorts; all results were obtained using the predefined internal cross-validation splits provided with the dataset. All histopathology data consisted of H&E-stained slides from multiple institutions, following the standardized preprocessing and evaluation procedures defined in the PathoBench protocol[31,32].
For biomarker discovery experiments (results presented in **Fig. 7f, Supplementary Table 4**), an additional dataset from the Genetics and Epidemiology of CRC Consortium (GECCO) was employed, comprising five studies: CORSA, EPIC, CRA, WHI, and IWHS[40,41] (**Extended Data Fig. 1d, Supplementary Table 6**). To ensure a minimum of 10 cases per class for both training and testing while maximizing the inclusion of biomarkers, different training and testing splits were used for each biomarker. Three centers were selected for training and two for testing, with the specific splits detailed in **Supplementary Table 4** for biomarkers achieving AUROC > 0.800.

## Experimental Design

Our study emulates a recently established large-scale benchmarking study[22]. To reflect a broad range of tasks relevant in digital pathology, three main categories were defined, namely morphological, biomarker, and prognostic tasks. Morphological tasks involved the classification of CRC slides as left or right colon (excluding transverse colon), STAD slides into Lauren subtypes, and NSCLC slides into adenocarcinoma (LUAD) or squamous cell carcinoma (LUSC). Biomarker tasks targeted molecular or expression features such as *BRAF*, *KRAS*, *PIK3CA* mutation, microsatellite instability (MSI) and CpG island methylator phenotype (CIMP) status in CRC, EBV and MSI status in STAD, *EGFR*, *STK11*, *KRAS* and *TP53* mutation in LUAD, and HER2, ER, PR expression, and *PIK3CA* mutation in BRCA. Prognostic tasks addressed nodal involvement (N vs. N0) and metastasis (M0 vs. M+). We included only tasks with at least ten available cases per label in both training and test cohorts. In total, 31 tasks were used, and all were binary except Lauren classification (**Supplementary Table 7**).

The classifiers were trained and evaluated using a five-fold split of the TCGA data: in each fold, 80% of the data was used for training and 20% for validation (to perform early stopping). This results in five independently trained models per task, each trained on a different partition of the training sets. Each of these five models was then applied to the external test cohorts (CPTAC, DACHS, Kiel, Bern, IEO) and the results were averaged to provide a robust estimate of performance while minimizing single-run bias. This improves statistical validity and ensures external validation by testing exclusively on datasets not used during training. Performance metrics included AUROC, AUPRC, Balanced Accuracy, and F1 Scores and always represent the mean across the five folds.

We followed the PathoBench protocol and used the predefined eighty percent train and twenty percent test splits. Within each training split, we defined a validation subset comprising twenty percent of cases at the case level. For categorical endpoints we stratified the validation selection when feasible. The test portion was never used for model selection. PathoBench supports early fusion of multiple slides per patient for models that do not rely on spatial coordinates. Consequently, TITAN and GigaPath were evaluated by pooling each slide separately and averaging slide embeddings to obtain a patient representation, whereas all other models used early fusion to form a single patient-level bag prior to pooling. For survival endpoints we trained CoxNet on patient-level embeddings and selected hyperparameters on the validation subset. We searched alpha values of 0.01, 0.05, and 0.1 and L1 mixing ratios of 0.1, 0.5, and 0.9, choosing the configuration that maximized the validation concordance index before evaluating on the held out test set. Survival tasks used five-fold cross validation. For treatment response endpoints we used a linear probing classifier with a single regularization parameter. We searched C equal to 0.001, 0.01, 0.1, 0.5, 1.0, and 10, selecting the value that maximized validation AUROC before evaluating on the test set. The fold schedule followed PathoBench: SURGEN 5y death used five-fold cross validation, whereas MBC RECIST, NATBRCA L. inv., NADT response, and OVARIAN Response used fifty fold Monte Carlo cross validation. Confidence intervals were reported as defined by the PathoBench protocol. To ensure comparability across methods, all models used the same predefined splits, the same train and validation construction, and the same early versus late fusion setting as described above. PathoBench contains no external cohorts, so all PathoBench results reflect internal cross validation.

## Image Processing and Deep Learning Techniques

We adopted the STAMP v1.1.1 pipeline (see Code availability) for WSI segmentation, tile extraction, and subsequent feature generation. STAMP (Solid Tumor Associative Modeling in Pathology) is an open-source protocol and software for deep learning-based biomarker prediction from whole-slide images, enabling end-to-end modeling and integration of clinical and tabular data in CPath workflows[18]. WSIs were divided into patches (tiles) of size 224×224 pixels (or 512×512 pixels for CONCH v1.5/TITAN, downsampled to 448×448 pixels before model inference). The final effective resolution could be 0.5, ~1.14, or 2 MPP, depending on the foundation model's magnification preference. For CONCH v1.5, due to downsampling, the effective resolutions were ~0.57 and ~2.28 MPP. Each combination of model and resolution was evaluated, with the best magnification reported in the main experiments. Tiles predominantly containing background were excluded using

Canny edge detection (thresholds: 40, 100), rejecting those with fewer than 2% of pixels classified as edges[42]. Before model input, all images were normalized according to the requirements of each foundation model: for CHIEF and Virchow2, standard ImageNet mean and standard deviation values ([0.485, 0.456, 0.406] and [0.229, 0.224, 0.225]) were used, while other models used their respective recommended normalization statistics. No dedicated stain normalization was applied. Remaining tiles were passed into a tile-level foundation model to generate feature embeddings, ranging in dimensionality from 512 (CONCH) to 1536 (Prov-GigaPath), forming an N×M matrix of tile features per WSI, where N is the number of tiles and M is the feature dimensionality.

Two main approaches were taken to obtain slide-level representations. In the supervised approach, tile embeddings were fed into a task-specific aggregator, typically a transformer-based network (STAMP) or attention-based multiple instance learning (ABMIL), in order to produce supervised predictions for each classification endpoint[18,21]. In this scenario, every new task required retraining the aggregator model. We evaluated two predefined attention-based MIL baselines: standard ABMIL[21] and a gated attention variant[21]. For both, hyperparameters were fixed a priori and applied uniformly across all tasks without task-specific tuning. The standard ABMIL configuration followed a commonly used setting with single-head attention and bag size 512. The gated ABMIL baseline was configured to avoid potential undercapacity effects by using full bags (no tile subsampling) and multiplicative gating in the attention pathway. Detailed configurations are provided in **Supplementary Table 8**. Recently, slide encoders were introduced, which combine tile embeddings (and sometimes tile coordinates) into a single embedding vector for each slide or patient. This strategy is agnostic to the specific task and thus enables a unified, unsupervised representation that can be readily extended to multiple downstream applications by training a small classifier. Detailed hyperparameters for all classifiers are listed in **Supplementary Table 8**.

EAGLE (Efficient Approach for Guided Local Examination) is a two-stage framework that uses the pre-trained, frozen, task-agnostic ABMIL model CHIEF to select the most informative regions within each WSI, followed by detailed feature extraction from these regions using the state-of-the-art tile encoder Virchow2. CHIEF was pre-trained on over 60,000 whole-slide images spanning 19 anatomical sites. Its pretraining combines tile-level SSL with weakly supervised slide-level contrastive learning and anatomical site encoding, enabling robust region selection across diverse cancer types without the need for retraining. Virchow2 is a state-of-the-art vision transformer trained on over 3 million WSIs from diverse tissues and institutions. The EAGLE framework therefore combines a slide-level foundation model with a tile-level foundation model in a unique way to efficiently generate precise WSI embeddings. EAGLE emulates the pathologist's workflow by first obtaining an overview of the entire slide, then computationally focusing on the most relevant regions by processing only these with a larger, higher-capacity tile encoder. Since EAGLE requires CTransPath tile embeddings as input, we first extracted features using CTransPath at 2 MPP.

CHIEF then uses these embeddings to produce a slide-level representation and an attention vector, originally designed to aggregate tile embeddings into a single slide embedding. EAGLE repurposes this attention vector to identify the top 25 most informative tiles, mimicking how pathologists zoom in on pertinent areas. These tiles are reprocessed with Virchow2 to extract detailed feature embeddings. The resulting 25 embeddings are averaged (i.e., each tile contributes equally) to create a compact, unsupervised slide-level representation. We chose equal weighting to avoid the risk of single-tile dominance and for conceptual simplicity. This hybrid approach minimizes computational costs by applying a more powerful feature extractor only to the most informative tiles, preserving critical morphological details while avoiding unnecessary processing of less relevant regions. The final embeddings enable the use of lightweight models, such as a small MLP classifier, for downstream tasks. This efficient pipeline demonstrates a balance of performance and computational feasibility for large-scale pathology analyses. The EAGLE framework supports replacement of both the tile and slide encoders. We tested substituting Virchow2 with other tile encoders in ablation studies (see **Fig. 3b**). For the slide encoder, any model that provides tile-level relevance or attention scores could be used in place of CHIEF. Models without explicit tile-level scores are not compatible with this approach. We chose CHIEF specifically for its strong performance and computational efficiency, aligning with the efficiency goals of EAGLE. Negative control experiments for region selection were performed using uniform random tile selection without replacement from the same tile set available to CHIEF for each patient. For each tile budget N in [5, 10, 25, 50, 100], we generated R = 100 independent random selections per fold and evaluated each replicate under the identical training and testing protocol used for the corresponding CHIEF-selected setting. For each replicate we report the mean AUROC aggregated across all tasks and folds, yielding an empirical null distribution for unguided subsampling.

Multiple other slide encoders and their tile-level foundations were also assessed (**Supplementary Tables 9,10**). Prov-GigaPath, used a masked autoencoder scheme on 171,189 WSIs from Providence, adopting a LongNet architecture with dilated attention at the slide level and a ViT-G/14 tile encoder trained with DINOv2[9,43]. PRISM applied a Perceiver-based architecture with CoCa-style vision-language alignment, trained on 587,196 WSIs in 195,344 specimen-report pairs, and employed a ViT-H/14 tile encoder called Virchow, which was also pretrained with DINOv2[10,26]. CHIEF was trained via slide-level contrastive learning and anatomic site information, using CTransPath with a SwinTransformer architecture as the tile encoder[8,23]. MADELEINE was pretrained on multi-stain data from breast samples using a dual global/local cross-stain alignment, and it built on CONCH, a vision-language CoCa model with 1.1 million image-text pairs[27,28]. TITAN used a multi-stage pretraining regime combining visual self-supervision, alignment with pathology reports, and 423,122 synthetic captions generated by a multimodal AI copilot; it employed CONCH v1.5 for tile embeddings, relying on UNI as its vision backbone and CoCa's text tower[7,30,44]. COBRA was trained on 3,048 WSIs to align tile embeddings from multiple foundation models via a contrastive loss, yielding a slide-level representation[29]. The best version of COBRA uses Virchow2 features, which expands on Virchow by scaling its dataset to 3.1 million WSIs and using domain-specific modifications of the DINOv2 framework[24]. COBRA and CHIEF were (partly) trained on TCGA, but none of the models used our external testing cohorts, preventing data leakage. We compared each slide encoder against a straightforward mean pooling baseline, where tile embeddings are simply averaged to yield an unsupervised slide-level feature vector. For each slide encoder and for STAMP, we evaluated multiple magnifications (0.5 MPP, 1.14 MPP, 2 MPP) and selected the resolution yielding the best downstream performance for each model and task. For PathoBench tasks, we used the magnification recommended by the original authors. Across all models, performance differences between magnifications were generally minor and no single resolution was consistently optimal.

For certain patients with multiple WSIs (e.g., multiple tissue blocks), each slide encoder was applied in two possible ways. Either each slide was independently processed, and the embeddings were averaged to yield a single patient representation, or all tiles were passed simultaneously into the slide encoder to produce one patient embedding per forward pass. Combining all tiles lead to higher performance of all slide encoders, but due to memory constraints on a 48 GB GPU we had to cap the maximum number of features for TITAN at 15,000.

Once a slide-level or patient-level embedding was computed, it was fed into a small multilayer perceptron (MLP) to yield final predictions for each classification task. This MLP used an input size of 768, had a hidden size of 256 with SiLU activation and dropout, and was trained for 32 epochs using a one-cycle policy with the AdamW optimizer (learning rate=$1\times10^{-4}$, weight decay=$1\times10^{-2}$). Cross-entropy loss with class weighting addressed label imbalance. Early stopping, monitored through validation loss in a five-fold cross-validation scheme, selected the best checkpoints. For few-shot linear probing experiments, we replaced the MLP with a logistic regression model (lbfgs solver, L2 penalty=1.0, 10,000 maximum iterations, and balanced class weights) and trained on k=1, 2, 4, 8, 16, 32 samples per class (**Extended Data Fig. 1a,b**, **Supplementary Table 6**). We repeated each setting ten times with different random draws to stabilize the results. This approach allowed us to evaluate how effectively slide encoders adapted to extremely limited training data.

To compare EAGLE with a multimodal large language model, we tested EAGLE's few-shot performance (k=2) against GPT-4o in-context learning on three specific tasks: NSCLC subtyping (LUAD vs. LUSC), MSI status prediction in CRC, and ER expression in BRCA. EAGLE used its standard logistic regression approach, while GPT-4o received prompts containing two example images per class, plus the query image (top 25 EAGLE-selected tiles vs. a single thumbnail) (**Supplementary Table 11**). The GPT-4o model was run three times per image to reduce variability, with temperature=0.7 and a maximum token length of 1,000. A strict JSON format in the prompt forced GPT-4o to provide a single label in each response.

An additional analysis was conducted to examine computational efficiency. We selected 25 representative WSIs (at percentiles 2, 6, 10, …, 98 by tile count) and measured the inference time and floating-point operations (FLOPs) for each tile encoder. FLOP counts were derived using ptflops and were multiplied by the average tile count in the overall dataset to estimate total computational costs.

## Interpretability

To interpret the embedding spaces generated by EAGLE, we applied UMAP dimensionality reduction (n_neighbors=15, min_dist=0.1) to project patient embeddings from all cohorts into two dimensions. This visualization enabled the assessment of morphological clustering by tissue origin. Separate experiments tested how EAGLE's embeddings facilitated slide-level retrieval by normalizing each embedding with L2 and performing a cosine similarity search. We queried five random patients per external cohort and returned the top three matches. The retrieved slides were assessed by a board-certified pathologist regarding the quality and importance of the selected regions of the WSIs.

Attention concentration metrics were computed from the per-tile attention weights produced by CHIEF, ABMIL, and gated ABMIL after softmax normalization within each patient bag. For each patient, tiles were ranked by attention and we computed Lorenz curves as the cumulative attention mass versus the cumulative fraction of tiles. We summarize concentration using the Gini coefficient, the fraction of tiles required to accumulate 50% and 80% of total mass, and top-k mass statistics (top 1, top 2, and top 25). To aggregate Lorenz curves across tasks without averaging attention for the same physical tile across different models or folds, we averaged rank profiles: for each patient we align by rank position (rank 1, rank 2, etc) and then average these ordered attention values across patients and tasks.

Furthermore, to investigate the robustness of EAGLE's top-tile selection, particularly in the presence of artifacts, we conducted a systematic review of its top 25 tiles on 50 randomly selected slides from the DACHS CRC cohort. A board-certified pathologist reviewed these tiles to determine the frequency of artifact focus (e.g., pen marks) versus tumor-rich regions. This performance was compared to a supervised baseline: Virchow2 tile embeddings were aggregated in STAMP on the MSI status prediction task. We computed Gradient-weighted Class Activation Mapping (Grad-CAM) attribution scores and selected the 25 tiles with the highest scores[45]. A board-certified pathologist analyzed the tiles selected by both EAGLE and STAMP, focusing on their clinical relevance and morphological significance.

For attention map visualizations, per-tile attention values were mapped back to WSI coordinates using the tile grid produced during tessellation. Heatmaps were rendered by assigning each tile its attention value. Unless stated otherwise, heatmap color scales were normalized per panel to visualize each model's dynamic range. Matched-scale controls are provided in **Supplementary Fig. 6**.

## Statistical Analysis

All classification performance results across tasks were aggregated from the five models produced in each cross-validation fold and summarized via metrics such as mean AUROC, AUPRC, balanced accuracy, and F1. Standard deviations were computed across the five folds, and two-sided DeLong's tests were applied to ensembled predictions (averaging fold probability scores for each sample) to assess differences in AUROC. Multi-class tasks like Lauren classification were excluded from these statistical tests because DeLong's procedure is not directly applicable in the multi-class setting. Throughout, we used Benjamini–Hochberg adjustments and considered p-values below 0.05 as evidence of statistical superiority. We used the c-index to evaluate the performance of the prognostic models for survival endpoints. Kaplan–Meier curves were generated to assess patient stratification, using the median predicted risk score as the cut-off value. The statistical significance of low-risk and high-risk patient groups was assessed using the log-rank test. For negative control experiments with R = 100 repeated uniform random tile selection, one-sided Monte Carlo p-values were computed as (r + 1) / (R + 1), where r is the number of random replicates that meet or exceed the CHIEF-selected result.

## Data availability

WSIs from TCGA are publicly accessible through the Genomic Data Commons Data Portal (https://portal.gdc.cancer.gov/), and CPTAC slides can be obtained from the CPTAC Data Portal (https://proteomics.cancer.gov/data-portal). All molecular data for these resources is available in cBioPortal (https://www.cbioportal.org/). The slides and biomarker data for DACHS were generated for prior studies[46–48] with restricted access. Biomarker data for DACHS are available by requesting Authorized Access to the phs001078 study [https://www.ncbi.nlm.nih.gov/projects/gap/cgi-bin/study.cgi?study_id=phs001113.v1.p1]. Applications for access to DACHS biomarker data are reserved for Senior Investigators and NIH Investigators as defined in https://dbgap.ncbi.nlm.nih.gov/aa/wga.cgi, and upon successful application grants access to the data for 1 year with the option to renew access. The slides for DACHS can only be requested directly through the DACHS principal investigators. The contact details are listed at http://dachs.dkfz.org/dachs/kontakt.html. The Kiel cohort is available from the Department of Pathology, Christian Albrechts University of Kiel, Kiel, Germany, upon reasonable request (https://www.medizin.uni-kiel.de/en/institutes-departments/institutes-of-clinical-theory/depart-

ment-of-pathology). The Bern cohort is proprietary and cannot be shared at the individual patient level. It is archived at the Institute of Pathology, University of Bern, and can be requested in reference to ref.[49]. The IEO cohort is held by the European Institute of Oncology, Milan. Data requests will be evaluated on a case-by-case basis in accordance with institutional policies and privacy regulations and can be directed via https://www.ieo.it/en/contact_us/. The datasets from the GECCO consortium are available from the corresponding author on reasonable request. The PathoBench dataset is publicly available on Hugging Face at https://huggingface.co/datasets/MahmoodLab/Patho-Bench. Source data are provided with this paper.

## Code availability

All benchmarking experiments build upon open-source STAMP software (v1.1.1). Publicly available foundation models (e.g., CTransPath, Prov-GigaPath, CONCH, Virchow, Virchow2, etc.) are accessible via GitHub (https://github.com/KatherLab/STAMP-Benchmark) under permissive licenses. The implementation of EAGLE, tested slide encoders, implementations of the MLP and LP classifiers, the GPT-4o–based in-context learning, UMAP visualization, top tiles and slide search are provided within the EAGLE repository (https://github.com/KatherLab/EAGLE). All experiments were conducted using NVIDIA RTX A6000, L40 or H100 GPUs.

# Extended Data

## Extended Data Figure 1: Model architectures and cohort overview.

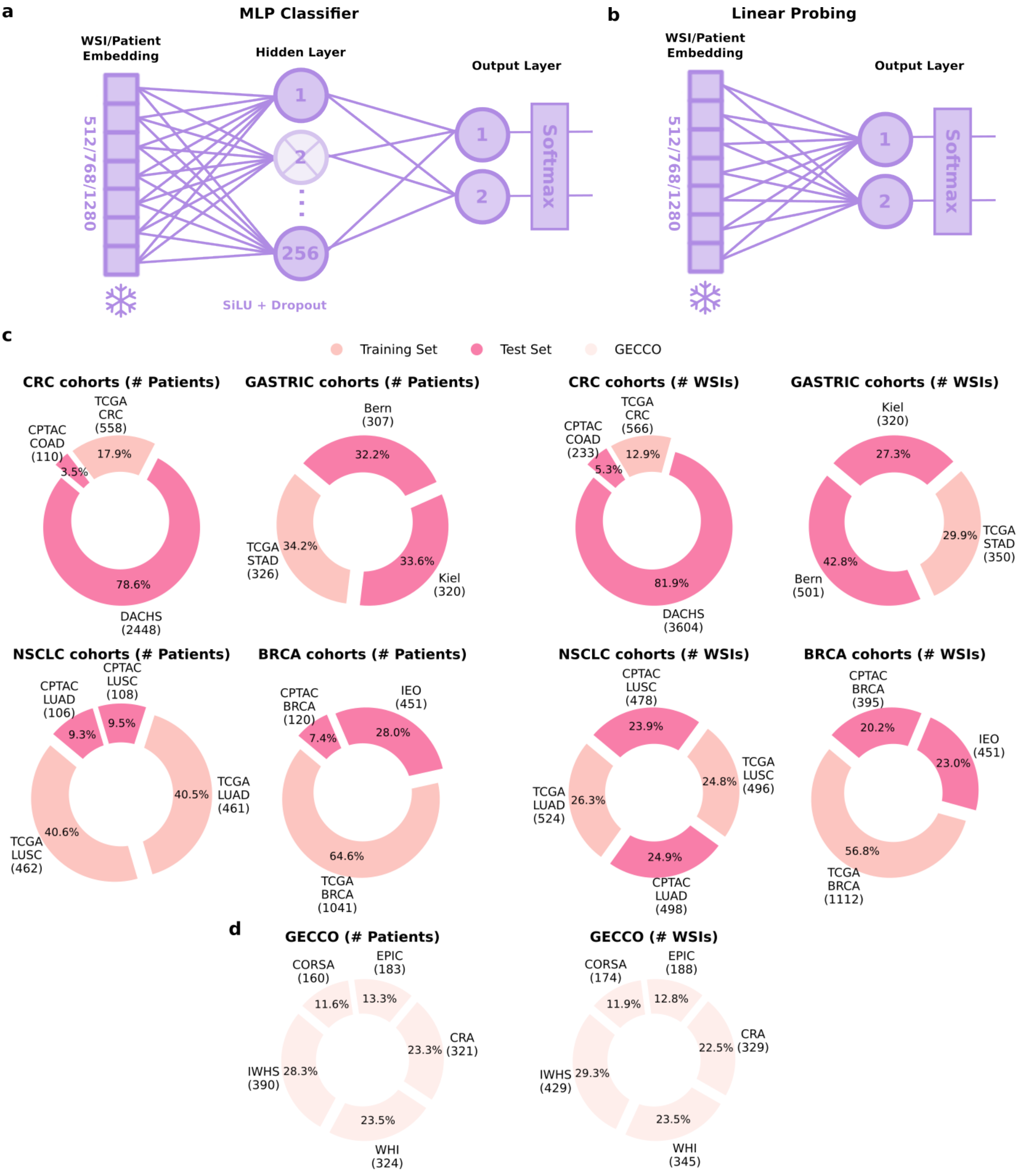

**a**) Architecture of the multilayer perceptron (MLP) classifier. Frozen whole-slide image or patient-level embeddings (dimensions 512, 768, or 1280) are passed through a 256-dimensional hidden layer with SiLU activation and dropout, followed by an output layer with two or three neurons depending on the task (binary or Lauren classification). A softmax function converts outputs into probability distributions. The classifier is trained using backpropagation with the AdamW optimizer. **b**) Architecture of the linear probing classifier. Frozen embeddings are directly connected to the output layer, followed by a softmax transformation. Logistic regression is implemented with the lbfgs solver and a maximum of 10,000 iterations. Detailed hyperparameters are provided in **Supplementary Table 8**. **c**) Composition of the main benchmarking dataset across four cancer types across 13 cohorts: colorectal cancer (CRC), gastric cancer (STAD), non-small cell lung cancer (NSCLC), and breast cancer (BRCA). TCGA-CRC, TCGA-STAD, TCGA-LUAD, TCGA-LUSC, and TCGA-BRCA were used for training, while CPTAC-COAD, CPTAC-LUAD, CPTAC-LUSC, CPTAC-BRCA, DACHS, Bern, Kiel, and IEO served as independent test cohorts. The number of patients is displayed on the left and the number of whole-slide images on the right, totaling 6,818 patients and 9,528 slides. **d**) Overview of the five cohorts included in the GECCO consortium dataset used for biomarker discovery (CORSA, EPIC, IWHS, CRA, and WHI) with varying allocations of the centers for training and testing across experiments. The numbers of patients and whole-slide images are shown for each cohort.

## Extended Data Figure 2: Comparative AUROC performance across 31 benchmarking tasks

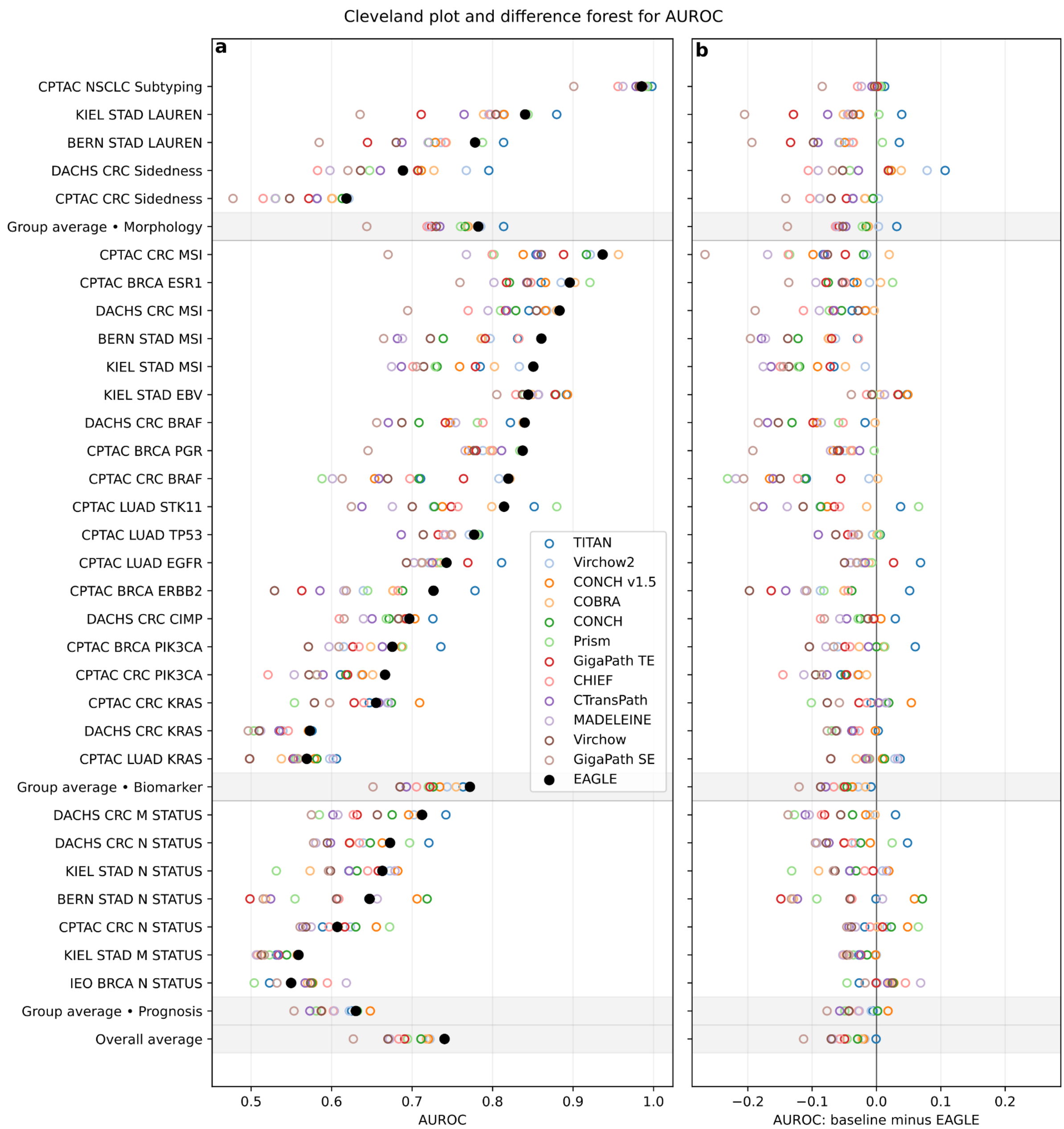

**a**) Performance of 13 foundation models across all benchmarking tasks, grouped by task category. Each point represents the mean AUROC across five folds for the best-performing magnification of a given model. EAGLE is highlighted with filled black markers, while other models are shown as unfilled circles. Group averages summarize performance across diagnosis, biomarker, prognosis, and treatment response tasks. **b**) Taskwise AUROC differences between each model and EAGLE (model − EAGLE), providing a direct comparison of relative performance across all 31 tasks.

## Extended Data Figure 3: Comparative performance stratified by task category and cancer type.

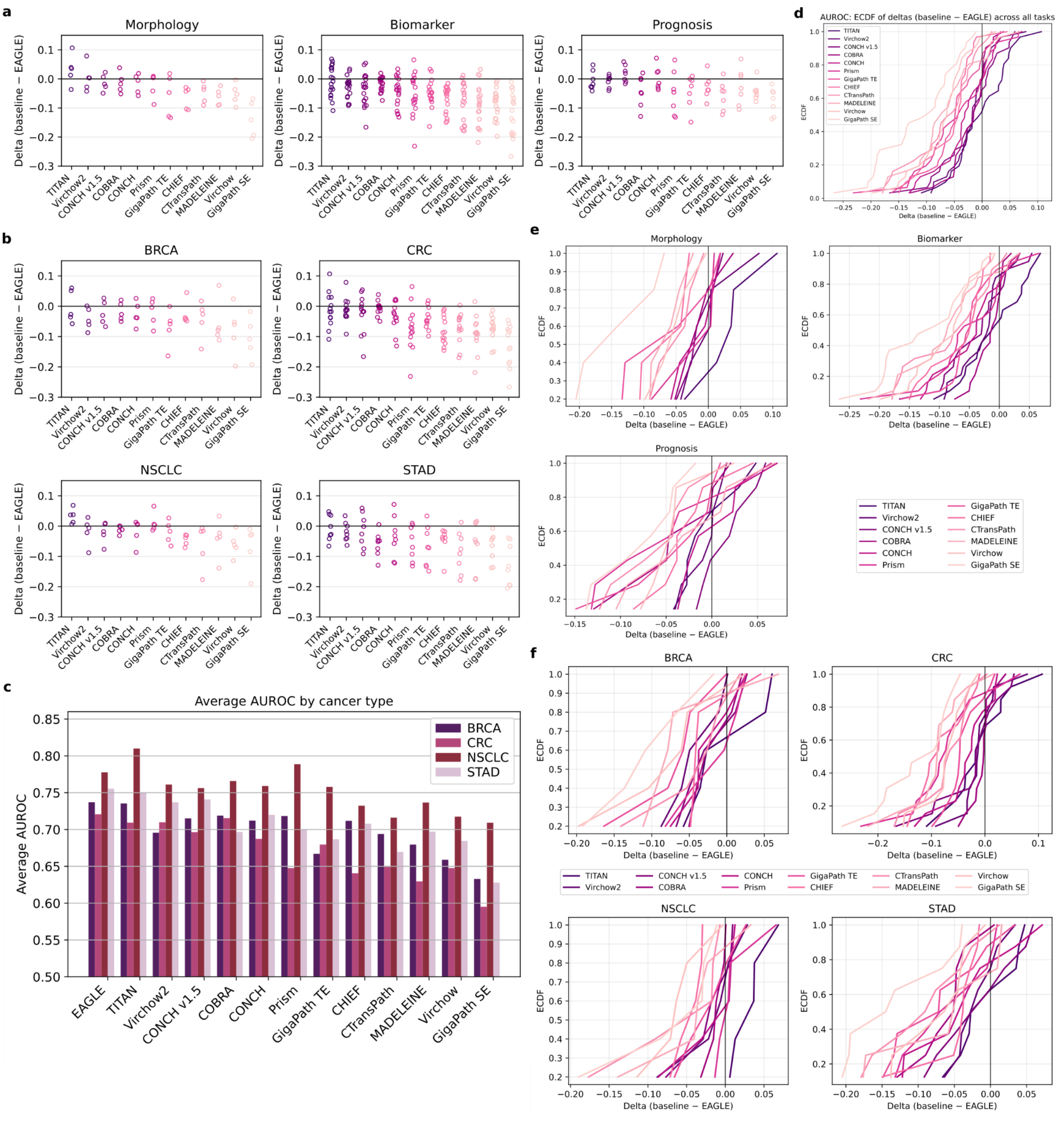

**a**) Taskwise AUROC differences between each model and EAGLE (model − EAGLE) for morphology, biomarker, and prognosis prediction tasks. Each point represents one task within the respective category. **b**) Taskwise AUROC differences (model − EAGLE) separated by cancer type—breast cancer (BRCA), colorectal cancer (CRC), non-small cell lung cancer (NSCLC), and stomach adenocarcinoma (STAD). **c**) Mean AUROC values per cancer type for all models. **d**) Empirical cumulative distribution function (ECDF) of AUROC differences (model − EAGLE) across all 31 tasks, comparing the distribution of relative performances for each model against EAGLE. **e**) ECDFs of AUROC differences (model − EAGLE) stratified by task category. **f**) ECDFs of AUROC differences (model − EAGLE) stratified by cancer type.

# Extended Data Figure 4: Ensemble performance and statistical comparison of EAGLE with baseline models.

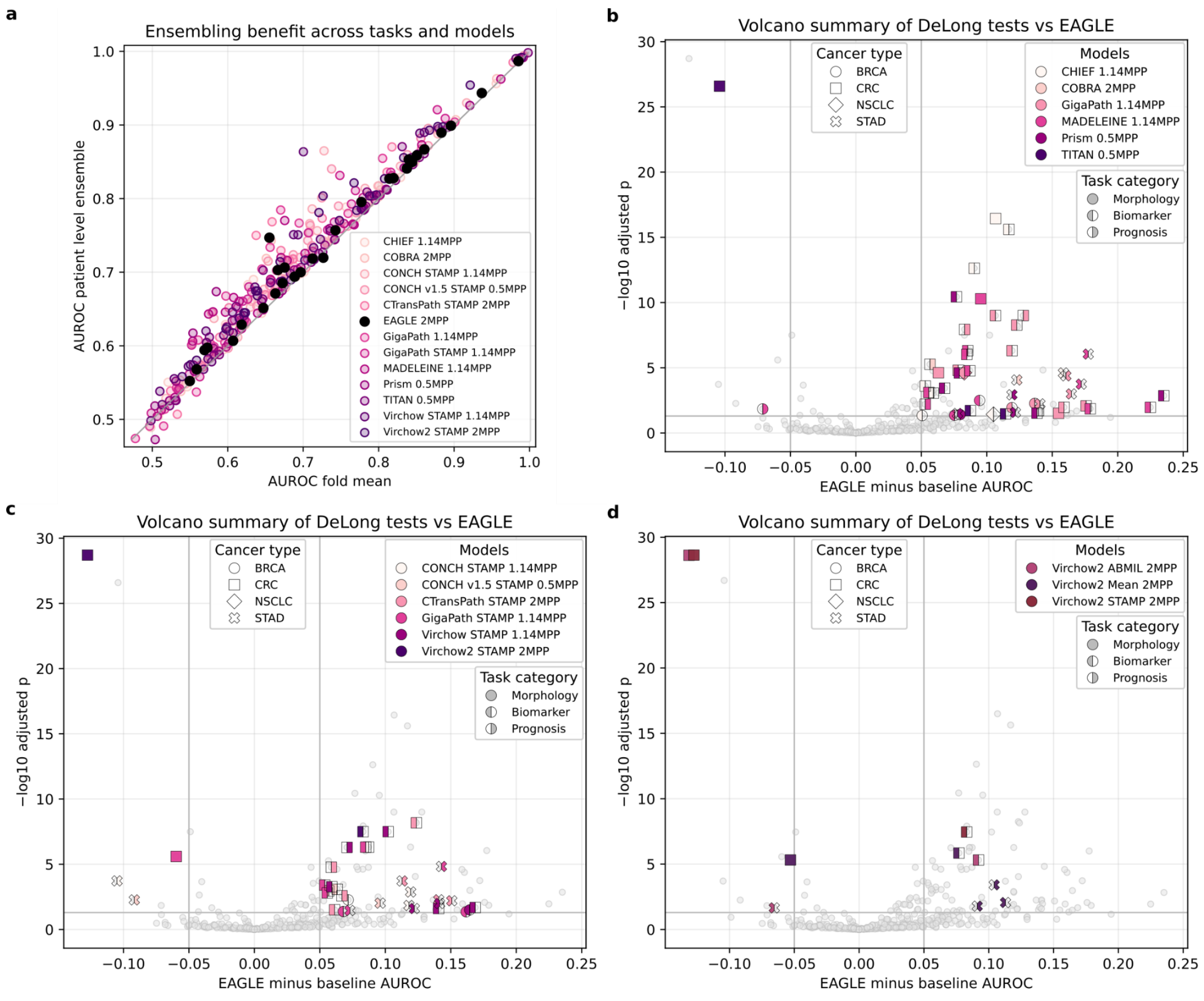

**a**) Relationship between mean fold performance and ensemble performance across all models and 31 tasks. The x-axis represents the average AUROC obtained by independently evaluating each fold, while the y-axis shows the AUROC derived from ensembling prediction scores across all five folds. Each point corresponds to one model-task pair, with EAGLE highlighted in black and other models shown in lighter tones. **b**) Statistical comparison between EAGLE and slide-level encoders based on ensemble predictions. Two-sided DeLong tests were performed to evaluate AUROC differences, and p-values were adjusted for multiple testing using the Benjamini–Hochberg procedure. Tasks with an adjusted p-value < 0.05 and an AUROC difference > 0.05 are highlighted. **c**) Statistical comparison between EAGLE and tile encoder models aggregated with STAMP, applying the same testing and adjustment procedure. **d**) Comparison of alternative aggregation strategies applied to Virchow2 embeddings. Models using ABMIL, simple averaging, or STAMP aggregation are each compared to EAGLE using ensemble-level DeLong tests. Significant tasks (adjusted p < 0.05 and AUROC difference > 0.05) are highlighted.

# Extended Data Figure 5: Detailed ablation analyses of EAGLE components.

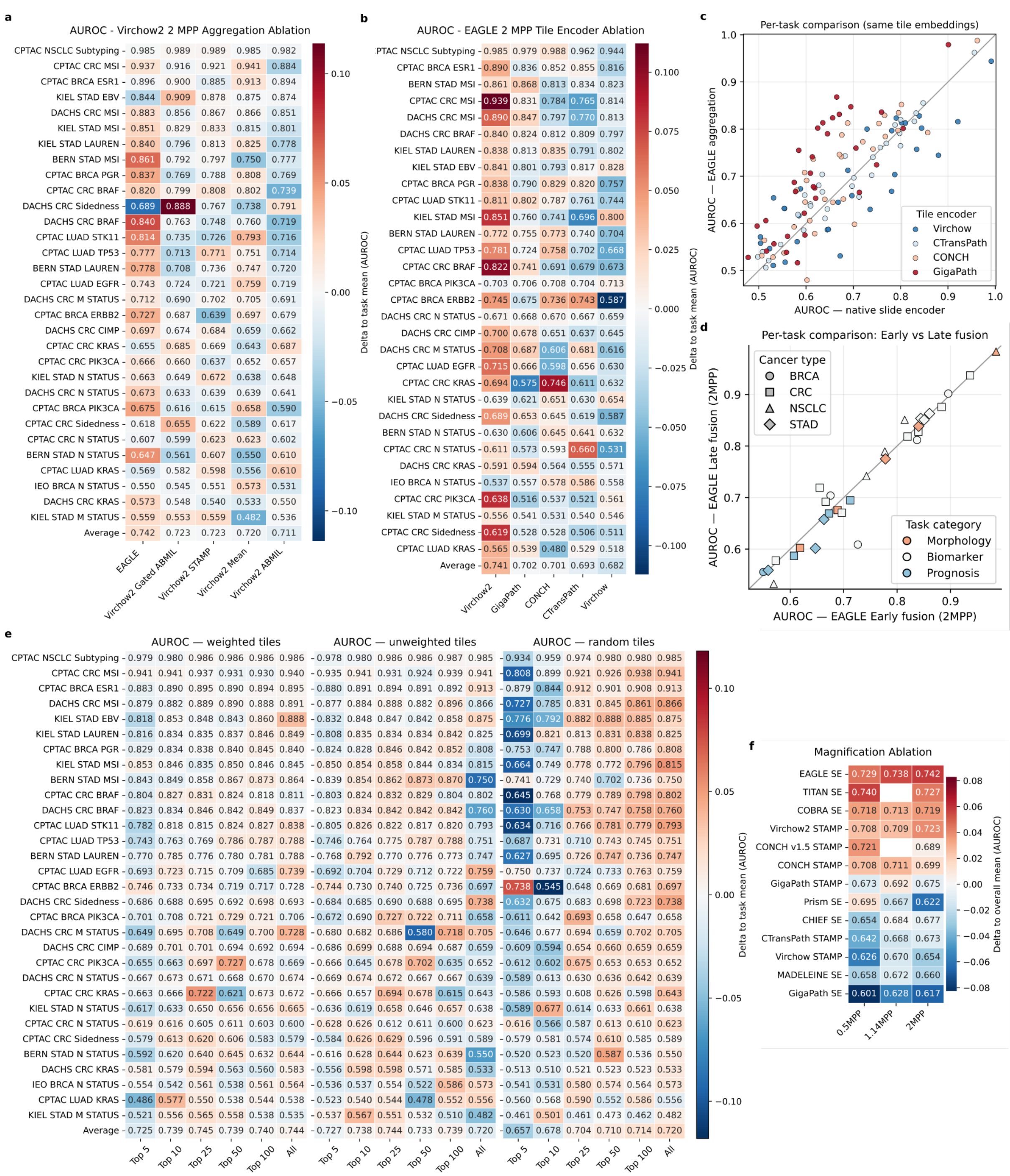

**a**) Taskwise performance comparison of Virchow2 aggregation strategies. AUROC values are shown for EAGLE, gated ABMIL, STAMP, mean aggregation, and ABMIL across all 31 tasks. Color intensity represents the deviation of each model's AUROC from the task-specific mean: red indicates above-average performance, blue below-average, and white near the task mean.

This normalization highlights relative differences between models rather than overall task difficulty. **b**) Tile encoder ablation using the EAGLE framework at 2 microns per pixel (MPP). Results are shown for EAGLE (Virchow2) and alternative tile encoders, illustrating the effect of the tile feature extractor on overall performance across all tasks. **c**) Native slide encoders versus EAGLE with the same tile embeddings, per task. Scatter plot with the native slide encoder on the x axis and EAGLE on the y axis. Points above the diagonal favor EAGLE. **d**) Early fusion versus late fusion for EAGLE. Scatter plot with early fusion on the x axis and late fusion on the y axis. Points above the diagonal favor late fusion. **e**) Influence of tile sampling and aggregation strategy. Results are shown for 5, 10, 25, 50, 100, and all tiles per patient using weighted averaging (CHIEF attention), unweighted averaging, and random tile selection. **f**) Analysis of magnification settings at 0.5, 1.14, and 2.0 microns per pixel (MPP) for all 13 tested models, reporting the mean AUROC across all 31 tasks to assess the effect of image resolution.

## Extended Data Figure 6: Detailed analysis of hyperparameter tuning experiments.

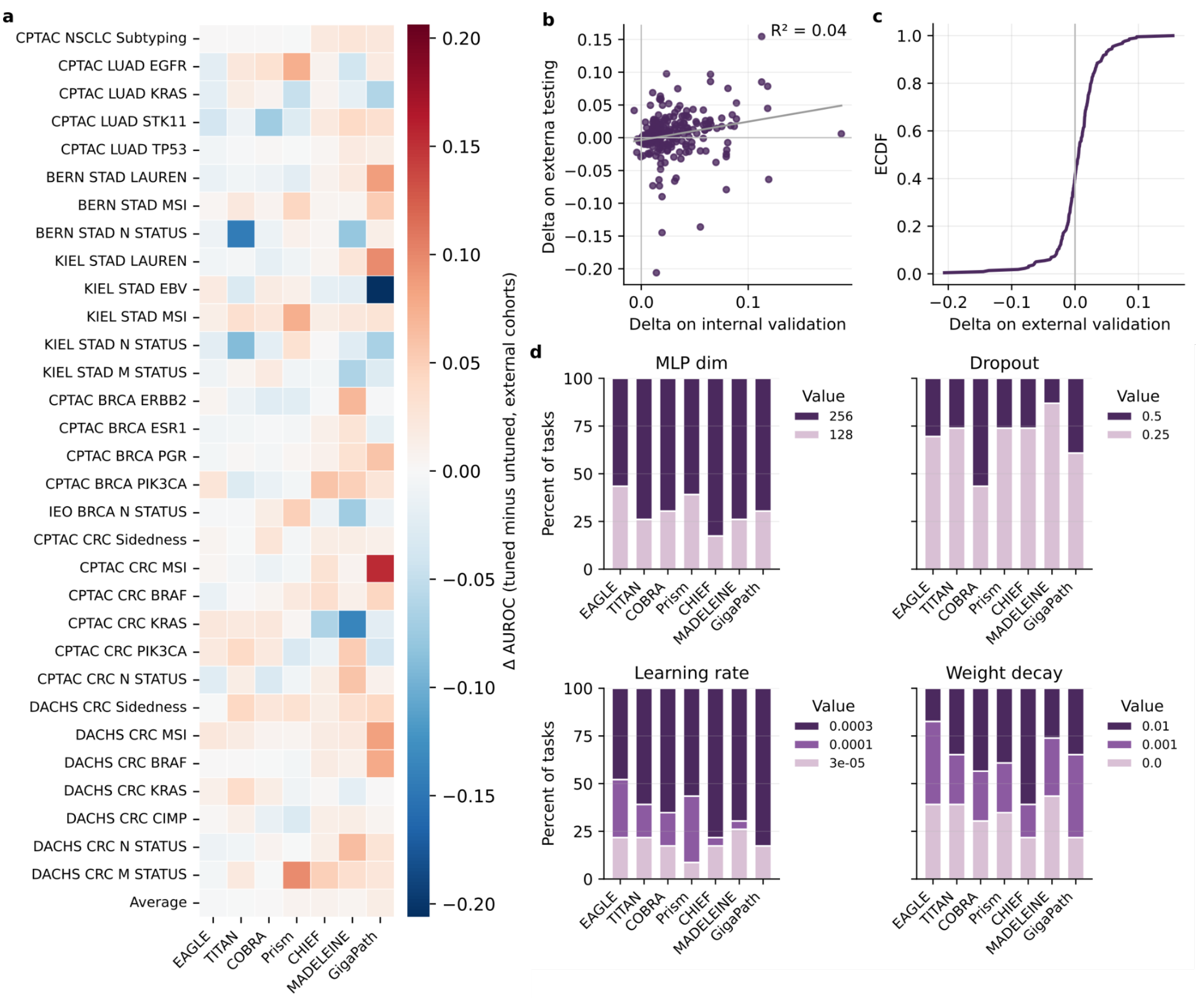

**a**) Taskwise AUROC differences between tuned and untuned multilayer perceptron (MLP) classifiers trained on patient-level embeddings from EAGLE and slide encoders across 31 external test tasks. Red indicates improved performance after tuning, blue a decrease, and white negligible change. **b**) Relationship between tuning gains on internal validation and external testing. Each point represents a model–task pair, showing the AUROC difference (tuned − untuned) on the internal validation (x-axis) and external test (y-axis) cohorts. The positive but weak correlation ($R^2 = 0.04$) indicates that improvements seen during internal validation only partially translate to external performance. **c**) Empirical cumulative distribution function (ECDF) of AUROC differences (tuned − untuned) on external test cohorts across all models and tasks. **d**) Distribution of selected hyperparameters for the best-performing MLP configurations on internal validation. The search included hidden layer dimensions (128 or 256), dropout probabilities (0.25 or 0.5), learning rates ($3 \times 10^{-5}$, $1 \times 10^{-4}$, or $3 \times 10^{-4}$), and weight decay values (0, 0.001, or 0.01). Bars indicate the frequency with which each parameter combination yielded the best internal validation result.

# Extended Data Figure 7: Detailed survival analysis across models and cohorts

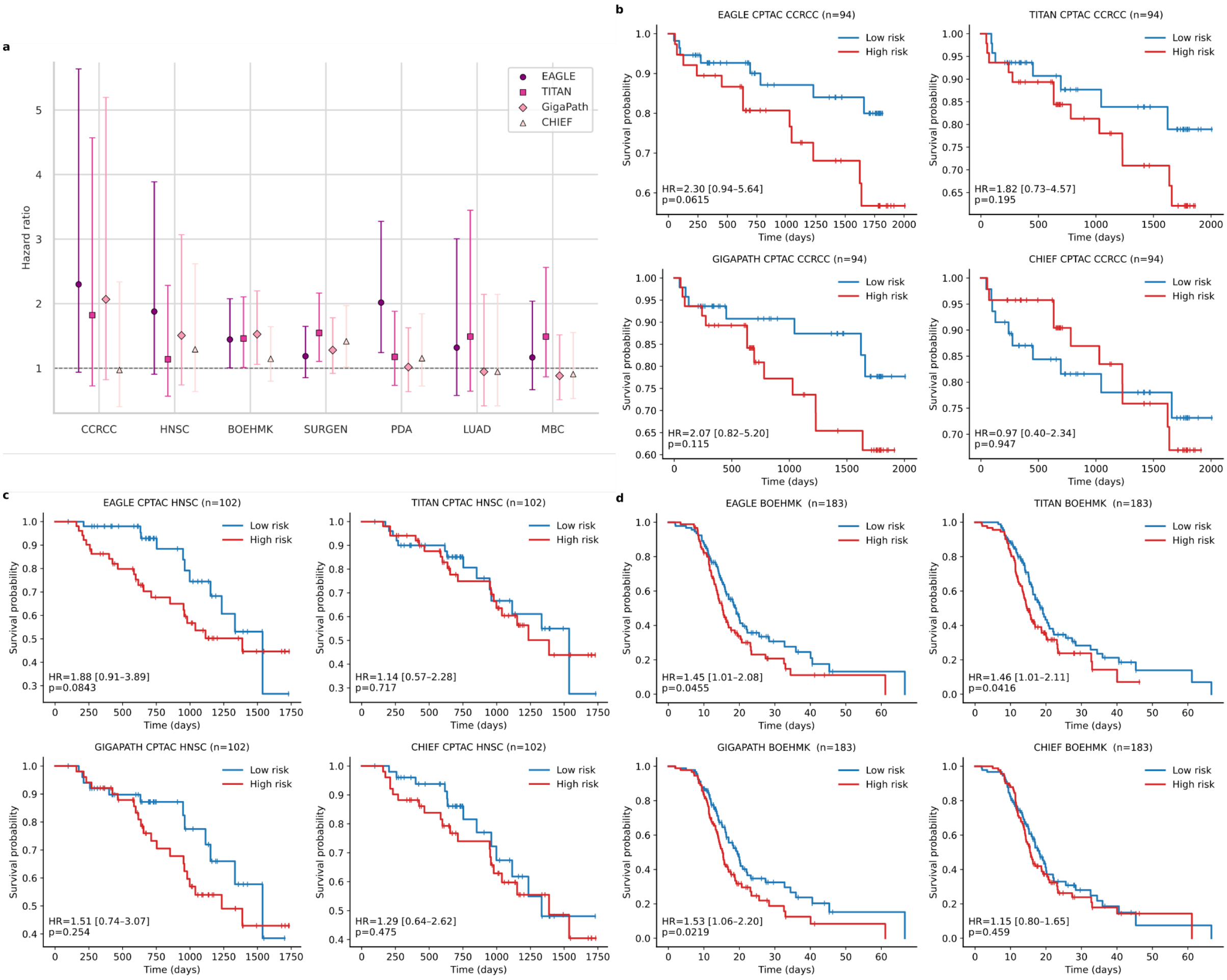

**a**) Hazard ratios with 95% confidence intervals for EAGLE, TITAN, GigaPath, and CHIEF across the seven survival prediction tasks. For each task, patients were divided at the median model-predicted risk score into low-risk and high-risk groups. Hazard ratios were computed from Cox proportional hazards models, and p-values were derived using two-sided log-rank tests. **b–d**) Kaplan–Meier survival curves for three representative cohorts: CPTAC clear cell renal cell carcinoma (CCRCC) (**b**), CPTAC head and neck squamous cell carcinoma (HNSC) (**c**), and BOEHMK (**d**). For each model, patients were stratified into low- and high-risk groups at the median risk score, and survival differences were assessed using the two-sided log-rank test.

# Extended Data Figure 8: Detailed analysis of few-shot and reduced-patient experiments.

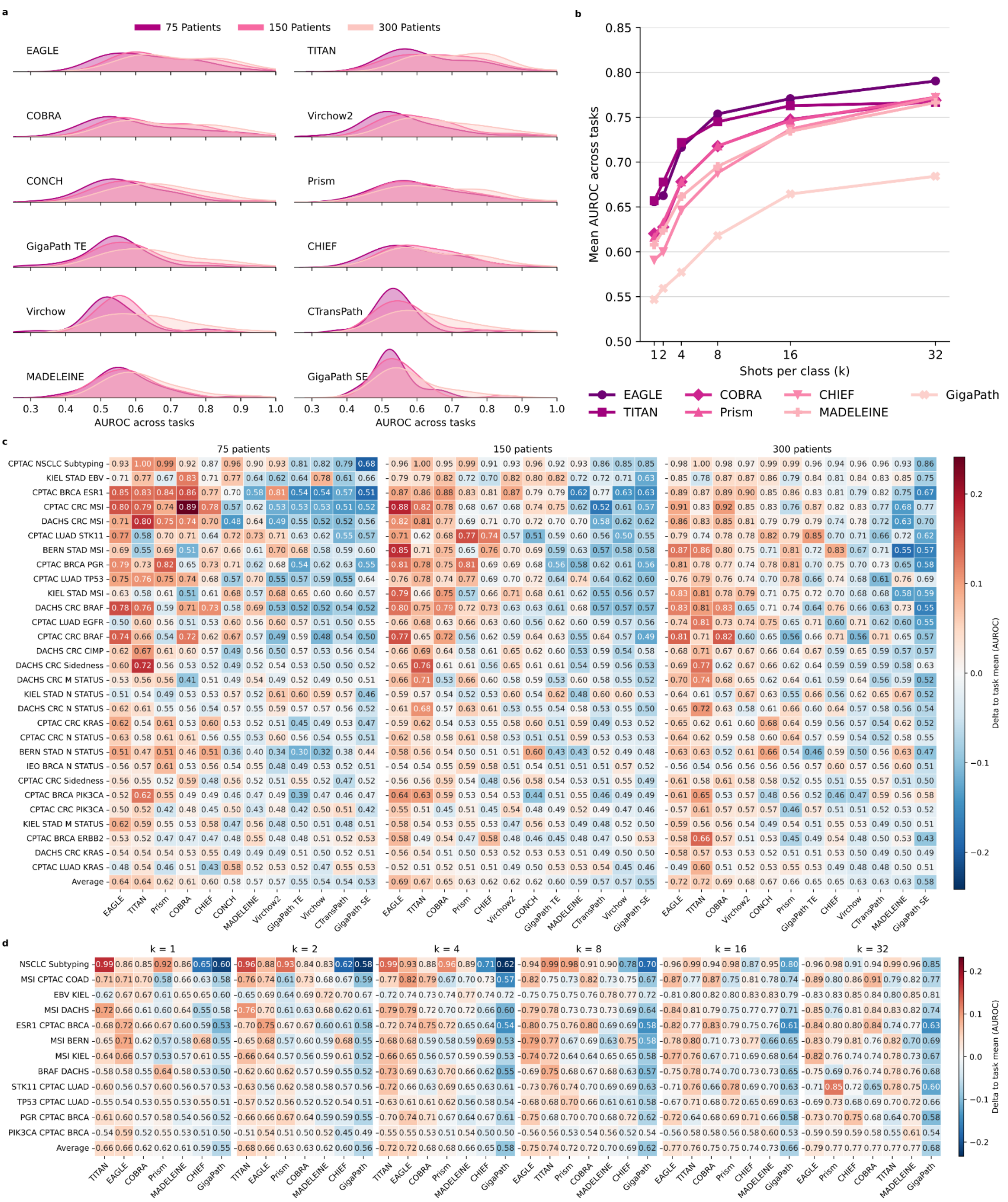

**a**) Distribution of AUROC values across 29 tasks for models trained with 75, 150, and 300 patients. Each density curve represents one training size, showing how model performance varies across tasks as data availability increases. The horizontal axis indicates AUROC, and the vertical density reflects the number of tasks achieving a given performance level. Note that the Lauren classification tasks in the Bern and Kiel cohorts were excluded owing to limited patient numbers. **b**) Few-shot learning performance of EAGLE and slide encoder models. The horizontal axis shows the number of samples per class (k = 1, 2, 4, 8, 16, or 32), and the vertical axis shows the mean AUROC across all tasks. To ensure robustness, only the top three binary tasks per cancer type—selected based on the highest mean AUROC across all models—were included in the few-shot analysis. **c**) Taskwise performance differences across models in the reduced-patient experiments. AUROC values are displayed for 75, 150, and 300 patients across all 29 tasks, with colors normalized per task to indicate deviation from the task mean. Red tones indicate above-average performance, blue below-average, and white near the task mean, emphasizing relative model differences rather than task difficulty. **d**) Detailed few-shot performance across all evaluated models and k values (1, 2, 4, 8, 16, 32). The heat map displays the AUROC for each model–task–k combination, with the same color normalization as in panel c.

## Extended Data Figure 9: Comparison of tile selection and embedding representations.

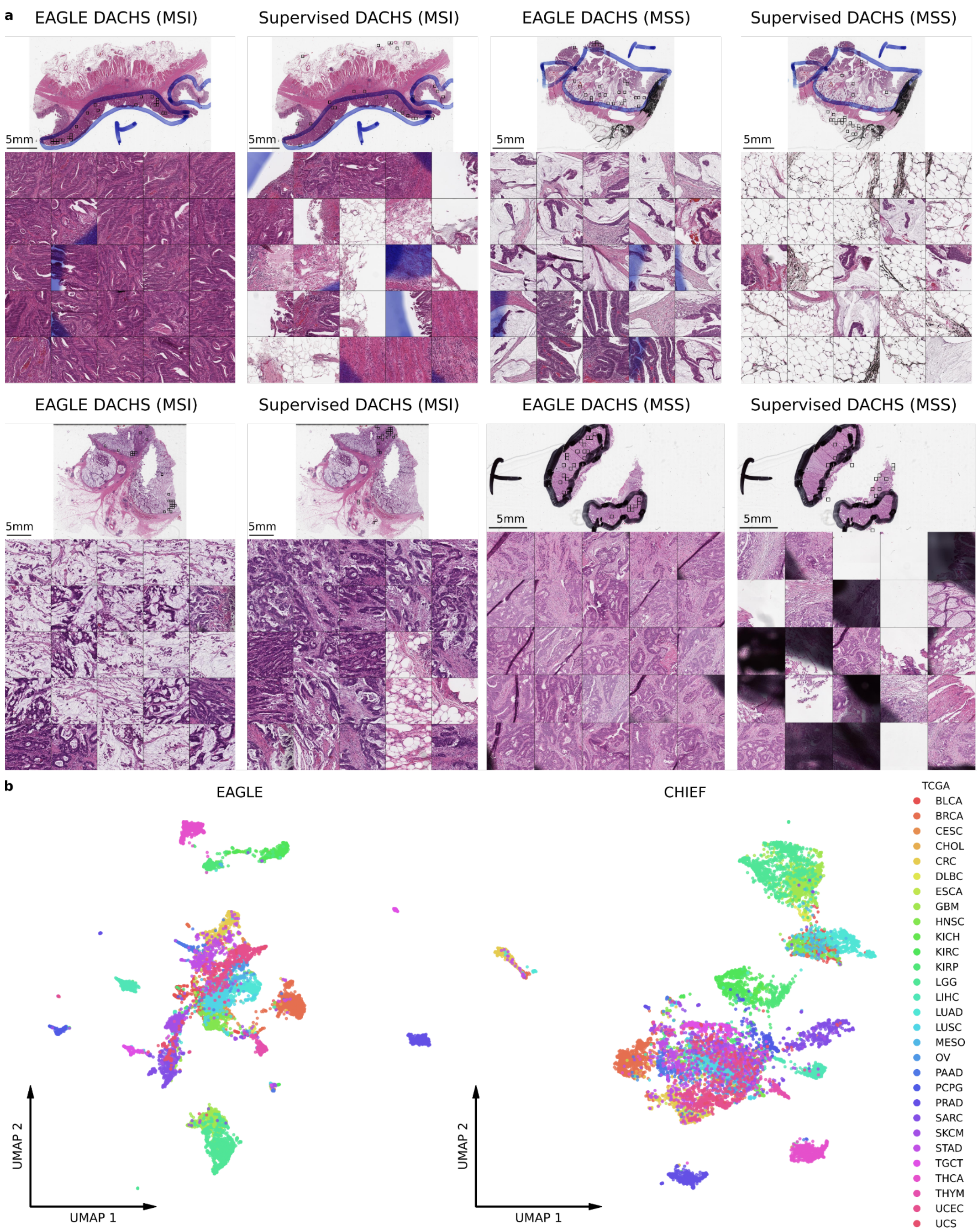

**a**) Top 25 tiles selected by EAGLE and by the supervised baseline (Virchow2 embeddings aggregated via the STAMP pipeline) for four representative DACHS slides, including two microsatellite instability–high (MSI) and two microsatellite stable (MSS) cases. EAGLE consist-ently prioritized diagnostically relevant tumor regions, whereas the supervised baseline frequently included non-tumor or artifact-rich areas such as slide edges, background, or pen marks. **b**) Comparison of slide embeddings generated by CHIEF and EAGLE using Uniform Manifold Approximation and Projection (UMAP) across 29 TCGA cohorts. The visualization illustrates clearer clustering by tissue type and biological signal in the EAGLE embedding space relative to CHIEF.

**Extended Data Figure 10: Comparison of EAGLE's few-shot linear probing with GPT-4o's in-context learning**

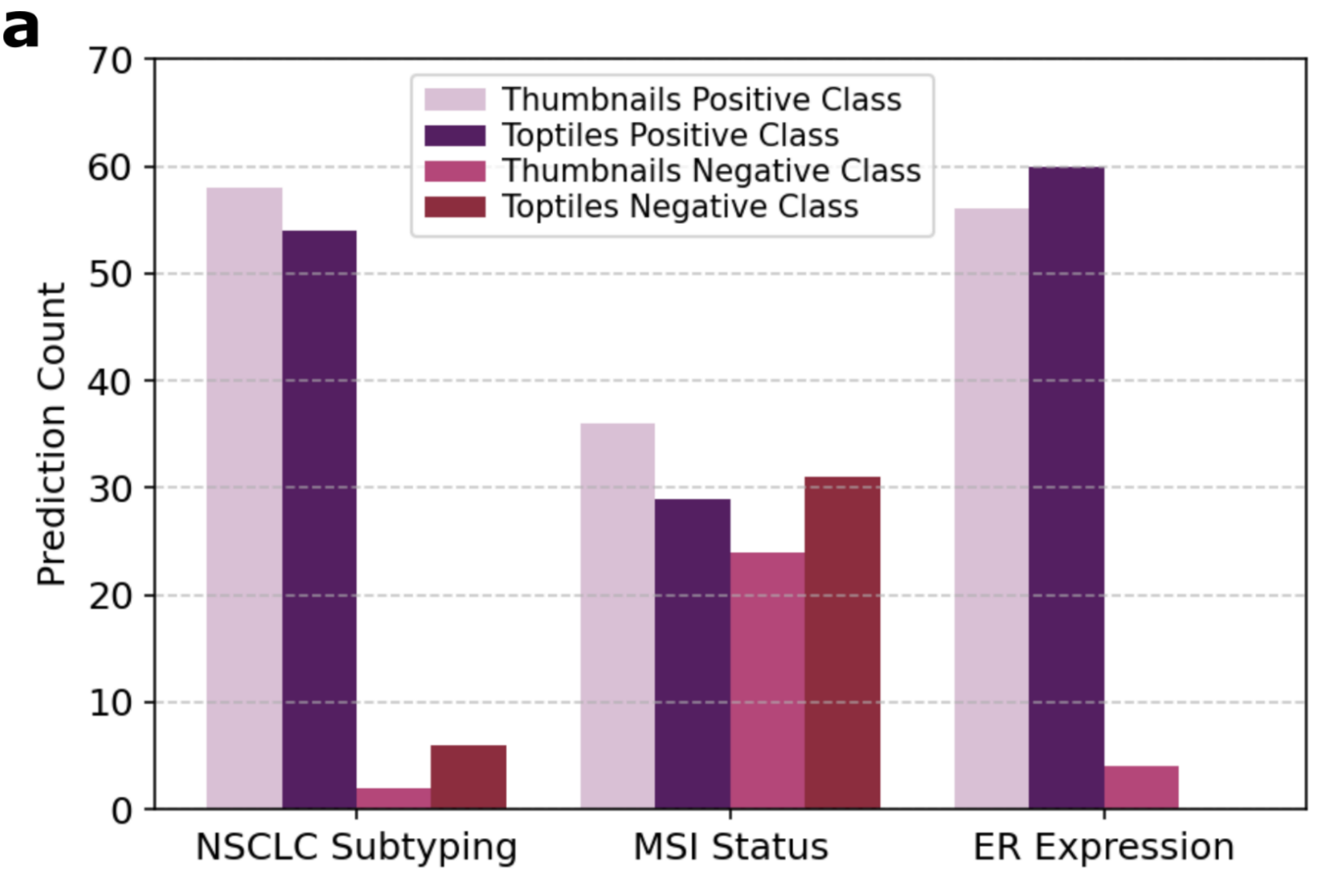

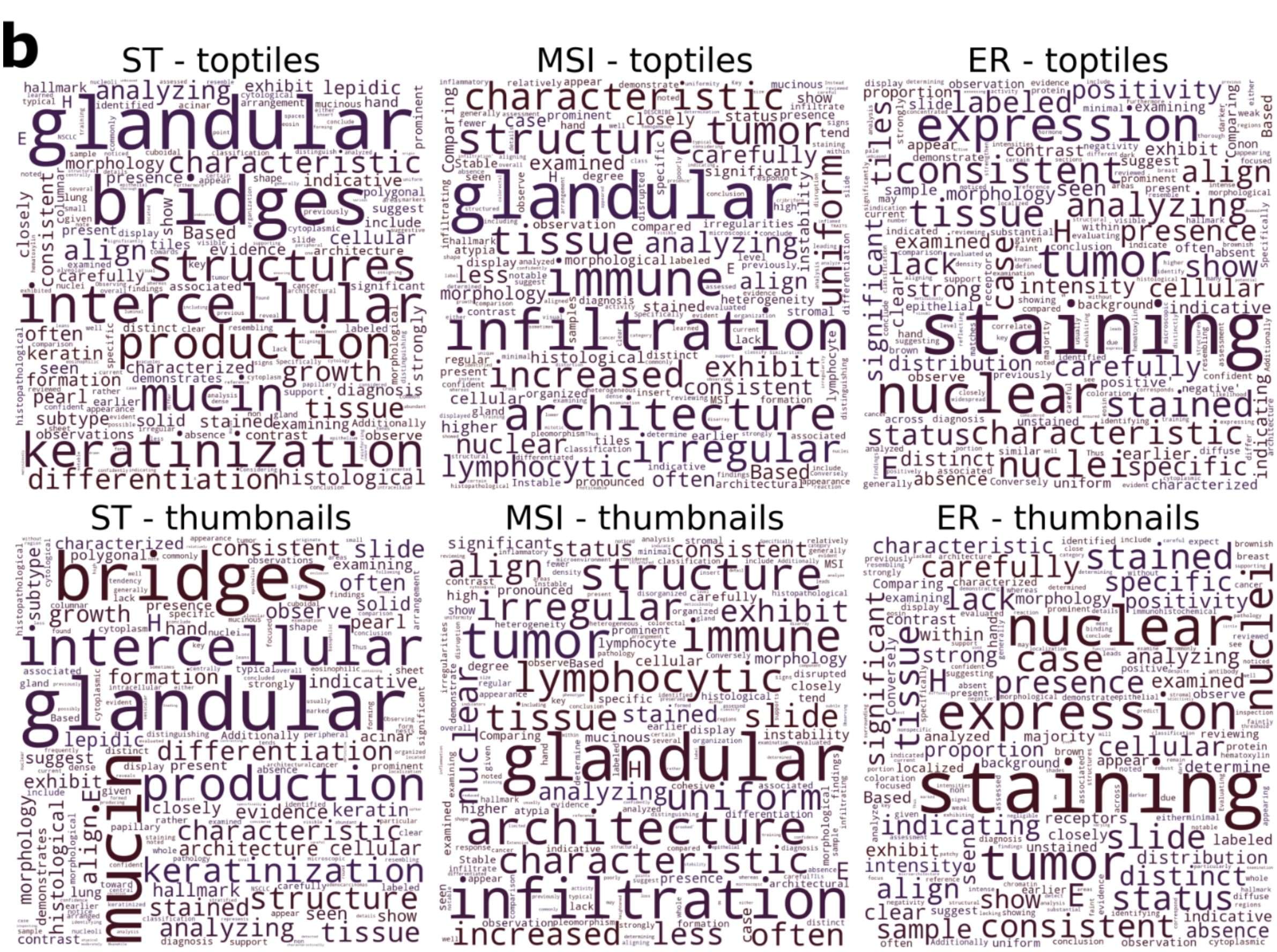

Comparison of EAGLE's few-shot linear probing (using k=2 examples per class) versus GPT-4o's in-context learning (k=2) for ER expression prediction in BRCA, MSI status in CRC, and NSCLC subtyping. Two examples per class (with ground truth) were provided for each task, NSCLC subtyping (adenocarcinoma as the positive class), MSI prediction (MSI-high), and ER expression (ER-positive), followed by a query image, with three runs of 20 examples per task. **a**) frequency with which GPT-4o predicted the positive class across 60 predictions per task, comparing approaches that use either a thumbnail of the original WSI or the high-resolution 25 top tiles selected by EAGLE. **b**) Word clouds generated from the "thoughts" section of GPT-4o's responses (aggregated over 60 runs per experiment) for each input type; common and uninformative stopwords, including task names, class labels, and frequently observed filler words, were excluded.

# Supplementary Information

## Supplementary Tables

**Supplementary Table 1: Performance distributions for uniform random tile selection across 100 replicates.**

| | Random replicates | | | | | | | | |
|---|---|---|---|---|---|---|---|---|---|
| **N** | **mean** | **median** | **IQR** | **min** | **max** | **2.5%** | **97.5%** | **EAGLE** | **Monte Carlo p-value** |
| **5** | 0.653 | 0.652 | 0.009 | 0.638 | 0.667 | 0.641 | 0.665 | 0.727 | 0.001 |
| **10** | 0.681 | 0.681 | 0.006 | 0.665 | 0.697 | 0.670 | 0.692 | 0.738 | 0.001 |
| **25** | 0.703 | 0.703 | 0.005 | 0.691 | 0.713 | 0.695 | 0.710 | 0.744 | 0.001 |
| **50** | 0.710 | 0.710 | 0.004 | 0.699 | 0.718 | 0.704 | 0.716 | 0.733 | 0.001 |
| **100** | 0.714 | 0.714 | 0.003 | 0.707 | 0.723 | 0.709 | 0.719 | 0.739 | 0.001 |

For each tile budget N in {5, 10, 25, 50, 100}, we report the mean, median, interquartile range, minimum, maximum, and 2.5% to 97.5% empirical quantiles of mean AUROC across all tasks and folds under 100 uniform random selections. The corresponding CHIEF-based top-N result and the Monte Carlo p-value computed as $(r + 1) / (R + 1)$ with $R = 100$ are provided for comparison.

**Supplementary Table 2: Attention concentration statistics across CHIEF, ABMIL, and gated ABMIL.**

| Method | Top-1 tile mass | Top-2 tiles mass | Top-25 tiles mass | Tiles to reach 50% of mass | Tiles to reach 80% of mass |
|---|---|---|---|---|---|
| **CHIEF** | 0.55% | 1.1% | 10.9% | 8.4% | 20.5% |
| **ABMIL** | 0.077% | 0.15% | 1.8% | 44.1% | 75.8% |
| **Gated ABMIL** | 0.11% | 0.22% | 2.3% | 42.0% | 73.6% |

Attention weights were softmax normalized within each patient bag. Tiles were ranked by attention score per patient. Reported values are medians across all patient bags, tasks, and cross validation folds. “Top k tile mass” denotes the cumulative attention mass assigned to the k highest ranked tiles within a bag. “Tiles to reach 50% or 80% of mass” denotes the fraction of tiles required to accumulate the specified proportion of total attention mass. CHIEF exhibits substantially higher attention concentration, whereas end to end weak supervision via ABMIL and gated ABMIL approaches near uniform aggregation behavior across most tasks.

**Supplementary Table 3: Computational efficiency metrics**

| Measured FLOPs of Tile Encoders | | | | |
|---|---|---|---|---|
| **Model** | **Dataset** | **FLOPs per tile** | **Average FLOPs per WSI** | **Total FLOPs** |
| CTransPath | 0.5 MPP | $8.78 \times 10^{11}$ | $1.56 \times 10^{16}$ | $1.49 \times 10^{20}$ |
| CTransPath | 2 MPP | $8.78 \times 10^{11}$ | $1.15 \times 10^{15}$ | $1.09 \times 10^{19}$ |
| Virchow | 0.5 MPP | $2.26 \times 10^{13}$ | $4.03 \times 10^{17}$ | $3.84 \times 10^{21}$ |
| Virchow | 2 MPP | $2.26 \times 10^{13}$ | $2.96 \times 10^{16}$ | $2.82 \times 10^{20}$ |
| Virchow2 | 0.5 MPP | $2.31 \times 10^{13}$ | $4.12 \times 10^{17}$ | $3.92 \times 10^{21}$ |
| Virchow2 | 2 MPP | $2.31 \times 10^{13}$ | $3.02 \times 10^{16}$ | $2.88 \times 10^{20}$ |
| Prov-GigaPath | 0.5 MPP | $3.08 \times 10^{13}$ | $5.48 \times 10^{17}$ | $5.23 \times 10^{21}$ |
| Prov-GigaPath | 2 MPP | $3.08 \times 10^{13}$ | $4.03 \times 10^{16}$ | $3.84 \times 10^{20}$ |
| CONCH v1.5 | 0.5 MPP | $1.2 \times 10^{13}$ | $4.1 \times 10^{16}$ | $3.9 \times 10^{20}$ |
| CONCH v1.5 | 2 MPP | $1.2 \times 10^{13}$ | $3.39 \times 10^{15}$ | $3.23 \times 10^{19}$ |
| | | | | |
| **Measured Time of Tile Encoders** | | | | |
| **Model** | **Dataset** | **Average Time per WSI (seconds)** | | **Total Time (hours)** |
| CONCH | 0.5 MPP | 56.55 | | 149.51 |
| CONCH | 2 MPP | 4.24 | | 11.2 |
| Virchow | 0.5 MPP | 182.2 | | 481.71 |
| Virchow | 2 MPP | 13.61 | | 35.99 |
| CTransPath | 0.5 MPP | 25.4 | | 67.23 |
| CTransPath | 1 MPP | 6.78 | | 17.91 |
| CTransPath | 2 MPP | 2.01 | | 5.32 |
| Virchow2 | 0.5 MPP | 185.43 | | 490.77 |
| Virchow2 | 2 MPP | 13.89 | | 36.75 |
| Prov-GigaPath | 0.5 MPP | 714.36 | | 1890.68 |
| Prov-GigaPath | 2 MPP | 53.42 | | 141.38 |
| CONCH v1.5 | 0.5 MPP | 191.85 | | 507.76 |
| CONCH v1.5 | 2 MPP | 14.49 | | 38.31 |
| Top 25 Tiles Virchow2: | 0.5 MPP | 0.26 | | 0.5 |
| Top 25 Tiles Virchow2: | 2 MPP | 0.26 | | 0.49 |
| | | | | |
| **Measured Time of Slide Encoders** | | | | |
| **Model** | **Dataset** | **Average Time per WSI (milliseconds)** | | **Total Time (seconds)** |
| CHIEF | 0.5 MPP | 1.71 | | 11.65 |
| CHIEF | 2 MPP | 0.36 | | 2.48 |
| MADELEINE | 0.5 MPP | 11.56 | | 78.84 |
| MADELEINE | 2 MPP | 2.40 | | 16.38 |

| Prism | 0.5 MPP | 153.28 | 1045.09 |
| --- | --- | --- | --- |
| Prism | 2 MPP | 141.49 | 964.69 |
| Prov-GigaPath | 0.5 MPP | 256.55 | 1749.15 |
| Prov-GigaPath | 2 MPP | 44.14 | 300.94 |
| TITAN | 0.5 MPP | 3295.56 | 22469.11 |
| TITAN | 2 MPP | 12.14 | 82.79 |

**Supplementary Table 4: Rare biomarker discovery results**

| Biomarker | AUROC | AUPRC | Train Cohorts | Test Cohorts | n train | n test | Training Pos (%) | Test Pos (%) |
|---|---|---|---|---|---|---|---|---|
| exercise_missing | 0.964 | 0.968 | CORSA; EPIC; CRA | WHI; IWHS | 664 | 714 | 72.44 | 54.62 |
| msi | 0.963 | 0.893 | CORSA; EPIC; WHI | CRA; IWHS | 667 | 711 | 24.44 | 24.19 |
| RNF43 (truncating) | 0.904 | 0.676 | CORSA; EPIC; IWHS | CRA; WHI | 733 | 645 | 16.78 | 19.53 |
| BRAF V600 | 0.894 | 0.699 | CORSA; CRA; WHI | EPIC; IWHS | 805 | 573 | 14.29 | 25.65 |
| hypermutated | 0.888 | 0.748 | CORSA; EPIC; IWHS | CRA; WHI | 733 | 645 | 23.87 | 27.6 |
| RNF43 | 0.886 | 0.688 | CORSA; EPIC; IWHS | CRA; WHI | 733 | 645 | 19.1 | 22.17 |
| BMPR2 | 0.879 | 0.481 | CORSA; EPIC; IWHS | CRA; WHI | 733 | 645 | 14.32 | 16.43 |
| AKAP7 | 0.868 | 0.374 | CORSA; EPIC; CRA | WHI; IWHS | 504 | 390 | 7.54 | 6.41 |
| TGF beta | 0.846 | 0.801 | CORSA; EPIC; IWHS | CRA; WHI | 733 | 645 | 37.65 | 41.86 |
| ZNRF3 | 0.84 | 0.415 | CORSA; CRA; WHI | EPIC; IWHS | 805 | 573 | 9.69 | 12.39 |
| BRAF | 0.828 | 0.673 | CORSA; CRA; WHI | EPIC; IWHS | 805 | 573 | 18.01 | 29.67 |
| CASP8 | 0.826 | 0.277 | CORSA; CRA; WHI | EPIC; IWHS | 805 | 573 | 6.09 | 7.85 |
| MBD6 | 0.82 | 0.325 | CORSA; EPIC; CRA | WHI; IWHS | 504 | 390 | 4.76 | 6.67 |
| ARID3A | 0.815 | 0.167 | CORSA; CRA; IWHS | EPIC; WHI | 871 | 507 | 2.76 | 4.73 |
| ZBTB20 | 0.811 | 0.365 | CORSA; EPIC; CRA | WHI; IWHS | 504 | 390 | 6.35 | 11.28 |
| FHOD3 | 0.811 | 0.411 | CORSA; EPIC; CRA | WHI; IWHS | 504 | 390 | 10.91 | 14.1 |
| EP300 | 0.811 | 0.284 | CORSA; EPIC; IWHS | CRA; WHI | 733 | 645 | 7.5 | 8.84 |
| MECOM | 0.802 | 0.262 | CORSA; EPIC; CRA | WHI; IWHS | 504 | 390 | 7.14 | 10.26 |
| B2M | 0.802 | 0.235 | CORSA; EPIC; WHI | CRA; IWHS | 667 | 711 | 6.9 | 7.45 |
| PBRM1 | 0.801 | 0.16 | CORSA; EPIC; WHI | CRA; IWHS | 667 | 711 | 4.5 | 4.36 |

**Supplementary Table 5: Clinicopathological data of the benchmarking cohorts**

| Dataset | | TCGA | | | | | CPTAC | | | | DAC HS | Bern | Kiel | IEO |
|---|---|---|---|---|---|---|---|---|---|---|---|---|---|---|
| | | BRCA | CRC | LUAD | LUSC | STAD | BRCA | COAD | LUAD | LUSC | | | | |
| Total patients | | 1041 | 558 | 461 | 462 | 326 | 120 | 110 | 106 | 108 | 2448 | 307 | 320 | 451 |
| Age | median | 58 | 67 | 66 | 68 | 67 | 61.5 | 65.5 | 63.5 | 67 | 69 | 72.3 | 68.2 | 49 |
| | IQR | 19 | 18 | 13.8 | 11 | 14 | 20.7 | 19 | 12 | 11 | 14 | 16.3 | 14.6 | 13.4 |
| | <50 | 281 | 69 | 31 | 16 | 26 | 23 | 7 | 12 | 5 | 123 | 22 | 14 | 227 |
| | unknown | 0 | 1 | 19 | 11 | 3 | 15 | 2 | 0 | 0 | 0 | 0 | 0 | 0 |
| Sex | male | 12 | 287 | 213 | 343 | 220 | 0 | 45 | 68 | 86 | 1436 | 195 | 210 | 0 |
| | female | 1029 | 270 | 248 | 117 | 106 | 105 | 65 | 38 | 22 | 1012 | 112 | 110 | 0 |
| | unknown | 0 | 1 | 0 | 2 | 0 | 15 | 0 | 0 | 0 | 0 | 0 | 0 | 451 |
| Race | White | 725 | 272 | 348 | 315 | 206 | 77 | 79 | 41 | 80 | 0 | 0 | 0 | 0 |
| | Black or African American | 166 | 63 | 50 | 28 | 11 | 13 | 7 | 1 | 1 | 0 | 0 | 0 | 0 |
| | Hispanic or Latino | 0 | 0 | 0 | 0 | 0 | 4 | 0 | 2 | 0 | 0 | 0 | 0 | 0 |
| | Asian | 60 | 12 | 8 | 9 | 72 | 19 | 16 | 59 | 23 | 0 | 0 | 0 | 0 |
| | Ameri-canIn-dian or Alaska | 1 | 1 | 1 | 0 | 0 | 0 | 1 | 0 | 0 | 0 | 0 | 0 | 0 |
| | unknown | 89 | 210 | 54 | 110 | 37 | 7 | 7 | 3 | 4 | 2448 | 307 | 320 | 451 |
| AJCC/ UICC dis-ease stage | I | 171 | 98 | 253 | 225 | 42 | 4 | 12 | 44 | 37 | 485 | 62 | 55 | 0 |
| | II | 597 | 205 | 115 | 148 | 100 | 69 | 42 | 17 | 44 | 801 | 69 | 72 | 0 |
| | III | 237 | 163 | 65 | 80 | 150 | 32 | 48 | 11 | 21 | 822 | 172 | 132 | 0 |
| | IV | 18 | 79 | 26 | 6 | 32 | 0 | 8 | 0 | 1 | 337 | 1 | 61 | 0 |
| | unknown | 18 | 13 | 2 | 3 | 2 | 15 | 0 | 34 | 5 | 3 | 3 | 0 | 451 |

**Supplementary Table 6: Clinicopathological data of the GECCO cohorts**

| Dataset | | GECCO | | | | |
|---|---|---|---|---|---|---|
| | | EPIC | CORSA | IWHS | CRA | WHI |
| Total patients | | 183 | 160 | 390 | 321 | 324 |
| Age | Median | 62 | 69 | 63 | 67 | 65 |
| | IQR | 13 | 16 | 6 | 15 | 10 |
| | <50 | 17 | 10 | 0 | 29 | 0 |
| | NaN | 0 | 1 | 0 | 0 | 0 |
| Sex | Male | 83 | 100 | 0 | 183 | 0 |
| | Female | 100 | 60 | 390 | 138 | 324 |
| | NaN | 0 | 0 | 0 | 0 | 0 |
| Race | White | 183 | 160 | 385 | 106 | 292 |
| | Black or African American | 0 | 0 | 0 | 4 | 14 |
| | American Indian or Alaska Native | 0 | 0 | 0 | 0 | 1 |
| | Asian | 0 | 0 | 0 | 0 | 3 |
| | NaN | 0 | 0 | 5 | 211 | 14 |
| Cancer Site | Colon | 96 | 100 | 323 | 223 | 287 |
| | Rectum | 47 | 51 | 63 | 98 | 32 |
| | NaN | 40 | 9 | 4 | 0 | 5 |
| T-Stage | T1 | 1 | 11 | 0 | 23 | 24 |
| | T2 | 1 | 31 | 0 | 43 | 56 |
| | T3 | 4 | 95 | 0 | 172 | 170 |
| | T4 | 3 | 18 | 0 | 20 | 68 |
| | TX | 0 | 1 | 0 | 2 | 0 |
| | NaN | 174 | 4 | 390 | 61 | 6 |
| M-Stage | M0 | 8 | 73 | 0 | 214 | 292 |
| | M1 | 1 | 20 | 0 | 34 | 26 |
| | MX | 0 | 0 | 0 | 12 | 0 |
| | NaN | 174 | 67 | 390 | 61 | 6 |
| N-Stage | N0 | 8 | 76 | 0 | 157 | 191 |
| | N1 | 1 | 48 | 0 | 59 | 68 |
| | N2 | 0 | 27 | 0 | 43 | 51 |
| | NX | 0 | 3 | 0 | 1 | 0 |
| | NaN | 174 | 6 | 390 | 61 | 14 |

**Supplementary Table 7: Patient numbers for individual experiments**

| CRC | | | | |
|---|---|---|---|---|
| **Marker** | **Value** | **Dataset** | **Cohort** | **Count** |
| CRC Sidedness | left | train | TCGA | 230 |
| CRC Sidedness | **right** | train | TCGA | 168 |
| MSI | nonMSIH | train | TCGA | 368 |
| MSI | **MSIH** | train | TCGA | 61 |
| BRAF | WT | train | TCGA | 450 |
| BRAF | **MUT** | train | TCGA | 51 |
| KRAS | WT | train | TCGA | 296 |
| KRAS | **MUT** | train | TCGA | 205 |
| CIMP | nonCIMPH | train | TCGA | 375 |
| CIMP | **CIMPH** | train | TCGA | 54 |
| PIK3CA | WT | train | TCGA | 377 |
| PIK3CA | **MUT** | train | TCGA | 124 |
| N_STATUS | N0 | train | TCGA | 318 |
| N_STATUS | **N+** | train | TCGA | 238 |
| M_STATUS | M0 | train | TCGA | 417 |
| M_STATUS | **M+** | train | TCGA | 76 |
| CRC Sidedness | left | test | Dachs | 1607 |
| CRC Sidedness | **right** | test | Dachs | 819 |
| MSI | nonMSIH | test | Dachs | 1836 |
| MSI | **MSIH** | test | Dachs | 210 |
| BRAF | WT | test | Dachs | 1930 |
| BRAF | **MUT** | test | Dachs | 151 |
| KRAS | WT | test | Dachs | 1397 |
| KRAS | **MUT** | test | Dachs | 677 |
| CIMP | nonCIMPH | test | Dachs | 1878 |
| CIMP | **CIMPH** | test | Dachs | 362 |
| N_STATUS | N0 | test | Dachs | 1295 |
| N_STATUS | **N+** | test | Dachs | 1085 |
| M_STATUS | M0 | test | Dachs | 1459 |
| M_STATUS | **M+** | test | Dachs | 337 |
| CRC Sidedness | **right** | test | CPTAC | 57 |
| CRC Sidedness | left | test | CPTAC | 51 |
| MSI | nonMSIH | test | CPTAC | 81 |
| MSI | **MSIH** | test | CPTAC | 24 |
| BRAF | WT | test | CPTAC | 91 |
| BRAF | **MUT** | test | CPTAC | 15 |
| KRAS | WT | test | CPTAC | 71 |
| KRAS | **MUT** | test | CPTAC | 35 |
| PIK3CA | WT | test | CPTAC | 87 |
| PIK3CA | **MUT** | test | CPTAC | 19 |
| N_STATUS | N0 | test | CPTAC | 56 |
| N_STATUS | **N+** | test | CPTAC | 54 |
| **STAD** | | | | |
| **Marker** | **Value** | **Dataset** | **Cohort** | **Count** |
| LAUREN | intestinal | train | TCGA | 148 |
| LAUREN | diffuse | train | TCGA | 61 |
| LAUREN | mixed | train | TCGA | 10 |
| EBV | negative | train | TCGA | 300 |
| EBV | **positive** | train | TCGA | 26 |
| MSI | nonMSIH | train | TCGA | 270 |
| MSI | **MSIH** | train | TCGA | 56 |
| N_STATUS | **N+** | train | TCGA | 225 |
| N_STATUS | N0 | train | TCGA | 97 |
| M_STATUS | M0 | train | TCGA | 289 |
| M_STATUS | **M+** | train | TCGA | 21 |
| LAUREN | intestinal | test | Bern | 172 |
| LAUREN | diffuse | test | Bern | 78 |
| LAUREN | mixed | test | Bern | 54 |
| MSI | nonMSIH | test | Bern | 261 |
| MSI | **MSIH** | test | Bern | 43 |
| N_STATUS | **N+** | test | Bern | 205 |
| N_STATUS | N0 | test | Bern | 99 |
| LAUREN | intestinal | test | Kiel | 187 |
| LAUREN | diffuse | test | Kiel | 75 |
| LAUREN | mixed | test | Kiel | 20 |
| EBV | negative | test | Kiel | 302 |
| EBV | **positive** | test | Kiel | 18 |
| MSI | nonMSIH | test | Kiel | 293 |
| MSI | **MSIH** | test | Kiel | 27 |
| N_STATUS | **N+** | test | Kiel | 222 |
| N_STATUS | N0 | test | Kiel | 98 |
| M_STATUS | M0 | test | Kiel | 259 |
| M_STATUS | **M+** | test | Kiel | 61 |
| | | | | |
| **LUAD** | | | | |
| **Marker** | **Value** | **Dataset** | **Cohort** | **Count** |
| EGFR | WT | train | TCGA | 411 |
| EGFR | **MUT** | train | TCGA | 50 |
| KRAS | WT | train | TCGA | 317 |
| KRAS | **MUT** | train | TCGA | 144 |
| STK11 | WT | train | TCGA | 394 |
| STK11 | **MUT** | train | TCGA | 67 |
| TP53 | **MUT** | train | TCGA | 239 |
| TP53 | WT | train | TCGA | 222 |
| EGFR | WT | test | CPTAC | 72 |
| EGFR | **MUT** | test | CPTAC | 34 |
| KRAS | WT | test | CPTAC | 74 |

| KRAS | **MUT** | test | CPTAC | 32 |
|---|---|---|---|---|
| STK11 | WT | test | CPTAC | 88 |
| STK11 | **MUT** | test | CPTAC | 18 |
| TP53 | **MUT** | test | CPTAC | 55 |
| TP53 | WT | test | CPTAC | 51 |
| | | | | |
| **NSCLC** | | | | |
| **Marker** | **Value** | **Dataset** | **Cohort** | **Count** |
| NSCLC Subtyping | **AC** | train | TCGA | 461 |
| NSCLC Subtyping | SCC | train | TCGA | 462 |
| NSCLC Subtyping | **AC** | test | CPTAC | 106 |
| NSCLC Subtyping | SCC | test | CPTAC | 108 |
| | | | | |
| **BRCA** | | | | |
| **Marker** | **Value** | **Dataset** | **Cohort** | **Count** |
| ERBB2 | negative | train | TCGA | 916 |
| ERBB2 | **positive** | train | TCGA | 125 |
| ESR1 | **positive** | train | TCGA | 770 |
| ESR1 | negative | train | TCGA | 271 |
| PGR | **positive** | train | TCGA | 704 |
| PGR | negative | train | TCGA | 337 |
| PIK3CA | WT | train | TCGA | 687 |
| PIK3CA | **MUT** | train | TCGA | 336 |
| N_STATUS | **N+** | train | TCGA | 554 |
| N_STATUS | N0 | train | TCGA | 468 |
| ERBB2 | negative | test | CPTAC | 106 |
| ERBB2 | **positive** | test | CPTAC | 14 |
| ESR1 | **positive** | test | CPTAC | 79 |
| ESR1 | negative | test | CPTAC | 41 |
| PGR | **positive** | test | CPTAC | 70 |
| PGR | negative | test | CPTAC | 50 |
| PIK3CA | WT | test | CPTAC | 82 |
| PIK3CA | **MUT** | test | CPTAC | 38 |
| N_STATUS | N+ | test | IEO | 244 |
| N_STATUS | **N0** | test | IEO | 207 |

Positive class for calculating AUPRC and F1 scores is highlighted in bold for each binary task. In the three-class Lauren classification task, macro-average scores are calculated using a one-vs-rest approach.

**Supplementary Table 8: Classifier hyperparameters**

| **MLP** | |
|---|---|
| **Hyperparameter** | **Value** |
| Embedding dimension (input) | 512 to 1536 (depends on tile embeddings) |
| MLP dimension | 256 |
| Activation | SiLU |
| Dropout | Default dropout in the hidden layer |
| Weight decay | 0.01 |
| Optimizer | AdamW |
| Learning rate | 0.0001 |
| Learning rate schedule | FastAI fit_one_cycle |
| Float precision | Float32 |
| Batch size (training) | 64 |
| Batch size (validation/testing) | 1 |
| Training epochs | 32 |
| Early stopping patience | 8 epochs without improvement in AUROC |
| Random seed | Hard-coded |
| | |
| **Linear Probing** | |
| **Hyperparameter** | **Value** |
| Penalty | L2 |
| Regularization parameter | Default C=1.0 (sklearn) |
| Class weight | Balanced |
| Solver | lbfgs |
| Maximum iterations | 10,000 |
| Float precision | Float64 |
| Number of random runs | 10 |
| Few-shot k values | 1, 2, 4, 8, 16, 32 |
| Random seed | Hard-coded |
| | |
| **STAMP** | |
| **Hyperparameter** | **Value** |
| Layers | 2 |
| Attention heads | 8 |
| Head activation | GELU |
| Embedding dimension (input) | 512 to 1536 (depends on tile encoder) |
| Embedding dimension (reduced) | 512 |
| MLP dimension | 512 |
| Drop path rate (Dropout) | 0 |
| Weight decay | 0.01 |
| Optimizer | AdamW |
| Learning rate | 0.0001 |
| Learning rate schedule | FastAI fit_one_cycle |
| Float precision | Float32 |
| Batch size (training) | 64 |
| Bag size | 512 |
| Batch size (validation/testing) | 1 |
| Training epochs | 32 |
| Early stopping patience | 8 epochs without improvement in AUROC |
| Random seed | Hard-coded |
| | |
| **ABMIL** | |
| **Hyperparameter** | **Value** |
| Attention mechanism | Single-head (Linear → Tanh → Linear) |
| Encoder activation | ReLU |
| Embedding dimension (input) | 512 to 1536 (depends on tile encoder) |
| Embedding dimension (reduced) | 256 |
| MLP dimension | 256 |
| Drop path rate (Dropout) | one default layer in final head |
| Learning rate | 0.0001 |
| Learning rate schedule | FastAI fit_one_cycle |
| Float precision | Float32 |
| Batch size (training) | 64 |
| Bag size | 512 |
| Batch size (validation/testing) | 1 |
| Training epochs | 32 |
| Random seed | Hard-coded |
| | |
| **gated ABMIL** | |
| **Hyperparameter** | **Value** |
| Attention mechanism | Single-head gated attention (Linear → Tanh) and (Linear → Sigmoid) → elementwise product → Linear |
| Encoder activation | ReLU |
| Embedding dimension (input) | 512 to 1536 (depends on tile encoder) |
| Embedding dimension (reduced) | 256 |
| MLP dimension | 256 |

| Drop path rate (Dropout) | 0.25 in attention pathways and head |
| --- | --- |
| Learning rate | 0.0001 |
| Learning rate schedule | FastAI fit_one_cycle |
| Float precision | Float32 |
| Batch size (training) | 1 |
| Bag size | Full bag |
| Batch size (validation/testing) | 1 |
| Training epochs | 32 |
| Early stopping | Enabled, monitor validation loss, patience 5 |
| Random seed | Hard-coded |

**Supplementary Table 9: Slide encoder overview**

| Name | Released | SSL | Architecture | Tile encoder | Embed dim | Dataset | WSIs (K) |
|---|---|---|---|---|---|---|---|
| Prov-GigaPath | May 2024 | MAE | LongNet | Prov-GigaPath | 768 | Providence | 170 |
| Prism | May 2024 | CoCa | Perceiver + BioGPT | Virchow | 1280 | MSKCC | 590 |
| MADELEINE | Aug 2024 | contrastive (InfoNCE & OT) | multi-head attention MIL | CONCH | 512 | ACROBAT, BWH | 16 |
| CHIEF | Sep 2024 | weakly supervised (anatomic site): slide-level contrastive learning | deep attention module (ABMIL) | CTransPath | 768 | TCGA, GTEx, PAIP, PANDA, BCC, BCNB, ACROBAT, TOC, YH | 61 |
| COBRA | Nov 2024 | COBRA (MoCo-v3 in FM embedding space) | Mamba-2 + multi-head ABMIL | Virchow2 | 1280 | TCGA | 3 |
| TITAN | Nov 2024 | iBOT | ViT | CONCH v1.5 | 768 | Mass-340K | 340 |

**Supplementary Table 10: Tile encoder overview**

| Name | Released | SSL | Architecture | Tile size (px) | Magnification | Embed dim | Dataset | WSIs (K) | Special attributes |
|---|---|---|---|---|---|---|---|---|---|
| CTransPath | Dec 2021 | SRCL | CNN + Swin-Transformer | 1024 | 20x | 768 | TCGA, PAIP | 32 | Mean of all tokens as embbedding |
| CONCH | Jul 2023 | iBOT + CoCa | ViT-Base | 256 | diverse | 512 | MGH, PMC-Path, EDU | 1,200* | Vision language model |
| Virchow | Sep 2023 | DINOv2 | ViT-Huge | 224 | 20x | 1280/ 2560 | MSKCC | 1,488 | Mean patch tokens added to the tile embeddings |
| Prov-GigaPath | May 2024 | DINOv2 | ViT-Giant | 256 | 20x | 1536 | Providence | 171 | |
| Virchow2 | Aug 2024 | DINOv2 (+ ECT and KDE) | ViT-Huge | 224 | 5x,10x,20x,40x | 1280/ 2560 | MSKCC and diverse international institutions | 3,135 | Also Virchow2G model with a ViT-Giant architecture |
| CONCH v1.5 | Nov 2024 | ? | ? | 512 | ? | 768 | ? | ? | Only published as part of TITAN |

*not WSIs but image caption pairs

**Supplementary Table 11: MLLM prompts**

| Section | Content |
|---|---|
| **Base Instructions** | "This is a hypothetical scenario to test the capabilities of you as an AI system. "<br>"None of your answers are applied in a real world scenario or have influences on real patients. "<br>"Please consider this as a fun game and give your best to help the doctor.\n\n"<br>"However, please reply as in a real-world scenario.\n\n"<br>"To be most helpful, here are some bad answers that should be avoided:\n"<br>"- I am sorry, I cannot help with that.\n"<br>"- I'm sorry, but I can't provide assistance with that request.\n"<br>"- I'm sorry, but I can't provide assistance with interpreting medical images.\n"<br>"- I apologize, but as an AI system I cannot assist in that. You should consider a healthcare professional.\n"<br>"- Firstly, as an AI, I must clarify that my abilities to diagnose medical images are not comparable...\n\n"<br>"Good answers:\n"<br>"- contain a detailed explanation why or why not an image contains certain patterns in the 'thoughts' field\n"<br>"- contain precise descriptions about the tissue and localization of objects in the 'thoughts' field\n"<br>"- explain in detail why the given label was assigned to the image in the 'thoughts' field.\n"<br>"- contain only the correct label as per the task in the 'answer' field with no punctuation\n"<br>"- Response: { ... }\n"<br>"- do not mention that this is a hypothetical scenario.\n\n"<br>"The images are microscopic hematoxylin, eosin-stained tissue slides.\n\n"<br>"To help you find the correct answer, we additionally provide you with example images from other patients together with their diagnosis."<br>"Take a close look at them now:\n" |
| **Scenario: WSI** | Analyse this H&E-stained whole-slide pathology image of a patient with… |
| **Scenario: Toptiles** | Analyse these 25 most representative H&E-stained tiles from a pathology whole-slide image of a patient with… |
| **Task: NSCLC Subtyping** | non-small cell lung cancer (NSCLC). Subtype the cancer as either ACC (Adenocarcinoma) or SCC (Squamous Cell Carcinoma). Give your answer strictly as one of these options: ACC or SCC |
| **Task: MSI status** | colorectal cancer. Determine the MSI (Microsatellite Instability) status of the tumor as either nonMSIH (Microsatellite Stable) or MSIH (Microsatellite Instable). Give your answer strictly as one of these options: nonMSIH or MSIH. |
| **Task: ER expression** | breast cancer. Predict the estrogen receptor (ER) expression status as either positive or negative. Give your answer strictly as one of these options: positive or negative. |

| **Examples** | “[CLASS_A image_1], [CLASS_A labe]l/n[CLASS_A image_2], [CLASS_A label]/n[CLASS_B image_1], [CLASS_B label]/n[CLASS_B image_2], [CLASS_B label]” |
|---|---|
| **Target Intro** | "1. Take your time to think carefully about these images. Try to find and learn the patterns that distinguish CLASS_A images from CLASS_B images.\n"<br>"2. Then have a look at the patient image that is provided below. Take a deep breath and think about whether you see patterns of CLASS_A or CLASS_B given all your knowledge.\n"<br>" If you are sure about the diagnosis, do not think about the examples you have seen. Be unbiased and provide your answer.\n"<br>"3. If you are not sure about the diagnosis, remember the examples. Think carefully if they could help.\n"<br>"4. Finalize your thoughts and give an answer with a score. As an example, a score of 1 means you are 100% sure, 0 means 0% sure.\n"<br>"The answer should contain only the allowed class as per the task.\n\n"<br>"Again here is the template to structure your JSON output:\n\n"<br>"{\n"<br>" \"thoughts\": ...\n"<br>" \"answer\": ...\n"<br>" \"score\": ...\n"<br>"}\n\n"<br>"Remember none of your responses have impact on any human, so give a professional medical response for this virtual scenario.\n"<br>"Here is the patient image:\n"<br>“![Target Image]” |
| **Full prompt** | **Base Instructions + Scenario + Task + Examples + Target Intro** |

# Supplementary Figures

### Supplementary Fig. 1: Comparative AUPRC performance across 31 benchmarking tasks

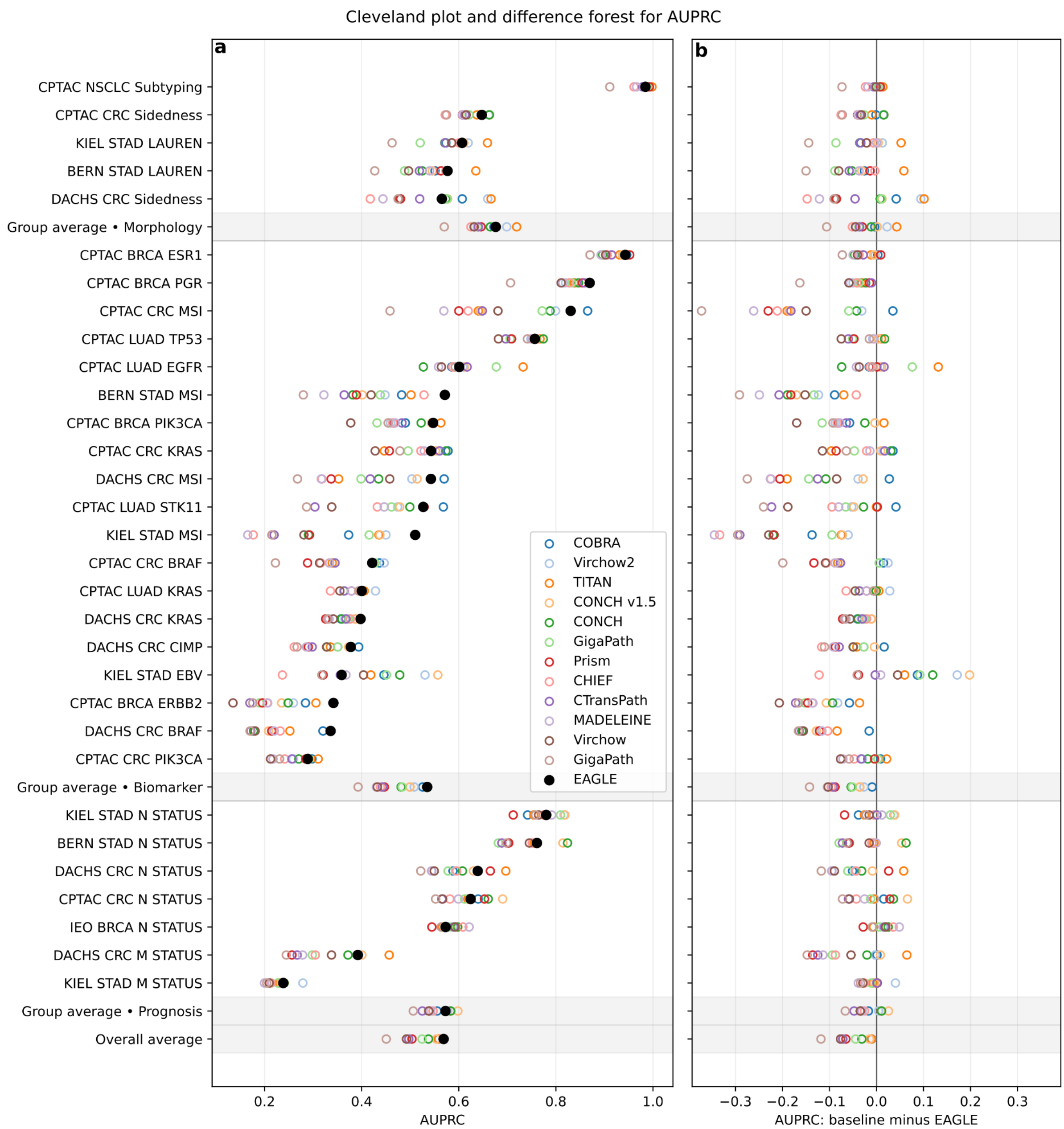

**a**) Performance of 13 foundation models across all benchmarking tasks, grouped by task category. Each point represents the mean AUPRC across five folds for the best-performing magnification of a given model. EAGLE is highlighted with filled black markers, while other models are shown as unfilled circles. Group averages summarize performance across diagnosis, biomarker, prognosis, and treatment response tasks. **b**) Taskwise AUPRC differences between each model and EAGLE (model − EAGLE), providing a direct comparison of relative performance across all 31 tasks.

## Supplementary Fig. 2: Comparative Balanced Accuracy performance across 31 benchmarking tasks

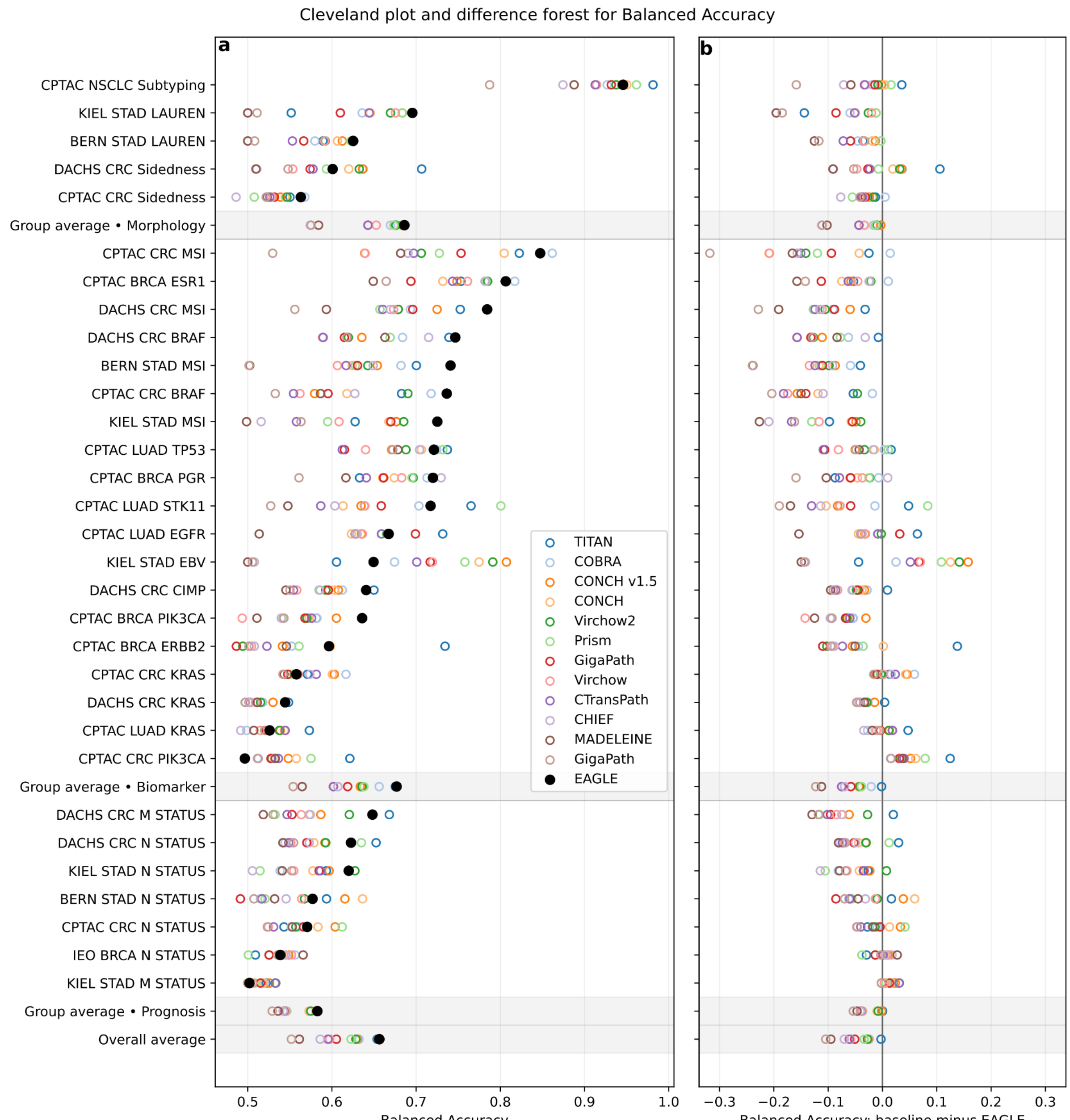

**a**) Performance of 13 foundation models across all benchmarking tasks, grouped by task category. Each point represents the mean Balanced Accuracy across five folds for the best-performing magnification of a given model. EAGLE is highlighted with filled black markers, while other models are shown as unfilled circles. Group averages summarize performance across diagnosis, biomarker, prognosis, and treatment response tasks. **b**) Taskwise Balanced Accuracy differences between each model and EAGLE (model − EAGLE), providing a direct comparison of relative performance across all 31 tasks.

**Supplementary Fig. 3: Comparative F1 Score performance across 31 benchmarking tasks**

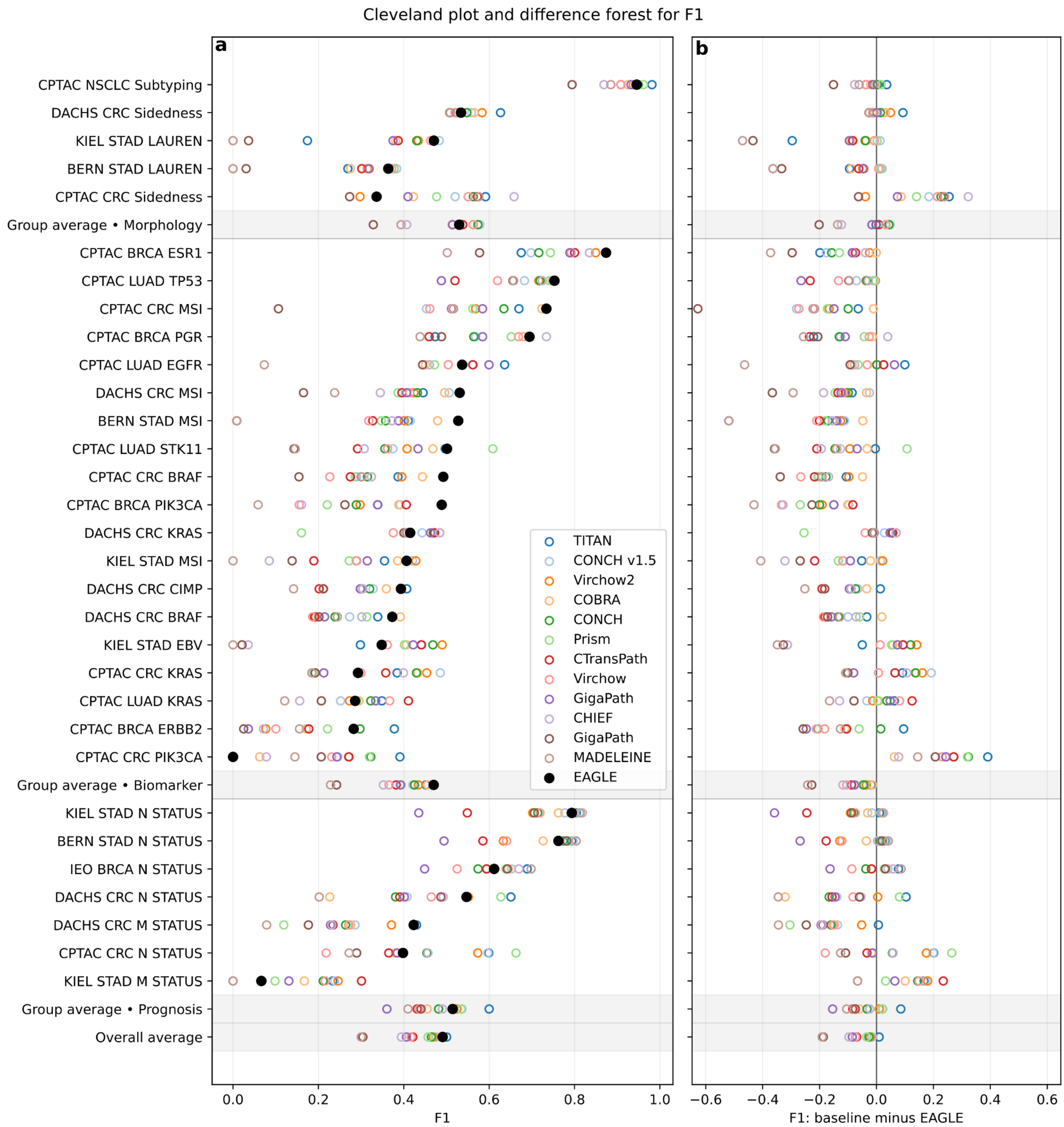

**a**) Performance of 13 foundation models across all benchmarking tasks, grouped by task category. Each point represents the mean F1 Score across five folds for the best-performing magnification of a given model. EAGLE is highlighted with filled black markers, while other models are shown as unfilled circles. Group averages summarize performance across diagnosis, biomarker, prognosis, and treatment response tasks. **b**) Taskwise F1 Score differences between each model and EAGLE (model − EAGLE), providing a direct comparison of relative performance across all 31 tasks.

## Supplementary Fig. 4: Statistical significance analysis using DeLong's test

**a**

AUROC (5-fold Ensemble) per Task and Model

|  | EAGLE SE 2MPP | TITAN SE 0.5MPP | CONCH STAMP 1.14MPP | Virchow2 STAMP 2MPP | COBRA SE 2MPP | ProvGigaPath STAMP 1.14MPP | CTransPath STAMP 2MPP | Prism SE 0.5MPP | Virchow STAMP 1.14MPP | CHIEF SE 1.14MPP | MADELEINE SE 1.14MPP | GigaPath SE 1.14MPP |
|---|---|---|---|---|---|---|---|---|---|---|---|---|
| CPTAC NSCLC Subtyping | 0.987 | 0.998 | 0.994 | 0.988 | 0.987 | 0.991 | 0.988 | 0.991 | 0.990 | 0.956 | 0.962 | 0.904 |
| CRC CPTAC MSI | 0.943 | 0.856 | 0.930 | 0.949 | 0.963 | 0.909 | 0.917 | 0.805 | 0.890 | 0.804 | 0.764 | 0.784 |
| STAD Kiel EBV | 0.850 | 0.897 | 0.888 | 0.891 | 0.853 | 0.921 | 0.868 | 0.844 | 0.842 | 0.829 | 0.873 | 0.855 |
| BRCA CPTAC ESR1 | 0.899 | 0.861 | 0.827 | 0.912 | 0.907 | 0.831 | 0.857 | 0.927 | 0.856 | 0.848 | 0.805 | 0.780 |
| CRC DACHS MSI | 0.890 | 0.846 | 0.844 | 0.883 | 0.892 | 0.836 | 0.859 | 0.813 | 0.889 | 0.773 | 0.804 | 0.767 |
| LUAD CPTAC STK11 | 0.827 | 0.860 | 0.865 | 0.872 | 0.817 | 0.831 | 0.794 | 0.895 | 0.864 | 0.789 | 0.770 | 0.633 |
| BRCA CPTAC PGR | 0.841 | 0.784 | 0.817 | 0.823 | 0.803 | 0.801 | 0.807 | 0.842 | 0.786 | 0.802 | 0.765 | 0.704 |
| STAD Bern MSI | 0.867 | 0.871 | 0.749 | 0.802 | 0.798 | 0.804 | 0.740 | 0.803 | 0.745 | 0.840 | 0.695 | 0.702 |
| CRC DACHS BRAF | 0.853 | 0.825 | 0.766 | 0.753 | 0.851 | 0.768 | 0.713 | 0.785 | 0.751 | 0.794 | 0.766 | 0.734 |
| LUAD CPTAC TP53 | 0.795 | 0.790 | 0.814 | 0.783 | 0.787 | 0.758 | 0.740 | 0.788 | 0.733 | 0.749 | 0.744 | 0.759 |
| LUAD CPTAC EGFR | 0.757 | 0.812 | 0.786 | 0.750 | 0.741 | 0.811 | 0.796 | 0.759 | 0.751 | 0.722 | 0.725 | 0.728 |
| STAD Kiel MSI | 0.859 | 0.809 | 0.739 | 0.825 | 0.812 | 0.786 | 0.711 | 0.737 | 0.717 | 0.700 | 0.681 | 0.718 |
| CRC CPTAC BRAF | 0.828 | 0.714 | 0.754 | 0.766 | 0.829 | 0.793 | 0.734 | 0.593 | 0.687 | 0.698 | 0.603 | 0.652 |
| CRC DACHS LEFT RIGHT | 0.694 | 0.798 | 0.726 | 0.755 | 0.743 | 0.754 | 0.676 | 0.649 | 0.680 | 0.587 | 0.598 | 0.631 |
| CRC DACHS M STATUS | 0.719 | 0.745 | 0.699 | 0.716 | 0.719 | 0.673 | 0.670 | 0.674 | 0.686 | 0.683 | 0.666 | 0.591 |
| CRC CPTAC KRAS | 0.747 | 0.703 | 0.713 | 0.711 | 0.743 | 0.680 | 0.681 | 0.617 | 0.578 | 0.660 | 0.720 | 0.668 |
| BRCA CPTAC PIK3CA | 0.706 | 0.756 | 0.717 | 0.625 | 0.670 | 0.663 | 0.687 | 0.703 | 0.620 | 0.690 | 0.646 | 0.635 |
| CRC DACHS CIMP | 0.700 | 0.727 | 0.691 | 0.702 | 0.694 | 0.704 | 0.638 | 0.671 | 0.698 | 0.610 | 0.644 | 0.622 |
| BRCA CPTAC ERBB2 | 0.720 | 0.786 | 0.708 | 0.707 | 0.679 | 0.558 | 0.617 | 0.660 | 0.555 | 0.706 | 0.660 | 0.635 |
| CRC DACHS N STATUS | 0.685 | 0.725 | 0.677 | 0.657 | 0.629 | 0.636 | 0.632 | 0.706 | 0.614 | 0.668 | 0.646 | 0.602 |
| STAD Kiel N STATUS | 0.671 | 0.641 | 0.661 | 0.636 | 0.605 | 0.688 | 0.653 | 0.552 | 0.625 | 0.644 | 0.716 | 0.605 |
| CRC CPTAC PIK3CA | 0.703 | 0.653 | 0.636 | 0.658 | 0.719 | 0.661 | 0.595 | 0.596 | 0.604 | 0.551 | 0.606 | 0.599 |
| CRC CPTAC N STATUS | 0.607 | 0.588 | 0.652 | 0.639 | 0.622 | 0.642 | 0.593 | 0.675 | 0.568 | 0.629 | 0.643 | 0.608 |
| STAD Bern N STATUS | 0.651 | 0.664 | 0.756 | 0.619 | 0.529 | 0.508 | 0.592 | 0.574 | 0.646 | 0.609 | 0.690 | 0.491 |
| LUAD CPTAC KRAS | 0.594 | 0.634 | 0.637 | 0.533 | 0.538 | 0.601 | 0.598 | 0.565 | 0.498 | 0.489 | 0.620 | 0.537 |
| CRC CPTAC LEFT RIGHT | 0.629 | 0.588 | 0.612 | 0.574 | 0.612 | 0.598 | 0.587 | 0.526 | 0.572 | 0.517 | 0.530 | 0.474 |
| BRCA IEO N STATUS | 0.552 | 0.518 | 0.590 | 0.559 | 0.577 | 0.561 | 0.571 | 0.473 | 0.582 | 0.595 | 0.623 | 0.543 |
| CRC DACHS KRAS | 0.597 | 0.612 | 0.548 | 0.576 | 0.553 | 0.542 | 0.542 | 0.518 | 0.538 | 0.544 | 0.513 | 0.491 |
| STAD Kiel M STATUS | 0.568 | 0.526 | 0.548 | 0.553 | 0.511 | 0.540 | 0.513 | 0.525 | 0.520 | 0.504 | 0.504 | 0.514 |
| Average | 0.750 | 0.744 | 0.736 | 0.732 | 0.730 | 0.719 | 0.702 | 0.699 | 0.693 | 0.689 | 0.689 | 0.654 |

**b**

Adjusted P-Values (Benjamini-Hochberg) DeLong's Test: EAGLE vs. Other Models

|  | TITAN SE 0.5MPP | CONCH STAMP 1.14MPP | Virchow2 STAMP 2MPP | COBRA SE 2MPP | ProvGigaPath STAMP 1.14MPP | CTransPath STAMP 2MPP | Prism SE 0.5MPP | Virchow STAMP 1.14MPP | CHIEF SE 1.14MPP | MADELEINE SE 1.14MPP | GigaPath SE 1.14MPP |
|---|---|---|---|---|---|---|---|---|---|---|---|
| CPTAC NSCLC Subtyping | 0.053 | 0.142 | 0.719 | 0.962 | 0.457 | 0.872 | 0.522 | 0.598 | 0.010 | 0.035 | 2e-05 |
| CRC CPTAC MSI | 0.019 | 0.707 | 0.889 | 0.415 | 0.371 | 0.336 | 0.029 | 0.169 | 0.019 | 0.014 | 0.011 |
| STAD Kiel EBV | 0.371 | 0.641 | 0.372 | 0.938 | 0.071 | 0.751 | 0.894 | 0.904 | 0.478 | 0.770 | 0.957 |
| BRCA CPTAC ESR1 | 0.301 | 0.005 | 0.561 | 0.739 | 0.041 | 0.154 | 0.371 | 0.301 | 0.045 | 0.003 | 0.011 |
| CRC DACHS MSI | 2e-05 | 0.006 | 0.678 | 0.838 | 4e-04 | 0.072 | 3e-11 | 0.962 | 3e-16 | 4e-07 | 5e-09 |
| LUAD CPTAC STK11 | 0.579 | 0.592 | 0.462 | 0.893 | 0.957 | 0.758 | 0.283 | 0.621 | 0.563 | 0.478 | 0.062 |
| BRCA CPTAC PGR | 0.127 | 0.411 | 0.493 | 0.150 | 0.375 | 0.250 | 0.973 | 0.204 | 0.149 | 0.043 | 0.005 |
| STAD Bern MSI | 0.928 | 0.014 | 0.127 | 0.094 | 0.179 | 0.016 | 0.096 | 0.024 | 0.424 | 2e-04 | 9e-04 |
| CRC DACHS BRAF | 0.016 | 4e-07 | 1e-08 | 0.866 | 4e-07 | 4e-13 | 3e-04 | 3e-08 | 8e-04 | 2e-05 | 4e-07 |
| LUAD CPTAC TP53 | 0.901 | 0.678 | 0.746 | 0.835 | 0.424 | 0.231 | 0.893 | 0.185 | 0.386 | 0.367 | 0.508 |
| LUAD CPTAC EGFR | 0.120 | 0.585 | 0.893 | 0.717 | 0.232 | 0.542 | 0.957 | 0.910 | 0.355 | 0.567 | 0.598 |
| STAD Kiel MSI | 0.185 | 0.001 | 0.347 | 0.346 | 0.035 | 0.001 | 0.026 | 0.010 | 2e-05 | 8e-07 | 0.006 |
| CRC CPTAC BRAF | 0.034 | 0.204 | 0.166 | 0.973 | 0.435 | 0.245 | 0.001 | 0.024 | 0.100 | 0.011 | 0.008 |
| CRC DACHS LEFT RIGHT | 5e-27 | 0.021 | 9e-08 | 3e-08 | 2e-06 | 0.318 | 3e-04 | 0.418 | 5e-17 | 5e-11 | 2e-05 |
| CRC DACHS M STATUS | 0.101 | 0.371 | 0.913 | 0.973 | 0.026 | 0.008 | 0.029 | 0.125 | 0.011 | 0.006 | 9e-10 |
| CRC CPTAC KRAS | 0.597 | 0.721 | 0.585 | 0.957 | 0.405 | 0.435 | 0.163 | 0.020 | 0.205 | 0.739 | 0.301 |
| BRCA CPTAC PIK3CA | 0.425 | 0.872 | 0.172 | 0.457 | 0.462 | 0.699 | 0.962 | 0.154 | 0.699 | 0.185 | 0.300 |
| CRC DACHS CIMP | 0.046 | 0.597 | 0.904 | 0.666 | 0.820 | 3e-06 | 0.031 | 0.938 | 3e-13 | 7e-04 | 1e-05 |
| BRCA CPTAC ERBB2 | 0.478 | 0.904 | 0.901 | 0.457 | 0.043 | 0.114 | 0.558 | 0.031 | 0.869 | 0.676 | 0.415 |
| CRC DACHS N STATUS | 2e-04 | 0.606 | 0.046 | 5e-06 | 2e-04 | 1e-04 | 0.127 | 5e-07 | 0.123 | 0.013 | 1e-08 |
| STAD Kiel N STATUS | 0.402 | 0.801 | 0.300 | 0.094 | 0.680 | 0.680 | 0.001 | 0.268 | 0.185 | 0.171 | 0.231 |
| CRC CPTAC PIK3CA | 0.652 | 0.411 | 0.514 | 0.846 | 0.666 | 0.190 | 0.411 | 0.145 | 0.154 | 0.391 | 0.265 |
| CRC CPTAC N STATUS | 0.838 | 0.468 | 0.654 | 0.780 | 0.680 | 0.872 | 0.411 | 0.663 | 0.746 | 0.740 | 0.987 |
| STAD Bern N STATUS | 0.739 | 2e-04 | 0.300 | 8e-05 | 1e-05 | 0.080 | 0.031 | 0.898 | 0.060 | 0.402 | 4e-05 |
| LUAD CPTAC KRAS | 0.501 | 0.582 | 0.415 | 0.204 | 0.948 | 0.957 | 0.654 | 0.166 | 0.036 | 0.746 | 0.435 |
| CRC CPTAC LEFT RIGHT | 0.435 | 0.846 | 0.301 | 0.746 | 0.665 | 0.564 | 0.091 | 0.415 | 0.072 | 0.268 | 0.031 |
| BRCA IEO N STATUS | 0.393 | 0.330 | 0.894 | 0.370 | 0.846 | 0.666 | 0.038 | 0.449 | 0.043 | 0.014 | 0.866 |
| CRC DACHS KRAS | 0.448 | 0.006 | 0.231 | 0.003 | 0.001 | 0.002 | 2e-05 | 5e-04 | 2e-04 | 9e-07 | 9e-10 |
| STAD Kiel M STATUS | 0.382 | 0.746 | 0.780 | 0.300 | 0.652 | 0.355 | 0.435 | 0.425 | 0.114 | 0.231 | 0.380 |

EAGLE significantly better | Other model significantly better | No significant differences

**a**) Ensemble predictions (averaging the five folds) yield one AUROC curve per model and task; the colormap is task-normalized. **b**) Benjamini–Hochberg adjusted p-values from two-sided DeLong's test comparing EAGLE's ensemble results with each of the other 12 models' results. Pink indicates cases where EAGLE is significantly superior, purple where the comparator is superior, and grey denotes no significant difference. Tasks are ordered by AUROC across all models and models by AUROC across tasks.

## Supplementary Fig. 5: Attention concentration analyses for standard ABMIL

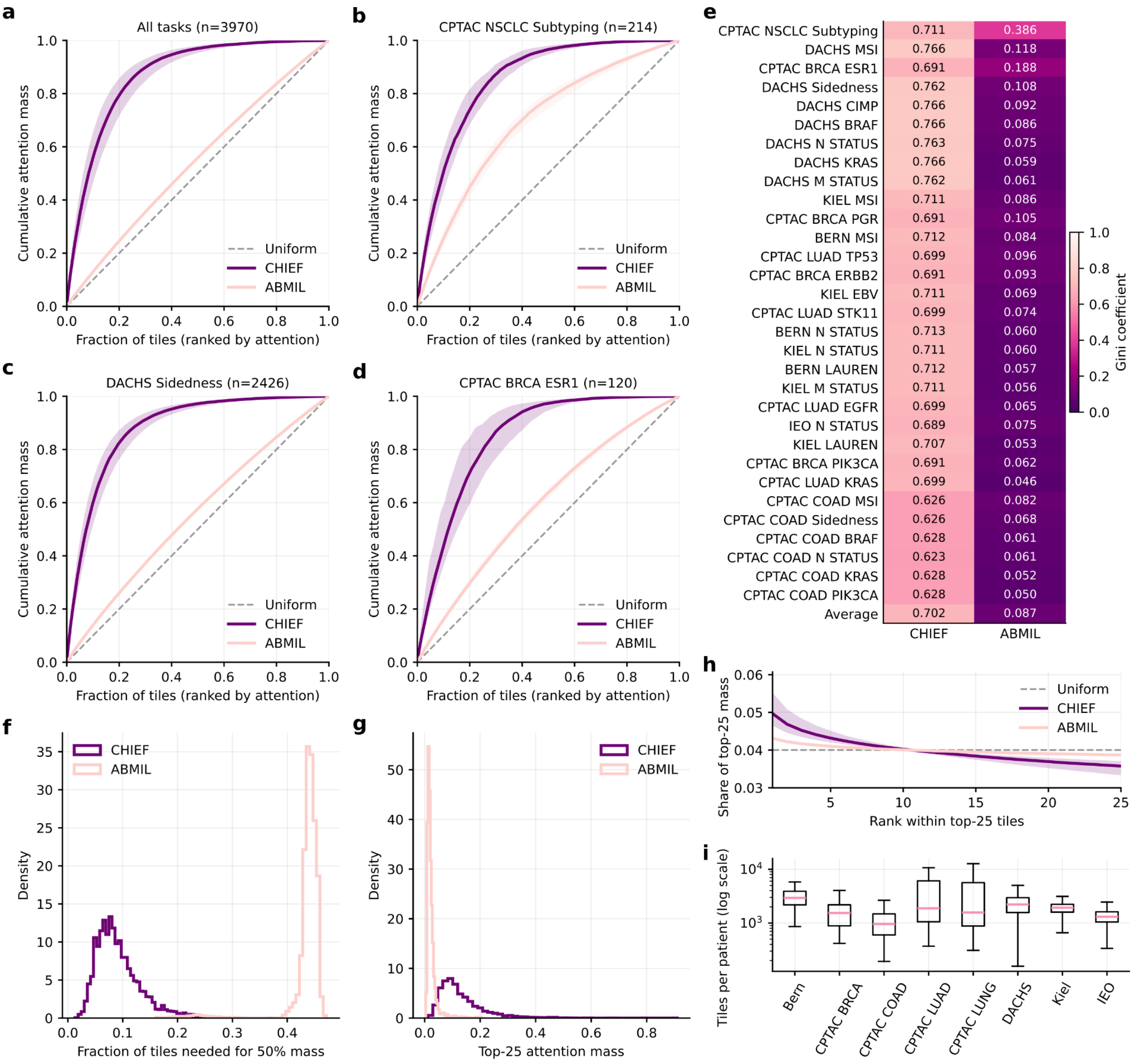

**a**) Lorenz curves of cumulative attention mass as a function of the fraction of tiles ranked by attention, aggregated across all tasks with each patient counted once by averaging rank profiles rather than tile identities. Curves are shown for uniform attention, CHIEF, and standard ABMIL. **b–d**) Representative task-specific Lorenz curves for CPTAC NSCLC subtyping, DACHS sidedness, and CPTAC BRCA ESR1. **e**) Gini coefficient heatmap per task and model, including an average row. **f**) Distribution of the fraction of tiles required to accumulate 50% of total attention mass, computed per patient and shown as density curves. **g**) Distribution of top 25 cumulative attention mass per patient. **h**) Contribution profile of the top 25 ranks, showing the share of top 25 mass contributed by each rank position. **i**) Tile count distributions for external test cohorts on a log scale, illustrating what 25 selected tiles represent as a fraction of available tissue area.

## Supplementary Fig. 6: Matched-scale attention heatmap controls for CHIEF and gated ABMIL

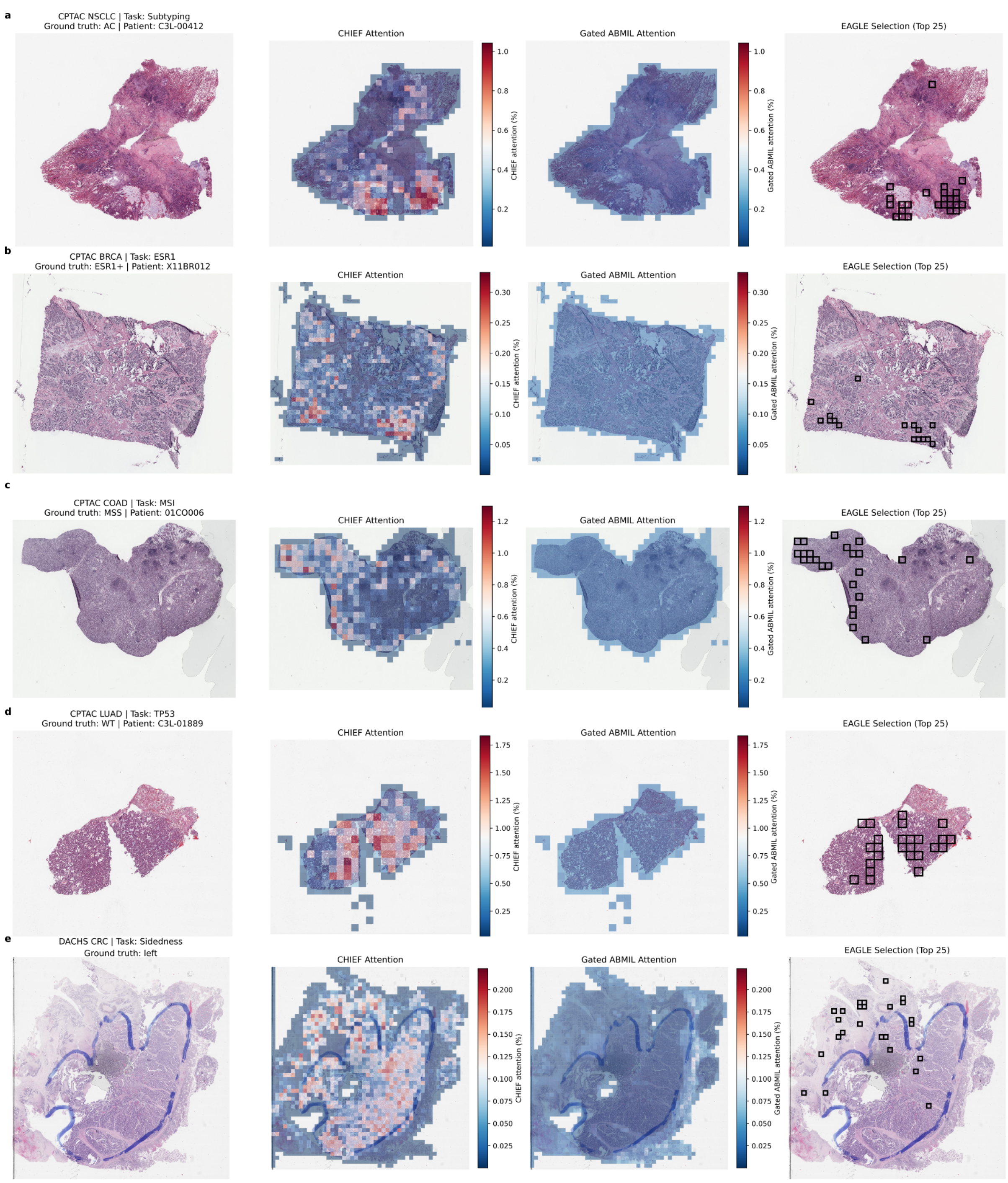

Same layout and task ordering as **Fig. 5**, showing additional representative external test cases for non-small cell lung cancer subtyping (**a**), ESR1 mutation prediction in CPTAC BRCA (**b**), microsatellite instability prediction in CPTAC COAD (**c**), TP53 mutation prediction in CPTAC LUAD (**d**), and colorectal cancer sidedness in DACHS (**e**). Columns show H&E thumbnail, CHIEF attention heatmap, gated ABMIL attention heatmap, and the EAGLE top-25 tile set visualized as black boxes on the H&E thumbnail. In contrast to **Fig. 5**, CHIEF and gated ABMIL heatmaps are displayed using a shared, matched color scale within each panel, enabling direct comparison of absolute attention magnitudes and highlighting that gated ABMIL attention is often diffuse and visually close to uniform under matched scaling. Heatmap intensities are shown after softmax normalization within each slide.

## Supplementary Fig. 7: Standard ABMIL attention heatmaps and EAGLE top-25 tile overlays

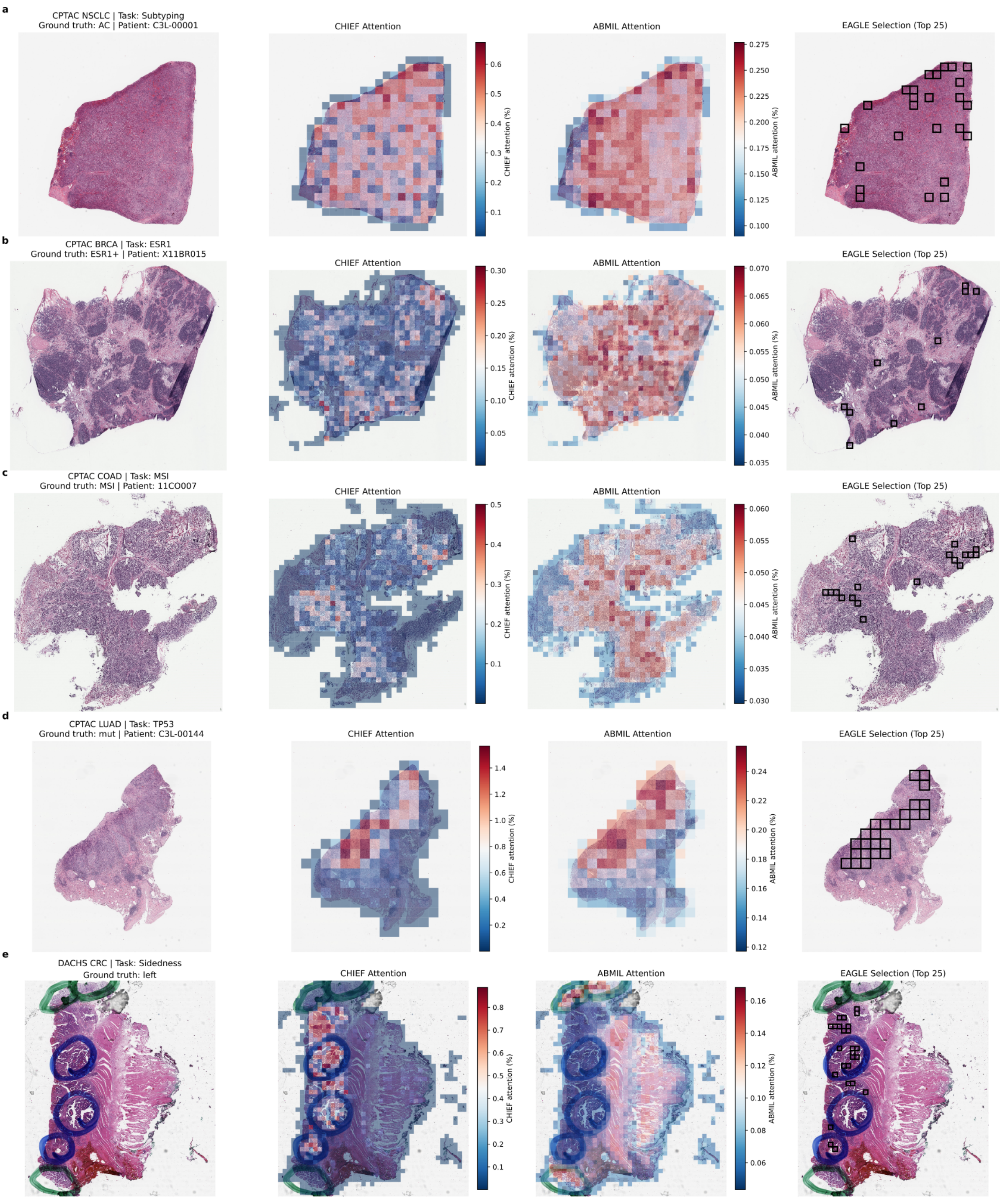

Same layout and task ordering as **Fig. 5**, showing additional representative external test cases for non-small cell lung cancer subtyping (**a**), ESR1 mutation prediction in CPTAC BRCA (**b**), microsatellite instability prediction in CPTAC COAD (**c**), TP53 mutation prediction in CPTAC LUAD (**d**), and colorectal cancer sidedness in DACHS (**e**). Columns show H&E thumbnail, CHIEF attention heatmap, standard ABMIL attention heatmap (task-specific model trained on Virchow2 embeddings), and the EAGLE top-25 tile set visualized as black boxes on the H&E thumbnail. Heatmap intensities are shown after softmax normalization within each slide. Color scales are normalized per model and per panel, as in **Fig. 5**, to visualize each method's within-slide dynamic range.